\documentclass{lcs2}

\usepackage{graphicx}
\usepackage{microtype}      
\usepackage{xcolor}         
\usepackage{amsmath}
\usepackage{amssymb}
\usepackage{mathtools}
\usepackage{amsthm}
\usepackage{amsmath}
\usepackage{amssymb}
\usepackage{mathtools}
\usepackage{amsthm}
\usepackage{booktabs}
\usepackage{multirow}
\usepackage{graphicx}
\usepackage{todonotes}
\usepackage[table, dvipsnames]{xcolor}
\usepackage{colortbl}
\usepackage{arydshln}
\usepackage{enumitem}
\usepackage{fancyvrb}
\usepackage{rotating}
\usepackage{subcaption}
\usepackage{natbib}
\usepackage{wrapfig}
\usepackage[capitalize,noabbrev]{cleveref}

\papergithub{https://github.com/parmanu-lcs2}  
\paperwebsite{https://parmanu.lcs2.in}  

\newcommand{\pruner}{\textsc{Sniper}}

\definecolor{blockblue}{RGB}{235, 245, 251}   
\definecolor{blockred}{RGB}{253, 237, 236}    
\definecolor{blockgreen}{RGB}{233, 235, 231}  
\definecolor{blockorange}{RGB}{254, 245, 231} 

\DeclareMathOperator*{\argmax}{arg\,max}

\theoremstyle{plain}

\theoremstyle{definition}

\theoremstyle{remark}

\papertitle{Unifying Depth and Width Pruning for LLMs \\via Binary Knapsack Optimization} 
\papershorttitle{Unifying Depth and Width Pruning for LLMs via Binary Knapsack Optimization}  

\papershortauthors{P. Goel, A. Sengupta, A. Nambi and T. Chakraborty} 
\paperauthors{%
  Palaash Goel \textsuperscript{\tiny 1}\quad
  Ayan Sengupta \textsuperscript{\tiny 2}\quad
  Akshay Nambi \textsuperscript{\tiny 3}\quad
  Tanmoy Chakraborty \textsuperscript{\tiny {1,2}}%
} 

\paperaffil{\textsuperscript{1} Yardi School of Artificial Intelligence, Indian Institute of Technology Delhi\\
\textsuperscript{2} Department of Electrical Engineering, Indian Institute of Technology Delhi\\
\textsuperscript{3} Microsoft Research (MSR) India\\} 
\paperemails{{goelpalaash@scai.iitd.ac.in, ayan.sengupta007@gmail.com, \\akshay.nambi@microsoft.com, tanchak@iitd.ac.in}} 

\paperkeywords{LLM compression, structured pruning, LLM efficiency, knapsack optimization, dual-axis pruning} 

\paperabstract{
Structured pruning is a promising approach for compressing large language models (LLMs), yet existing methods rely heavily on greedy heuristics that produce myopic decisions, and often fail to precisely meet target compression budgets. 
We present \pruner, a two-stage structured pruning framework that solves a knapsack optimization over coarse-granularity components to
yield conditionally optimal parameter allocations with respect to fixed importance estimates, followed by a fine-grained pruning stage to meet strict budget constraints.
We introduce the Compression Ratio Adherence Factor (\texttt{CRAFT}) to quantify budget fidelity, showing that while existing pruners deviate from target compression ratios by up to 33\%, \pruner\ achieves near-exact adherence with a \texttt{CRAFT} score of 0.98. Evaluations across four diverse architectures over a set of 18 tasks spanning five domains demonstrate \pruner's consistent improvements in average performance retention and task-level stability over six state-of-the-art pruners. Across all pruning configurations, \pruner\ achieves an excellent mean rank of 1.25, indicating its robust cross-architectural generalizability and excellent reliability. 
}

\appendixtocon 
\appendixtocname{Structure of the Appendix}   
\begin{document}
\makelabtitle

\section{Introduction}
\label{sec:intro}

Large language models (LLMs) have become the foundation of modern natural language processing, driving advances in reasoning, generation, and comprehension across a wide range of tasks~\citep{grattafiori2024llama3herdmodels,qwen3technicalreport,deepseekai2024deepseekv3technicalreport,openai2025gptoss120bgptoss20bmodel}. These capabilities, however, come at the cost of rapidly increasing parameter counts, posing significant challenges for deployment on resource-constrained hardware such as edge devices and on-device accelerators~\citep{zhu2024surveymodelcompressionlarge,nguyen_survey_2024}. While techniques such as quantization~\citep{bhandare2019efficient8bitquantizationtransformer,dettmers2023qloraefficientfinetuningquantized,ding2025cbqcrossblockquantizationlarge} and knowledge distillation~\citep{gu2023knowledge,sengupta2023good} reduce memory footprint or training cost, they largely preserve the original model architecture and therefore do not yield proportional inference speedups. In contrast, structured model pruning directly removes parameter groups, offering a principled route to practical acceleration.

\begin{figure*}[!htb]
    \centering
    \begin{subfigure}[t]{0.45\linewidth}
        \centering
    \includegraphics[width=\linewidth]{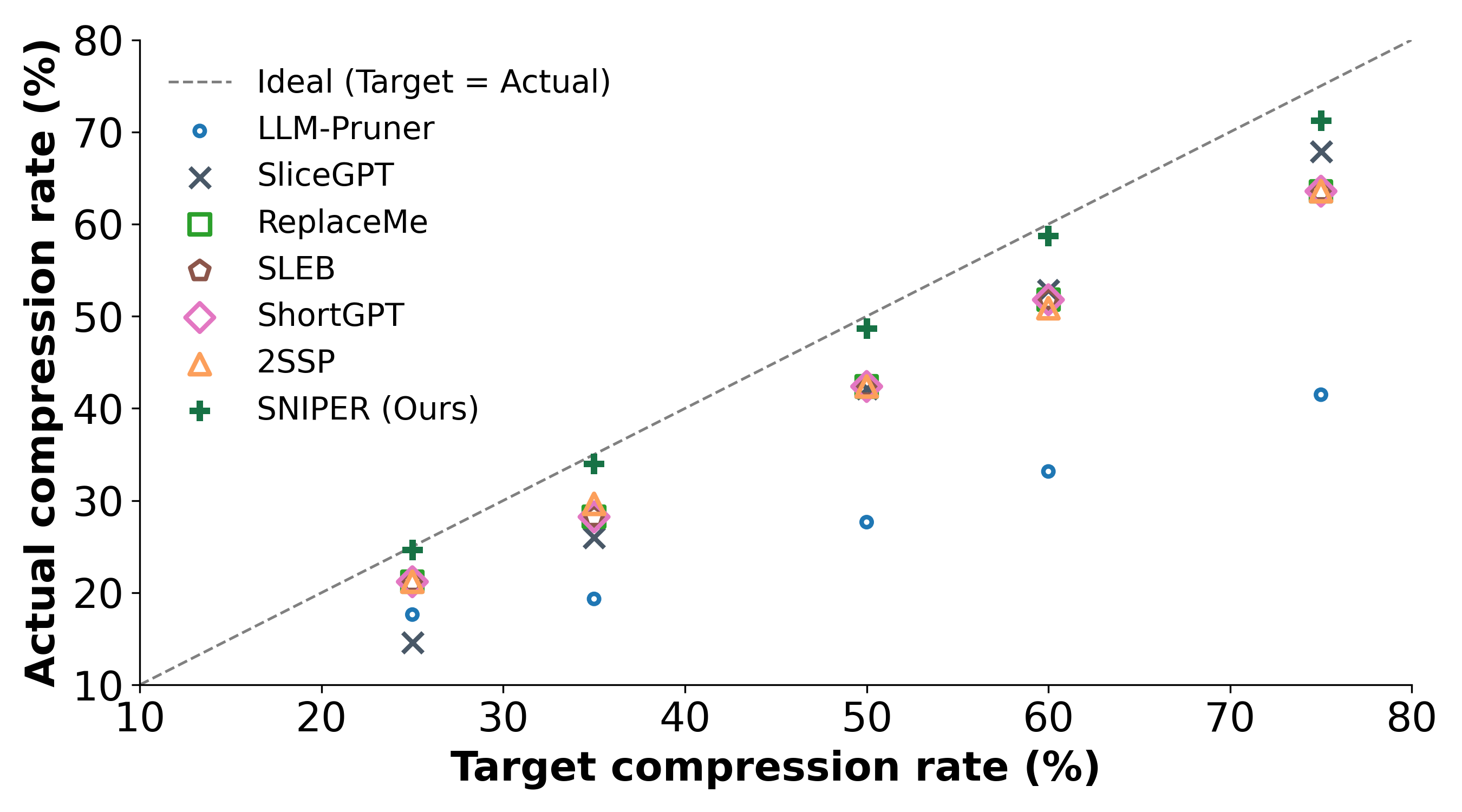}
        \caption{Target v/s actual compression ratio.}
        \label{fig:motivation-budget-drift}
    \end{subfigure}
    \begin{subfigure}[t]{0.45\linewidth}
        \centering
        \includegraphics[width=\linewidth]{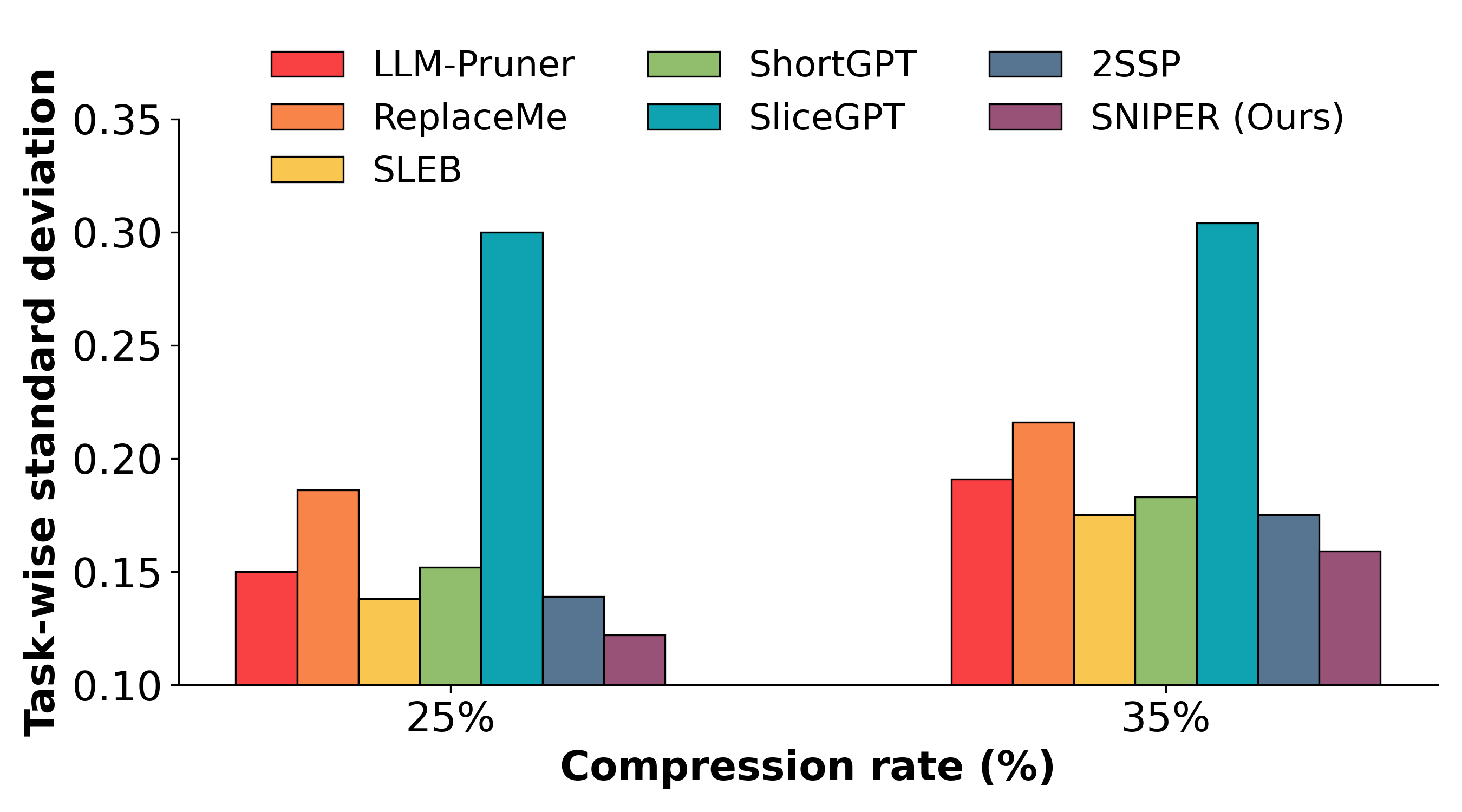}
        \caption{Comparison of task-wise standard deviation.}
        \label{fig:motivation-taskwise-std-dev}
    \end{subfigure}
    
    \caption{(a) Existing structured pruners exhibit severe ``capacity slack'', with the actual compression ratio deviating from the target by up to $33\%$. \pruner\ achieves near-exact budget adherence with a \texttt{CRAFT} of $0.98$, compared to the baseline average of $0.78$. Quantitative results are provided in Table~\ref{tab:craft-analysis} of Appendix~\ref{appendix:craft-detailed-results}. (b) \pruner\ reduces task-wise standard deviation by up to $60\%$ relative to prior methods, mitigating the instability induced by existing pruning strategies.}
    \label{fig:motivation}
\end{figure*}

Despite their appeal, existing structured pruning methods suffer from fundamental limitations arising from the interplay between pruning granularity and greedy optimization. Depth-based pruners \citep{shopkhoev2025replaceme0, men2024shortgptlayerslargelanguage, sleb, gromov2025the} operate on coarse atomic units and often fail to adhere to target compression ratios, while width-based methods \citep{ma2023llmpruner, sengupta2025pruneoncedesigningcalibrationfree, ashkboos2024slicegptcompresslargelanguage} induce sparsity-driven irregularities in weight tensor shapes, leading to minimal inference speedup \citep{kim2024shortenedllamadepthpruning, slidingwindowmerging}. Additionally, existing methods tend to rely on local and greedy heuristics that ignore inter-component dependencies across layers and may discard components that appear weak individually but are globally important. These design choices lead to three recurring failure modes: (i) calibration-induced distributional bias~\citep{jibeaware2025} and high task-wise deviations outside the calibration distribution (Figure~\ref{fig:motivation-taskwise-std-dev} shows deviation grows to $30\%$); (ii) irreversible local pruning decisions that cannot guarantee optimal architectural configurations, even with respect to their own importance estimates; and (iii) unreliable adherence to target compression ratios, undermining deployment on memory-constrained systems (Figure~\ref{fig:motivation-budget-drift} highlights deviations upto $33\%$). 

We address these limitations in this work and summarize our contributions as follows:
\begin{itemize}[leftmargin=*, itemsep=0pt, parsep=0pt, topsep=0pt]
    \item We propose \pruner\ (\underline{\textbf{S}}tructured K\underline{\textbf{n}}apsack-opt\underline{\textbf{i}}mization-based \underline{\textbf{P}}run\underline{\textbf{er}}), a novel dual-axis structured pruning framework for LLMs. \pruner\ first compresses along the depth-axis by solving a 0/1 knapsack problem over coarse-grained components, then applies a width-pruning phase that lets the model hit the target compression ratio near-perfectly. 
    \item Unlike existing systems that rely on greedy and myopic pruning decisions, \pruner\ utilizes dynamic programming to guarantee a conditionally optimal selection of coarse-grained components while pruning along the depth-axis.  
    \item We introduce the Compression Ratio Adherence Factor (\texttt{CRAFT}) to quantify budget fidelity, and show that \pruner\ achieves near-exact compression targets, whereas existing structured pruners suffer from substantial capacity slack, thereby undermining utility.
    \item We construct a comprehensive evaluation suite spanning six modern baselines tested across 18 tasks, five domains, and four diverse architectures -- including dense, reasoning-specialized, fused-MLP, and mixture-of-experts models. \pruner\ consistently achieves state-of-the-art performance retention, and lower task-level variance under multiple compression regimes.
\end{itemize}

\section{Related Work}
\label{sec:related_work}

The escalating computational demands of Large Language Models (LLMs) have catalyzed extensive research into efficient deployment strategies. Existing approaches generally fall into the categories of quantization \citep{bhandare2019efficient8bitquantizationtransformer, kim-integer-only-quantization, Liu_2023, guan_2024, ding2025cbqcrossblockquantizationlarge}, knowledge distillation \citep{hinton2015distillingknowledgeneuralnetwork, jiao-etal-2020-tinybert, wang2025metalearnedmodalityweightedknowledgedistillation}, and activation sparsity \citep{Dhar_2024, luo2025sparsinglawlargelanguage, liu2025trainingfreeactivationsparsitylarge, zhang2025rsparserankawareactivationsparsity}. While effective, these methods typically preserve the original model architecture. In contrast, \textit{model pruning} explicitly removes parameters to achieve tangible inference speedups and memory reductions.

\subsection{Unstructured Model Pruning} Early research predominantly focused on \textit{unstructured} or \textit{semi-structured} sparsity, identifying individual weights for removal based on magnitude \citep{han2015learningweightsconnectionsefficient, sun2024simpleeffectivepruningapproach} or second-order gradient information \citep{optimalbrainsurgeon, frantar2023sparsegptmassivelanguagemodels, ouderaa2024the}. Although methods like SparseGPT \citep{frantar2023sparsegptmassivelanguagemodels} and Wanda \citep{sun2024simpleeffectivepruningapproach} achieve high sparsity with minimal perplexity degradation, they typically produce irregular sparse matrices. Consequently, specialized hardware accelerators or N:M sparsity support are often required to translate theoretical FLOPs reductions into actual latency gains \citep{mishra2021acceleratingsparsedeepneural}.

\subsection{Structured Pruning for LLMs} To circumvent the hardware dependencies of unstructured sparsity, recent work has shifted towards \textit{structured pruning}, which removes coherent architectural units. We categorize these approaches based on their granularity:

\paragraph{Width Pruning.} These methods prune along the channel or embedding dimension, effectively narrowing the model's matrices. Techniques range from pruning individual attention heads \citep{voita-etal-2019-analyzing-heavy-lifting, aresixteenheadsbetterthanone, guo2025slimllm} to removing entire neurons in MLPs \citep{sengupta2025pruneoncedesigningcalibrationfree}. LLM-Pruner \citep{ma2023llmpruner} advances this by constructing a dependency graph to ensure structurally coupled parameters are pruned simultaneously. Similarly, SliceGPT \citep{ashkboos2024slicegptcompresslargelanguage} projects weight matrices into a lower-dimensional space via PCA, effectively slicing rows and columns. Other decomposition-based methods, such as FWSVD \citep{hsu2022language}, ASVD \citep{yuan2023asvd}, and SVD-LLM \citep{wang2024svd}, adopt greedy strategies to select and drop redundant neuron blocks from MLP and self-attention modules. While effective for fine-grained compression, width pruning often struggles to fully excise large, redundant computational blocks.

\paragraph{Depth Pruning.} Operating at a coarser granularity, depth pruning removes entire transformer layers \citep{gromov2025the, chen2025streamlining, chen2025dlp}. Approaches such as ShortGPT \citep{men2024shortgptlayerslargelanguage} and SLEB \citep{sleb} identify and discard redundant layers using block influence metrics. ReplaceMe \citep{shopkhoev2025replaceme0} mitigates the representation collapse caused by layer removal by substituting contiguous blocks with a learned linear projection. Dynamic schemes, such as PuDDing \citep{wee2025promptbased}, learn input-conditioned routing policies for selective layer skipping at inference time. Recent hybrid approaches \citep{yang2025moregreencodelarge, chen2025dlp} attempt to jointly prune across neurons, heads, and layers, guided by learned or internal importance signals.

\subsection{Limitations of Prior Work} Most aforementioned structured approaches share a critical limitation: reliance on \textit{greedy} selection criteria. Decisions are typically made locally (e.g., layer-by-layer) without an optimality-centric view of the parameter-performance trade-off. 
Furthermore, methods tend to not adhere to the required compression budget, introducing significant ``capacity slack'', with deviations of up to $33\%$ from the target budget (Figure~\ref{fig:motivation-budget-drift}). \pruner\ addresses these gaps by pruning along two axes: first along the depth wherein component selection is driven by a \textit{conditionally optimal} knapsack optimization algorithm, followed by a width-pruning stage that ensures that the target compression budget is met with high accuracy.
\section{Methodology}
\label{sec:methodology}

\begin{figure*}[t]
    \centering
    {\includegraphics[width=0.9\linewidth]{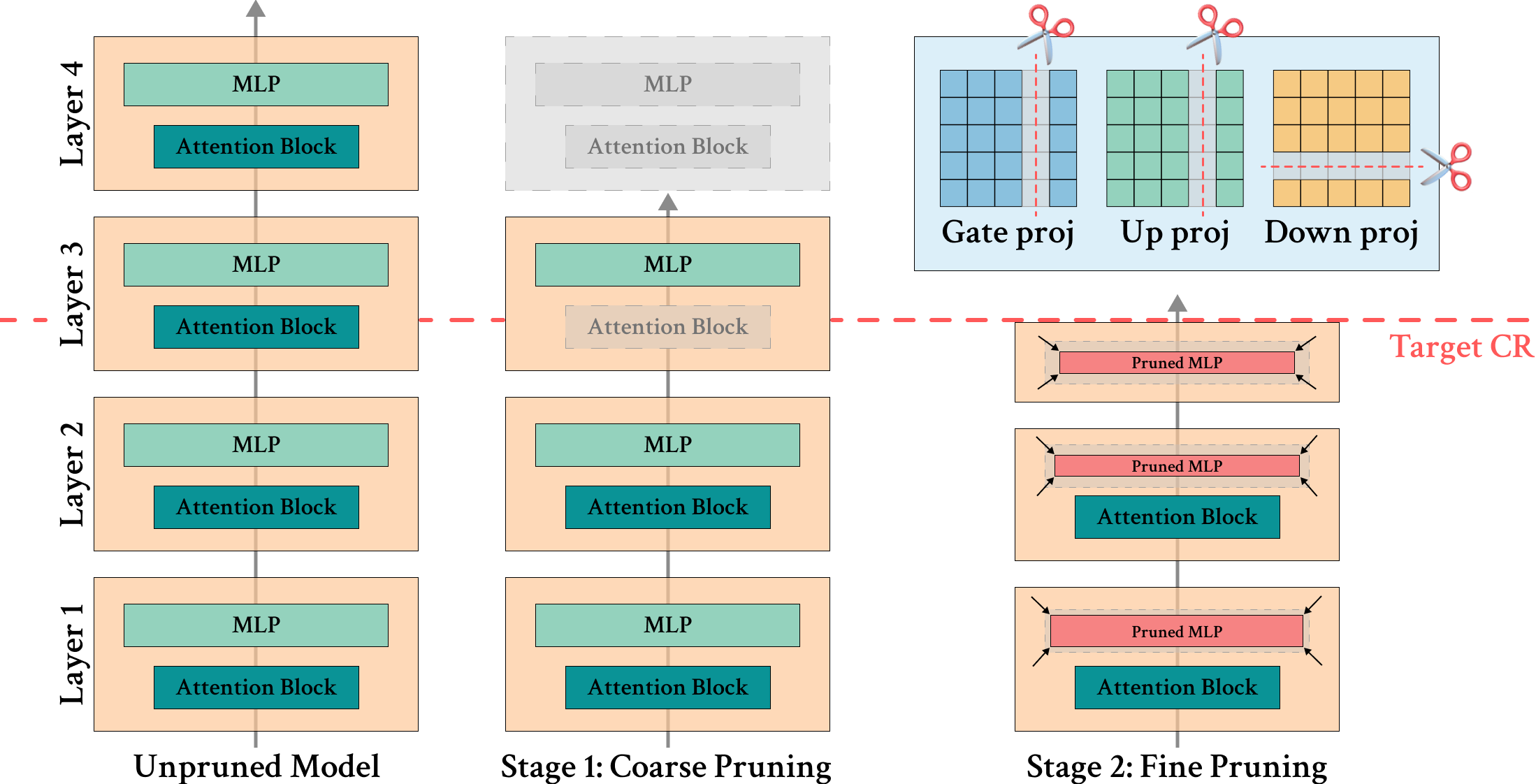}}
    \caption{A schematic overview of \pruner\ which performs structured pruning in two stages under a strict target compression ratio (CR). \textbf{Stage 1: Coarse pruning} removes entire redundant components such as full transformer layers or attention blocks, enabling notable efficiency gains while preserving architectural coherence. \textbf{Stage 2: Fine pruning} then operates within the remaining layers, selectively pruning the internal dimensions of the model's MLPs to precisely meet the target CR. This dual-axis strategy allows \pruner\ to combine the efficiency of depth pruning with the precision of width pruning, eliminating capacity slack and avoiding greedy, layer-wise decisions.} 
    \vspace{-2mm}
    \label{fig:schematic}
\end{figure*}

In this section, we provide a detailed account of the dual-stage pruning mechanism utilized by \pruner 
, as illustrated in Figure~\ref{fig:schematic}.  

\subsection{Problem Formulation}
\label{sec:problem_formulation}

Consider an LLM $\mathcal{M}$ consisting of a sequence of $L$ transformer layers. We decompose the model into a set of discrete, prunable components $\mathcal{X}$. 
Specifically, we adopt a \textit{mixed-granularity} decomposition wherein each transformer layer $l$ is decomposed into its constituent sub-blocks: the Multi-Head Attention block ($A_l$) and the Feed-Forward Network block ($M_l$). Thus, the universe of items is defined as: $\mathcal{X} = \bigcup_{l=1}^{L} \{ A_l, M_l \} \cup \{ T_l \}$, where $T_l$ represents the atomic retention of the \textit{entire} layer $l$. Since retaining $T_l$ effectively implies retaining both $A_l$ and $M_l$, it follows that for any layer $l$, the choice is mutually exclusive between selecting the composite item $T_l$ or a subset of its components $\{A_l, M_l\}$.
Each component $x \in \mathcal{X}$ is associated with a weight $w(x) \in \mathbb{Z}^+$, representing its parameter count, and a value $v(x) \in \mathbb{R}$, 
representing its contribution to model performance (Section~\ref{sec:importance_estimation}).
Given a target parameter budget $C$, our objective is to identify the subset $\mathcal{S}^* \subseteq \mathcal{X}$ that maximizes total importance while satisfying the capacity constraint:

\begin{equation}
\label{eq:knapsack_obj}
    \mathcal{S}^* = \argmax_{\mathcal{S} \subseteq \mathcal{X}} \sum_{x \in \mathcal{S}} v(x) \quad \text{s.t.} \quad \sum_{x \in \mathcal{S}} w(x) \le C
\end{equation}

This formulation maps the coarse-grained pruning stage directly to the 0/1 Knapsack Problem \citep{kellerer2013knapsack}.


\subsection{Stage 1: Coarse-Grained Pruning via Dynamic Programming}
\label{sec:coarse_pruning}

To solve the optimization problem in Equation~\ref{eq:knapsack_obj}, we employ a dynamic programming approach. Let the components be indexed sequentially (due to the inherent hierarchy in a typical transformer model, the components can be deterministically sequenced) $i = 1, \dots, N$, where $N = |\mathcal{X}|$. We define the state function $f(i, j)$ as the maximum importance value achievable using a subset of the first $i$ items subject to a capacity limit $j$. The recurrence relation is defined as:
\begin{equation}
\label{eq:dp_recurrence}
    f(i, j) = 
    \begin{cases} 
        f(i-1, j) & \text{if } w(x_i) > j \\
        \max \left(f(i-1, j), \Delta\right) & \text{if } w(x_i) \le j
    \end{cases}
\end{equation}
where, $\Delta  = f(i-1, j-w(x_i)) + v(x_i)$. The first case corresponds to excluding item $x_i$ due to capacity violation, while the second case evaluates the trade-off between exclusion and inclusion. To enforce the mutual exclusivity constraint (i.e., one cannot select both $T_l$ and $A_l$), we group mutually exclusive items and process them as a single decision step with multiple branches.

Notably, while this algorithm originally runs in $\Theta(N \cdot C)$ time which is intractable when $C$ is in the order of billions, we improve its tractability by discretizing all parameter counts by a discretizing factor, $\alpha$. This reduces the algorithm's time complexity to $\Theta(N\lfloor\frac{ C}{\alpha}\rfloor)$. As expected, increasing $\alpha$ makes the algorithm faster at the cost of assigning less precise (more coarsely rounded) parameter-count values to each component, leading to a suboptimal knapsack solution or a final compression ratio that deviates further from the target. We provide a detailed analysis of the same in Appendix~\ref{app:discretizing-factor-sweep}.

Upon computing the terminal state $f(N, \lfloor\frac{C}{\alpha}\rfloor)$, we backtrack to recover the optimal set $\mathcal{S}^*$. We note that $S^*$ does not represent a global optimum over all possible pruning configurations but rather a \textit{conditional optimum} with respect to fixed component-level importance estimates. Despite being a weaker guarantee of optimality, it surpasses what existing greedy methods provide since they do not offer any optimality guarantee whatsoever, conditional or otherwise.

\subsection{Iterative Importance Estimation}
\label{sec:importance_estimation}

A critical challenge in Knapsack-based pruning is assigning static importance values $v(x_i)$ to components that interact non-linearly. To address this, we propose an iterative importance estimation scheme based on marginal contribution.
Let $\mathcal{M}^{(i-1)}$ denote the model pruned up to component $i-1$. We define the importance of component $x_i$ as the divergence induced by removing it from $\mathcal{M}^{(i-1)}$. Let $\mathbf{Z} \in \mathbb{R}^{B \times V}$ be the logits produced by the full model on a calibration batch $\mathbf{X}$, where $B$ and $V$ are the batch size and vocabulary size of the model, respectively. For the $i$-th component, we evaluate two scenarios:
\begin{enumerate}[leftmargin=*, itemsep=0pt, parsep=0pt, topsep=0pt]
    \item \textbf{Retention:} The component is kept. Let $\mathbf{Z}_{\text{retain}}^{(i)}$ be the logits of the model where $x_i$ is present.
    \item \textbf{Drop:} The component is pruned. Let $\mathbf{Z}_{\text{drop}}^{(i)}$ be the logits of the model where $x_i$ is removed.
\end{enumerate}
The importance $v(x_i)$ is quantified as the marginal degradation in prediction fidelity:
\begin{equation}
\label{eq:importance_score}
    v(x_i) = \| \mathbf{Z} - \mathbf{Z}_{\text{drop}}^{(i)} \|_2^2 - \| \mathbf{Z} - \mathbf{Z}_{\text{retain}}^{(i)} \|_2^2
\end{equation}

Unlike greedy parameter importance estimation metrics that capture the importance of individual components in isolation, Equation~\ref{eq:importance_score} ensures that the importance scores reflect the effect of jointly pruning multiple components, thereby making the scoring function more reliable and representative.
We note that \pruner\ is similar to existing methods in that it utilizes heuristic proxies for component importance. However, it greatly improves upon these baselines by not making pruning decisions greedily and instead, computing \textit{provably optimal} compression configurations with respect to these importance scores.


\subsection{Stage 2: Fine-Grained Residual Pruning}
\label{sec:fine_grained}

The discrete nature of components in Stage 1 often leaves a residual capacity $\Delta C > 0$. To fully utilize the budget $C$, we introduce a second stage of fine-grained structured pruning targeting the MLPs within the transformer blocks.

\paragraph{Budget Allocation.}
We distribute the fine-grained pruning budget $N_{total}$ (total MLP columns to remove) across layers inversely proportional to their coarse importance. Let $\mathbf{u} \in \mathbb{R}^L$ be the vector of layer importance scores where $u_l = v(T_l)$. We define the layer-wise pruning ratio $\rho_l$ via a softmax over negated importance:
\begin{equation}
\label{eq:fine-grained-budget-distribution}
    \rho_l = \frac{\exp(-u_l / \tau)}{\sum_{k=1}^L \exp(-u_k / \tau)}
\end{equation}
where $\tau$ (defaulted to 1) is a temperature parameter. The number of columns to prune from layer $l$ is computed as $N_l = \lfloor \rho_l \cdot N_{total} \rfloor$.

\paragraph{Column Selection.}
Modern LLMs~\citep{grattafiori2024llama3herdmodels,qwen_qwen2._2025} utilize variants of the Gated Linear Unit (GLU), where the MLP computation is defined as:
\begin{equation}
    \text{MLP}(\mathbf{X}) = (\sigma(\mathbf{X}\mathbf{G}^T) \odot (\mathbf{X}\mathbf{U}^T))\mathbf{D}^T
\end{equation}
where $\mathbf{U}, \mathbf{G} \in \mathbb{R}^{d_{model} \times d_{ff}}$ are the up- and gate-projections, respectively, and $\mathbf{D} \in \mathbb{R}^{d_{ff} \times d_{model}}$ is the down-projection. Here, $d_{model}$ refers to the model's hidden dimension and $d_{ff}$ is the higher dimension to which embeddings are projected within its MLPs. To maintain structural consistency, pruning a column index $j$ requires removing the $j$-th column of $\mathbf{U}$ and $\mathbf{G}$, and the $j$-th row of $\mathbf{D}$.
We define the sensitivity of the $j$-th neuron index as the magnitude of its contribution to the output:
\begin{equation}
\label{eq:col-importance}
    \Omega_j = | \mathbf{D}_{j,:} \cdot (\mathbf{G}_{:,j} \odot \mathbf{U}_{:,j}) |
\end{equation}
Indices with the lowest $\Omega_j$ scores are pruned until $N_l$ columns are removed. This ensures that the residual capacity is filled with the least salient parameters, 
maxing out the available budget.

By integrating the conditionally optimal configurations computed by the Knapsack solver with the precision of fine-grained residual pruning, \pruner\ demonstrates high retention capabilities with respect to the unpruned model, while also ensuring high-fidelity adherence to the parameter budget, as reflected by its excellent CRAFTs (Appendix~\ref{appendix:craft-detailed-results}). 

\section{Experimental Setup}
\label{sec:experiments}

\paragraph{Pruned models.}
To evaluate architectural generality, we benchmark \pruner\ on a diverse set of modern LLMs\footnote{All pretrained checkpoints are obtained from the HuggingFace model hub.} with different architectural paradigms. These include the instruction-tuned LLaMA-3.1-8B-Instruct~\citep{grattafiori2024llama3herdmodels}, the reasoning-oriented Qwen3-8B~\citep{qwen3technicalreport}, Phi-4~\citep{abdin2024phi4technicalreport}, which employs fused MLP projections, and the Mixture-of-Experts model GPT-OSS-20B~\citep{openai2025gptoss120bgptoss20bmodel}. 

\paragraph{Baselines.} We compare \pruner\ against six contemporary structured pruning baselines. Width-based methods include SliceGPT~\citep{ashkboos2024slicegptcompresslargelanguage} and LLM-Pruner~\citep{ma2023llmpruner}, while depth-based methods include ReplaceMe~\citep{shopkhoev2025replaceme0}, SLEB~\citep{sleb}, and ShortGPT~\citep{men2024shortgptlayerslargelanguage}. We also include 2SSP \citep{sandri2025ssp} since it mirrors \pruner\ via its dual-axis approach, allowing us to evaluate \pruner's effectiveness with respect to analogous pruning approaches.
Several baselines are re-engineered to support recent LLM architectures\footnote{For instance, ReplaceMe and ShortGPT do not natively support newer LLM variants.}. 
We note that vector-space width pruners such as SliceGPT are inherently incompatible with expert routing layers and therefore cannot be applied to GPT-OSS. 
In order to ensure a fair evaluation, we adjust the compression ratios of all methods to yield pruned models of comparable sizes. We provide all implementation details in Appendix~\ref{appendix:implementation-details}.  

\paragraph{Evaluation task suite.}
Many pruning evaluations rely on a narrow set of tasks, which can obscure degradation in a model's linguistic capabilities, reasoning, or alignment.  
For a more comprehensive assessment, we evaluate all models on a testbed of 18 tasks spanning five domains. Generative performance is measured using log-perplexity on WikiText-2~\citep{merity2016pointer} and LAMBADA~\citep{paperno2016lambada}. World understanding is evaluated using PIQA~\citep{bisk2020piqa}, PROST~\citep{aroca-ouellette-etal-2021-prost}, and CommonsenseQA~\citep{talmor-etal-2019-commonsenseqa}. Domain-specific knowledge covers STEM reasoning tasks including ARC-Easy and ARC-Challenge~\citep{clark2018thinksolvedquestionanswering}, mathematical reasoning tasks such as MathQA~\citep{amini-etal-2019-mathqa} and OpenBookQA~\citep{OpenBookQA2018}, and medical question answering using MedQA~\citep{jin2021disease}. Natural language understanding is evaluated using BLiMP~\citep{warstadt2020blimp}, BoolQ~\citep{clark2019boolq}, Winogrande~\citep{sakaguchi2021winogrande}, and CoQA~\citep{reddy2019coqa}. To assess alignment preservation, we additionally evaluate safety and ethics using TruthfulQA~\citep{lin-etal-2022-truthfulqa}, Winogender~\citep{rudinger-etal-2018-gender}, and Moral Stories~\citep{emelin-etal-2021-moral}. A detailed description of each task is provided in Appendix~\ref{appendix:dataset-descriptions}.



\section{Results}
\label{sec:results}
Table~\ref{tab:main_results} compares \pruner\ against state-of-the-art structured pruning methods on LLaMA-3.1-8B-Instruct and Qwen3-8B, under two compression ratios (25\% and 35\%), both with and without recovery fine-tuning (RFT). Across \textit{all configurations except one}, \pruner\ achieves the highest average retention performance (Avg RP) while maintaining strong performance across different task groups.  
While it is marginally outperformed by ReplaceMe and ShortGPT on Llama-3.1-8B-Instruct (25\% compression with RFT), \pruner\ maintains a near-perfect mean rank of 1.25 across all pruning configurations, yielding competitive performance even at its worst rank of merely 3 (Tables~\ref{tab:main_results} and ~\ref{tab:method-ranking}). In contrast, every baseline has at least one configuration in which its performance deteriorates drastically. Unlike its baselines, \pruner\ \textit{never} disintegrates. 
This indicates \pruner's unparalleled robustness and generalization capabilities across different architectures and compression regimes. We provide detailed per-task results in Appendix~\ref{appendix:task-wise-performance}.

\begin{table*}[!htb]
\centering
\resizebox{\textwidth}{!}{%
\begin{tabular}{l||clccccc||cc}
\hline
\textbf{Model} & \textbf{CR} & \textbf{Method} & \textbf{Generative} & \textbf{World} & \textbf{Domain} & \textbf{NLU \& NLI} & \textbf{Safety} & \textbf{Avg RP (\%)} & \textbf{Std RP (\%)}\\
 & & & (Log-PPL $\downarrow$) & (Acc $\uparrow$) & (Acc $\uparrow$) & (Acc $\uparrow$) & (Score $\uparrow$) & ($\uparrow$) & ($\downarrow$)\\
\hline
\multirow{15}{*}{LLaMA-3.1-} & \multirow{1}{*}{-} & Base & 1.69 & 66.55 & 56.28 & 78.52 & 55.56 & - & -\\
\cdashline{2-10}
\multirow{15}{*}{8B-Instruct} & \multirow{7}{*}{25\%} & ReplaceMe & 4.91 (2.36) & 59.30 (58.31) & 40.78 (43.35) & 54.97 (75.80) & 55.62 (54.79) & 74.86 (\textbf{88.90}) & 20.40 (10.50) \\
 &  & SliceGPT & 5.11 (3.60) & 40.78 (49.10) & 26.87 (30.08) & 53.75 (62.49) & 50.23 (49.50) & 62.06 (72.49) & 20.20 (17.30) \\
 &  & LLM-Pruner & 3.96 (2.54) & 40.56 (44.37) & 32.14 (36.32) & 49.85 (62.43) & 51.87 (50.99) & 64.56 (75.11) & 22.40 (17.70) \\
 &  & SLEB & 3.37 (2.25) & 42.20 (47.52) & 37.06 (38.79) & 50.78 (65.98) & 51.01 (50.74) & 67.91 (78.45) & 20.30 (16.80) \\
 &  & ShortGPT & 13.20 (2.46) & 48.70 (59.22) & 31.68 (42.42) & 34.50 (74.77) & 51.01 (55.34) & 57.50 (88.47) & 26.10 (11.20) \\
 &  & 2SSP & 2.77 (2.13) & 49.33 (54.23) & 36.67 (39.75) & 68.14 (71.59) & 50.93 (51.42) & 76.65 (83.19) & 14.00 (13.90)\\ 
 &  & \pruner & 3.15 (2.29) & 51.81 (57.89) & 39.00 (41.71) & 64.95 (74.41) & 53.02 (53.97) & \textbf{77.03} (87.34) & \textbf{12.70} (\textbf{10.00}) \\
\cdashline{2-10}
& \multirow{7}{*}{35\%} & ReplaceMe & 5.86 (3.19) & 38.42 (39.81) & 29.01 (29.62) & 36.09 (50.55) & 51.24 (49.58) & 56.25 (65.16) & 27.10 (22.60) \\
 &  & SliceGPT & 6.44 (4.51) & 38.43 (44.77) & 25.68 (28.41) & 42.72 (55.22) & 50.86 (51.23) & 56.74 (68.08) & 22.50 (18.60) \\
 &  & LLM-Pruner & 4.94 (2.94) & 37.18 (38.49) & 29.61 (33.17) & 44.60 (58.50) & 51.33 (50.28) & 59.66 (69.93) & 23.70 (19.40) \\
 &  & SLEB & 4.47 (2.71) & 39.20 (40.57) & 30.99 (32.95) & 44.13 (57.87) & 50.87 (51.29) & 60.72 (70.58) & 24.10 (20.10) \\
 &  & ShortGPT & 13.14 (2.94) & 40.77 (43.64) & 28.99 (32.50) & 36.52 (69.91) & 55.03 (54.67) & 56.14 (76.98) & 28.00 (18.70) \\
 & & 2SSP & 3.77 (2.45) & 37.28 (47.62) & 29.39 (35.66) &  55.55 (68.03) & 50.50 (50.19) & 64.21 (77.26) & 19.10 (15.80)\\
 & & \pruner & 4.70 (2.63) & 41.16 (48.52) & 32.18 (34.73) & 53.52 (68.31) & 51.29 (50.95) & \textbf{64.96} (\textbf{77.26}) & \textbf{19.50} (\textbf{14.80}) \\
\cline{1-10}
\multirow{16}{*}{Qwen3-8B} & \multirow{1}{*}{-} & Base & 2.01 & 66.32 & 58.60 & 76.20 & 55.90 & - & - \\
\cdashline{2-10}
 & \multirow{7}{*}{25\%} & ReplaceMe & 4.37 (2.76) & 37.40 (43.98) & 30.21 (36.83) & 41.44 (55.52) & 50.47 (50.66) & 59.92 (72.30) & 23.80 (18.60) \\
 &  & SliceGPT & 15.72 (10.75) & 34.01 (34.48) & 24.64 (25.05) & 29.46 (29.54) & 51.00 (51.35) & 48.68 (50.07) & 27.50 (30.00) \\
 &  & LLM-Pruner & 3.20 (2.46) & 48.73 (53.44) & 31.85 (37.28) & 63.89 (67.05) & 50.99 (52.20) & 72.64 (80.11) & 15.90 (15.00) \\
 &  & SLEB & 3.75 (2.61) & 43.11 (50.41) & 36.93 (38.38) & 48.84 (64.50) & 51.48 (53.12) & 67.53 (78.89) & 20.20 (13.80) \\
 &  & ShortGPT & 8.98 (3.14) & 43.76 (55.87) & 33.09 (39.34) & 46.98 (67.98) & 55.02 (54.14) & 63.03 (80.21) & 22.90 (15.20) \\
 & & 2SSP & 6.37 (3.11) & 34.79 (50.10) & 26.31 (33.86) & 39.98 (58.74) & 52.19 (52.44) & 56.09 (72.92) & 23.10 (16.30)\\
 & & \pruner & 3.65 (2.72) & 50.81 (55.93) & 36.21 (41.36) & 61.04 (69.31) & 54.90 (53.70) & \textbf{74.41} (\textbf{82.64}) & \textbf{15.00} (\textbf{12.20}) \\
\cdashline{2-10}
& \multirow{7}{*}{35\%} & ReplaceMe & 5.99 (4.20) & 37.63 (38.12) & 32.30 (28.86) & 40.78 (52.27) & 51.47 (52.59) & 59.40 (64.00) & 25.90 (21.60) \\
 &  & SliceGPT & 16.13 (10.82) & 33.49 (34.39) & 24.78 (24.58) & 29.53 (29.87) & 50.99 (52.44) & 48.67 (50.17) & 27.70 (30.40) \\
 &  & LLM-Pruner & 5.25 (3.46) & 39.24 (45.25) & 26.47 (31.86) & 50.05 (55.02) & 50.75 (51.39) & 60.63 (68.34) & 21.90 (19.10) \\
 &  & SLEB & 4.94 (3.10) & 38.46 (45.15) & 31.95 (34.40) & 38.99 (53.56) & 51.55 (52.43) & 59.67 (70.47) & 24.70 (17.50) \\
 &  & ShortGPT & 10.67 (3.98) & 45.56 (50.25) & 30.96 (34.82) & 38.09 (59.54) & 54.02 (53.11) & 58.69 (72.27) & 27.10 (18.30) \\
 & & 2SSP & 7.42 (3.56) & 34.71 (46.37) & 25.92 (31.94) & 36.63 (56.68) & 51.57 (51.68) & 54.15 (69.23) & 24.40 (17.50)\\
 &  & \pruner & 5.11 (3.21) & 44.04 (50.81) & 30.43 (35.45) & 48.68 (63.35) & 52.76 (52.81) & \textbf{63.71} (\textbf{75.13}) & \textbf{20.10} (\textbf{15.90}) \\
\hline
\end{tabular}%
\vspace{-3mm}
}
\caption{Comparison of different pruning methods across various compression ratios (CR) without and (with) recovery fine-tuning (RFT). 
We report the scores for each task group, the average retention performance (RP) (ratio of pruned and base model performances), as well as the standard deviation across tasks for each method.
Per-task results are highlighted in Tables~\ref{tab:main_results_llama_full_norft}, \ref{tab:main_results_llama_full_rft}, \ref{tab:main_results_qwen_full_norft} and \ref{tab:main_results_qwen_full_rft} of Appendix~\ref{appendix:task-wise-performance}.}
\label{tab:main_results}
\end{table*}

\subsection{Consistency and robustness}
Beyond average performance, \pruner\ exhibits substantially lower task-level standard deviations (Std RP) across \textit{all configurations}, even including settings where it does not outperform all baselines. For example, at 35\% compression on Qwen3-8B with RFT, \pruner\ reduces task-wise variance to $15.90\%,$ compared to 17--30\% for competing pruners. This behavior highlights \pruner's ability to avoid a common failure mode of greedy pruning strategies, which often over-optimize for isolated metrics (e.g., perplexity) at the expense of broader task performance. By combining the benefits of conditional optimality during coarse-grained pruning with that of a budget-filling fine-grained stage, \pruner\ is able to maintain superior cross-task consistency across diverse model architectures.
\begin{table}[!htb]
\centering
\resizebox{0.5\textwidth}{!}{%
\begin{tabular}{c|c|c|c}
\hline
\textbf{Method} & \textbf{Mean Rank} & \textbf{Best Rank} & \textbf{Worst Rank} \\ 
& $(\downarrow)$ & $(\downarrow)$ & $(\downarrow)$\\\hline
ReplaceMe       & 4.75                                    & 1                                       & 7                                        \\ 
SliceGPT        & 6.50                                    & 5                                       & 7                                        \\ 
LLM-Pruner      & 4.00                                    & 2                                       & 6                                        \\ 
SLEB            & 3.63                                    & 3                                       & 5                                        \\ 
ShortGPT        & 3.70                                    & 2                                       & 7                                        \\ 
2SSP            & 3.55                                    & 1                                       & 6                                        \\ 
 \pruner          & 1.25                                    & 1                                       & 3                                        \\ \hline
\end{tabular}
}
\caption{Ranking different pruning methods across all available configurations.}
\label{tab:method-ranking}
\end{table}

\begin{table*}[!htb]
\centering
\resizebox{\textwidth}{!}{%
\begin{tabular}{l|clccccc|cc}
\hline
\textbf{Model} & \textbf{CR} & \textbf{Method} & \textbf{Generative} & \textbf{World} & \textbf{Domain} & \textbf{NLU \& NLI} & \textbf{Safety} & \textbf{Avg RP (\%)} & \textbf{Std RP (\%)} \\
 & & & (Log-PPL $\downarrow$) & (Acc $\uparrow$) & (Acc $\uparrow$) & (Acc $\uparrow$) & (Score $\uparrow$) & ($\uparrow$) & ($\downarrow$) \\
\hline
\multirow{4}{*}{Phi-4-14B} & - & Base & 1.65 & 70.11 & 56.73 & 79.89 & 59.46 & - & - \\
\cdashline{2-10}
& \multirow{3}{*}{35\%} & ShortGPT & 3.35 & 54.18 & 40.05 & 65.39 & 57.44 & 77.18 & 16.60\\
& & 2SSP & 3.07 & 47.61 & 33.35 & 62.66 & 51.62 & 70.16 & 16.20\\
&  & \pruner & 2.55 & 58.07 & 44.14 & 70.10 & 53.70 & \textbf{81.58} & \textbf{12.60}\\
\hline
\multirow{4}{*}{GPT-OSS-20B} & - & Base & 2.22 & 65.65 & 56.07 & 75.37 & 58.16 & - & - \\
\cdashline{2-10}
& \multirow{3}{*}{35\%} & ShortGPT & 3.06 & 46.18 & 38.03 & 45.16 & 52.24 & 70.51 & 22.10\\
& & 2SSP & 2.99 & 56.16 & 43.19 & 68.99 & 56.38 & 84.68 & 11.40\\
&  & \pruner & 2.71 & 55.62 & 45.28 & 69.94 & 55.47 & \textbf{86.99} & \textbf{8.80}\\
\hline
\end{tabular}%
}
\caption{Performance analysis on larger and architecturally diverse models -- Phi-4 and GPT-OSS-20B with RFT. Detailed results provided in Tables~\ref{tab:main_results_phi_full_rft} and \ref{tab:main_results_gpt_full_rft} of Appendix~\ref{appendix:task-wise-performance}.
}
\label{tab:phi-gpt-results}
\end{table*}

\subsection{Results on larger and diverse architectures}
Table~\ref{tab:phi-gpt-results} evaluates \pruner\ on larger and architecturally distinct models, i.e, Phi-4-14B and GPT-OSS-20B, under a 35\% compression regime with RFT. Despite these departures from standard dense transformers, \pruner\ consistently attains the highest Avg RP while maintaining the lowest Std RP. On Phi-4-14B, \pruner\ attains an Avg RP of $81.58\%$, substantially outperforming ShortGPT and 2SSP by $4.4\%$ and $11.42\%$, respectively. \pruner\ demonstrates similar superiority on GPT-OSS-20B, achieving an Avg RP of $86.99\%$ - $2.31\%$ higher than 2SSP and $16.48\%$ better than ShortGPT. 
Notably, on GPT-OSS-20B, 2SSP operates solely via its first stage due to how it distributes the pruning budget between its two stages, effectively turning its second stage into a no-op. In contrast, \pruner\ ensures that it leverages both of its stages to yield a superior compressed model.  
\subsection{Adherence to Compression Ratio} We analyze the effect of fixing the target compression ratio for each method and observing its adherence to the same. We observe that existing methods demonstrate significant deviations of upto 33\% from the target ratio, leading to severe under-pruning and an erosion of trust in the pruning process (Figure~\ref{fig:motivation} and Appendix~\ref{appendix:craft-detailed-results}). In contrast, \pruner\ exhibits remarkable reliability by adhering strictly to these targets with a near-ideal \texttt{CRAFT} of 0.98 across a wide range of budgets. Unlike its baselines, \pruner\ eliminates reliance on hit-and-trial strategies to obtain a model of a specific size, leading to significant practical advantages.   

\subsection{Ablation on \pruner}
\begin{figure} 
    \centering
    \includegraphics[width=0.5\textwidth]{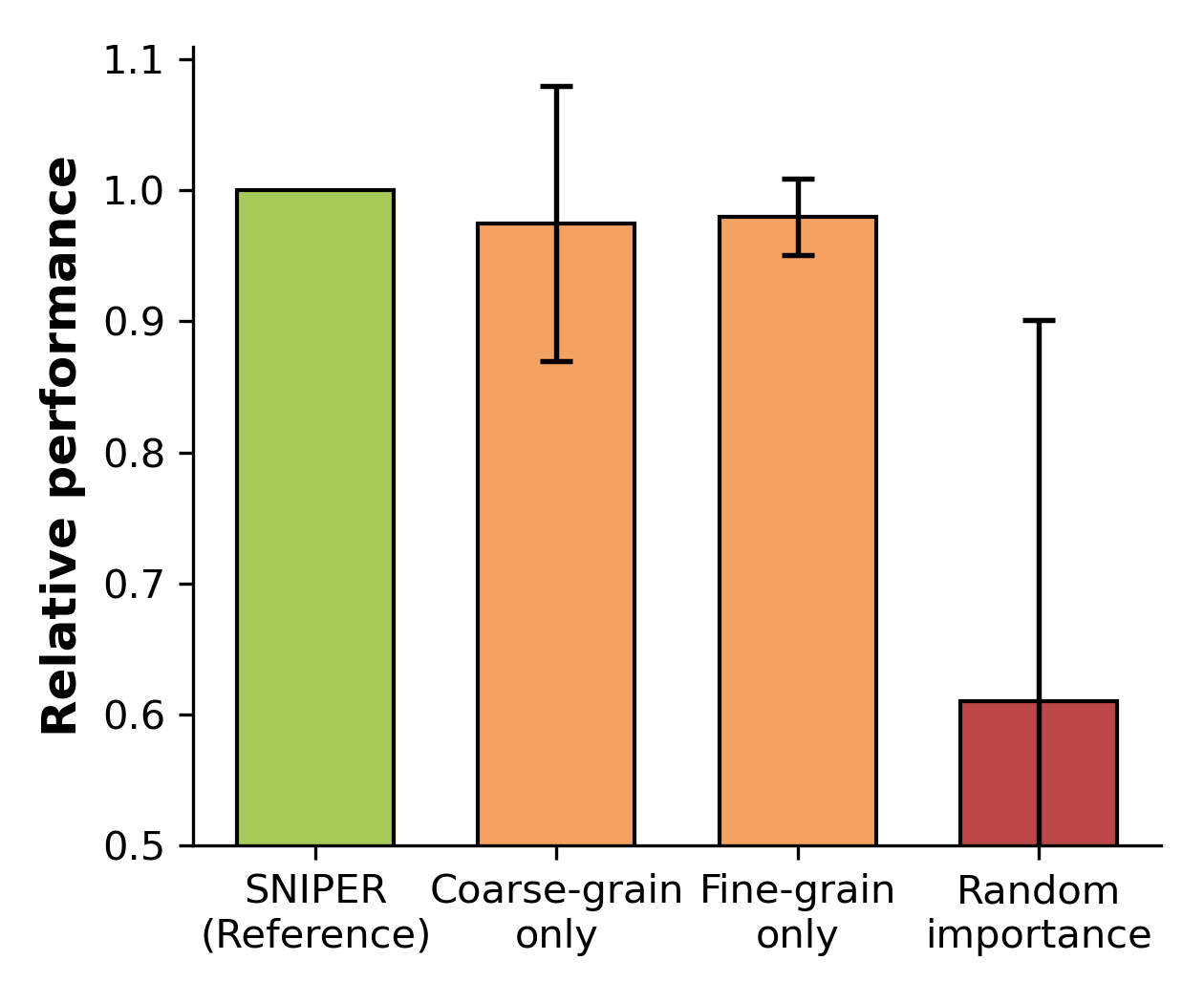}
    \caption{Ablation study of \pruner\ showing the effect of coarse and fine pruning stages. Combining both stages yields the best performance and stability, while removing either component leads to lower performance retention or instability.}
    \label{fig:ablation}
\end{figure}

We provide additional ablation experiments in Table~\ref{table:additional-ablations} and Figure~\ref{fig:ablation} and analyze the effectiveness of various components and design choices of \pruner.

\paragraph{Dual-Axis Pruning.}
We ablate \pruner's coarse- and fine-grained components, measuring Avg RP and Std RP (Figure~\ref{fig:ablation}). Coarse pruning alone reaches a strong Avg RP of 97.47\%, but its high Std RP (10.50\%) reflects a key limitation: operating on entire structural groups rather than individual neurons causes some important neurons to inevitably be pruned, destabilizing downstream performance. Its structured removals, however, preserve weight tensor shapes, making it the primary driver of inference speedup.
The fine-grained variant attains a marginally better Avg RP (97.95\%) with far lower Std RP (2.90\%), confirming that precise, neuron-level interventions better preserve cross-task consistency. However, it produces irregularly-shaped weight tensors, yielding little inference speedup.
The two stages are thus complementary: coarse pruning delivers hardware-friendly speedups at the cost of consistency, while fine-grained pruning restores consistency but lacks efficiency gains due to induced irregularities. 
Additionally, random importance assignment yields a low Avg RP (63.24\%) and high Std RP ($26.90\%$), confirming the necessity of principled importance estimation. 
\begin{table*}[t]
\centering
\resizebox{\textwidth}{!}{%
\begin{tabular}{cccccc|cc}
\hline
\textbf{Variant} & \textbf{Generative} & \textbf{World} & \textbf{Domain} & \textbf{NLU \& NLI} & \textbf{Safety} & \textbf{Avg RP (\%)} & \textbf{Std RP (\%)} \\
& (Log-PPL $\downarrow$) & (Acc $\uparrow$) & (Acc $\uparrow$) & (Acc $\uparrow$) & (Score $\uparrow$) & ($\uparrow$) & ($\downarrow$) \\
\hline
\pruner-MAG & 3.02 & 54.76 & 39.61 & 64.48 & 52.49 & 78.87 & 13.80 \\
\pruner-ACT & 2.74 & 53.81 & 40.27 & 65.90 & 54.23 & 80.87 & 13.80 \\
\pruner-LeaveOneOut & 4.08 & 37.33 & 29.11 & 47.93 & 51.44 & 62.01 & 23.30 \\
\pruner-Uniform & 2.77 & 52.40 & 41.19 & 68.22 & 53.51 & 81.26 & 13.70 \\
\hline
\pruner & \textbf{2.72} & \textbf{55.93} & \textbf{41.36} & \textbf{69.31} & \textbf{53.70} & \textbf{82.64} & \textbf{12.20} \\
\end{tabular}%
}
\caption{Additional ablation results for \pruner\ obtained after compressing Qwen3-8B by 25\% (with RFT).}
\label{table:additional-ablations}
\end{table*}

\paragraph{Column Selection Heuristic.} We analyze the efficacy of our column selection heuristic (Equation~\ref{eq:col-importance}) by replacing it with two popular alternatives: \textit{(a)} magnitude-based \citep{han2015learningweightsconnectionsefficient} and \textit{(b)} activation-aware \citep{sun2024simpleeffectivepruningapproach} pruning, corresponding to the \pruner-MAG and \pruner-ACT variants in Table~\ref{table:additional-ablations}, respectively. Both variants perform approximately 2--4\% worse than \pruner\ on average and exhibit 13\% higher per-task deviation, demonstrating the effectiveness of our strategy at maintaining strong and stable performance.

\paragraph{Iterative Pruning for Importance Estimation.} To estimate component importance, \pruner\ computes the marginal degradation in performance incurred by dropping a component from a partially pruned model, whose pruning configuration is determined by our dynamic programming solver (Section~\ref{sec:importance_estimation}). We compare this against the leave-one-out strategy employed by baselines such as SLEB \citep{sleb}, denoted \pruner-LeaveOneOut. This variant incurs a significant drop of 20\% in average performance and nearly twice the task-wise standard deviation of \pruner, demonstrating that estimating component importance on iteratively optimal pruning configurations captures global inter-component dependencies—something leave-one-out strategies fail to model due to their limited scope.

\paragraph{Fine-Grained Pruning Budget Distribution.} To determine how many columns to drop from each MLP in Stage 2 (Section~\ref{sec:fine_grained}), \pruner\ uses the component importance scores from Section~\ref{sec:importance_estimation} (Equation~\ref{eq:importance_score}) to distribute the pruning budget inversely proportional to importance (Equation~\ref{eq:fine-grained-budget-distribution}), pruning redundant MLPs more aggressively while compressing salient ones more conservatively. We evaluate the \pruner-Uniform variant, which distributes the fine-grained pruning budget uniformly across all MLPs. As shown in Table~\ref{table:additional-ablations}, this variant not only incurs a 1.5\% drop in average performance but, more critically, exhibits 12.3\% higher task-wise standard deviation, indicating uneven performance retention across domains.

\section{Discussion}
\label{sec:discussion}
\begin{table*}[!htb]
\centering
\begin{subtable}[t]{\textwidth}
\resizebox{\textwidth}{!}{%
\begin{tabular}{ccccccc|cc}
\hline
\textbf{\# Calibration Samples} & \textbf{Method} & \textbf{Generative} & \textbf{World} & \textbf{Domain} & \textbf{NLU \& NLI} & \textbf{Safety} & \textbf{Avg RP (\%)} & \textbf{Std RP (\%)} \\
 & & (Log-PPL $\downarrow$) & (Acc $\uparrow$) & (Acc $\uparrow$) & (Acc $\uparrow$) & (Score $\uparrow$) & ($\uparrow$) & ($\downarrow$) \\
\hline
\multirow{3}{*}{50} & ShortGPT & 3.14 & 55.87 & 39.34 & 67.98 & 54.14 & 80.21 & 15.20\\
 & 2SSP & 3.11 & 50.10 & 33.86 & 58.74 & 52.44 & 72.92 & 16.30\\
 &  \pruner & 2.72 & 55.93 & 41.36 & 69.31 & 53.70 & \textbf{82.64} & \textbf{12.20}\\
 \cdashline{1-9}
 \multirow{3}{*}{250} & ShortGPT & 3.10 & 58.02 & 39.95 & 67.66 & 53.90 & 80.98 & 13.30\\
 & 2SSP & 3.02 & 51.68 & 33.62 & 63.59 & 53.30 & 75.31 & 15.10\\
 & \pruner & 2.76 & 54.34 & 40.21 & 68.25 & 53.58 & \textbf{81.26} & \textbf{12.40}\\
 \cdashline{1-9}
 \multirow{3}{*}{1000} & ShortGPT & 3.10 & 58.60 & 39.96 & 67.86 & 54.09 & 81.21 & 14.00\\
 & 2SSP & 3.00 & 52.98 & 33.90 & 60.05 & 52.20 & 74.26 & 15.40\\
 &  \pruner & 2.75 & 53.79 & 40.08 & 69.30 & 53.93 & \textbf{81.50} & \textbf{12.90}\\
\hline
\end{tabular}%
}
\caption{Calibration data sample size with Slim Orca~\citep{SlimOrca}.}
\label{subtable:calibration-size-results}
\end{subtable}
\hfill
\begin{subtable}[t]{\textwidth}
\resizebox{\textwidth}{!}{%
\begin{tabular}{ccccccc|cc}
\hline
\textbf{Calibration Dataset} & \textbf{Method} & \textbf{Generative} & \textbf{World} & \textbf{Domain} & \textbf{NLU \& NLI} & \textbf{Safety} & \textbf{Avg RP (\%)} & \textbf{Std RP (\%)} \\
 & & (Log-PPL $\downarrow$) & (Acc $\uparrow$) & (Acc $\uparrow$) & (Acc $\uparrow$) & (Score $\uparrow$) & ($\uparrow$) & ($\downarrow$) \\
\hline
\multirow{2}{*}{Slim Orca} & ShortGPT & 3.14 & 55.87 & 39.34 & 67.98 & 54.14 & 80.21 & 15.20\\
\multirow{2}{*}{~\citep{SlimOrca}} & 2SSP & 3.11 & 50.10 & 33.86 & 58.74 & 52.44 & 72.92 & 16.30\\
 &  \pruner & 2.72 & 55.93 & 41.36 & 69.31 & 53.70 & \textbf{82.64} & \textbf{12.20} \\
 \cdashline{1-9}
 \multirow{2}{*}{Alpaca} & ShortGPT & 2.93 & 58.48 & 41.19 & 68.64 & 53.73 & \textbf{82.59} & 13.50\\ 
 \multirow{2}{*}{~\citep{alpaca}} & 2SSP & 2.71 & 53.44 & 39.26 & 63.75 & 52.84 & 79.06 & 14.30\\
 &  \pruner & 2.75 & 54.80 & 40.49 & 68.67 & 53.79 & 81.79 & \textbf{13.10}\\
 \cdashline{1-9}
 \multirow{2}{*}{C4} & ShortGPT & 3.65 & 61.19 & 39.28 & 66.70 & 54.47 & 80.63 & 15.30\\
 \multirow{2}{*}{~\citep{c4-dataset}} & 2SSP & 3.35 & 39.65 & 31.05 & 55.10 & 52.19 & 67.48 & 18.80\\
 & \pruner & 2.75 & 54.77 & 40.66 & 68.61 & 54.11 & \textbf{81.93} & \textbf{13.10}\\
\hline
\end{tabular}%
}
\caption{Impact of calibration dataset with sample size of 50.}
\label{subtable:calibration-distribution-results}
\end{subtable}
\caption{Impact of calibration on compressed Qwen3-8B with RFT. Detailed results provided in Tables~\ref{tab:calibration-size-all-results} and~\ref{tab:calibration-distribution-all-results} of Appendix \ref{appendix:calibration-results}.}
\label{tab:calibration-dataset-results}
\end{table*}
\subsection{Impact of calibration on pruning robustness.}
\label{app:calibration-impact}
We analyze the impact of calibration data on structured pruning by varying both \emph{calibration set size} and \emph{calibration distribution} using Qwen3-8B with post-pruning RFT (Table~\ref{tab:calibration-dataset-results}). As the number of calibration samples increases from 50 to 1000, 2SSP demonstrates a marked increase in its Avg RP ($72.92\% \rightarrow 75.31$), followed by a decline in the same ($75.31\% \rightarrow 74.26\%$), while its Std RP exhibits the reverse pattern ($16.30\% \rightarrow 15.10\% \rightarrow 15.40\%$), indicating its relative instability and sensitivity to calibration sample count. While ShortGPT remains comparatively stable in average performance, it consistently suffers from higher variance across task groups. In contrast, \pruner\ achieves the best trade-off between performance and consistency across all calibration sizes, reducing task-wise standard deviation by an average of 11.45\% relative to ShortGPT. A more noticeable trend is observed when varying calibration distributions: both ShortGPT and 2SSP show large performance fluctuations across Slim Orca~\citep{SlimOrca}, Alpaca~\citep{alpaca}, and C4~\citep{c4-dataset}, whereas \pruner\ maintains consistently high Avg RP and low Std RP. These results indicate that \pruner's pruning strategy is significantly less sensitive to calibration choices, further reinforcing its robustness with respect to its baselines. Detailed results are provided in Appendix~\ref{appendix:calibration-results}.

\begin{table*}[!htb]
\centering
\resizebox{\textwidth}{!}{%
\begin{tabular}{lcccccc|cc}
\hline
\textbf{CR}& \textbf{Transfer regime} & \textbf{Generative} & \textbf{World} & \textbf{Domain} & \textbf{NLU \& NLI} & \textbf{Safety} & \textbf{Avg RP (\%)} & \textbf{Std RP (\%)} \\
 & & (Log-PPL $\downarrow$) & (Acc $\uparrow$) & (Acc $\uparrow$) & (Acc $\uparrow$) & (Score $\uparrow$) & ($\uparrow$) & ($\downarrow$) \\
\hline
\multirow{2}{*}{25\%} & $25\% \rightarrow 25\%$ (No transfer) & 3.65 & 50.81 & 36.21 & 61.04 & 54.90 & {74.41} & {15.00}\\
& $35\% \rightarrow 25\%$ (High-to-low) & 3.66 & 50.59 & 36.12 & 60.78 & 54.89 & 74.24 & 17.00 \\
& $50\% \rightarrow 25\%$ (High-to-low) & 3.63 & 49.48 & 35.73 & 61.58 & 54.64 & 73.92 & 17.40\\
\cdashline{1-9}
\multirow{2}{*}{35\%} & $35\% \rightarrow 35\%$ (No transfer) & 5.11 & 44.04 & 30.43 & 48.68 & 52.76 & {63.71} & {20.10}\\
& $25\% \rightarrow 35\%$ (Low-to-high) & 5.07 & 42.73 & 29.29 & 49.72 & 53.75 & 63.48 & 22.90 \\
& $50\% \rightarrow 35\%$ (High-to-low) & 5.08 & 44.42 & 30.43 & 47.44 & 54.15 & 63.83 & 22.90\\
\cdashline{1-9}
\multirow{2}{*}{50\%} & $50\% \rightarrow 50\%$ (No transfer) & 11.56 & 36.37 & 26.80 & 33.63 & 53.51 & {53.30} & {28.60}\\
& $25\% \rightarrow 50\%$ (Low-to-high) & 9.55 & 35.51 & 25.86 & 30.69 & 51.97 & 51.46 & 30.00\\
& $35\% \rightarrow 50\%$ (Low-to-high) & 11.46 & 36.80 & 26.23 & 31.95 & 53.71 & 52.45 & 30.90\\
\hline
\end{tabular}%
}
\caption{Transferability of importance scores across compression ratios. We reuse importance scores computed at one compression ratio to prune the model at a different target ratio (e.g., 35\% $\rightarrow$ 25\%), and compare against recomputing scores from scratch (Table~\ref{tab:main_results}). 
Table~\ref{tab:transferability-detailed-results} of Appendix~\ref{appendix:transferability} displays transferability metrics across each task.}
\label{tab:transferability-results}
\end{table*}
\subsection{Transferability of importance scores.}
\label{app:transferability}
Table~\ref{tab:transferability-results} evaluates the robustness of \pruner's importance scores when transferred across compression ratios. Although \pruner\ is designed to recompute scores for each target budget, transferring them across ratios (35\% $\rightarrow$ 25\% and vice versa) results in only marginal degradation in average retention performance with a modest increase in task-wise variance. For example, the 35\% $\rightarrow$ 25\% transfer preserves nearly identical average RP (74.24\% vs.\ 74.41\%), with the reverse exhibiting a similarly small drop (63.48\% vs.\ 63.71\%). This stability indicates that \pruner's importance signals reflect intrinsic structural salience rather than budget-specific artifacts, while the slight variance increase highlights the benefit of recomputing scores for strict optimality. Overall, transferability offers a favorable efficiency--robustness trade-off, allowing users to bypass \pruner's most time-intensive step, i.e, importance computation, with minimal performance sacrifice. We provide task-wise results in Appendix~\ref{appendix:transferability}.

\subsection{Interpretable structural salience across layers.}
\label{app:interpretable-structural-salience}
\begin{figure*}[t]
    \centering
    \includegraphics[width=\linewidth]{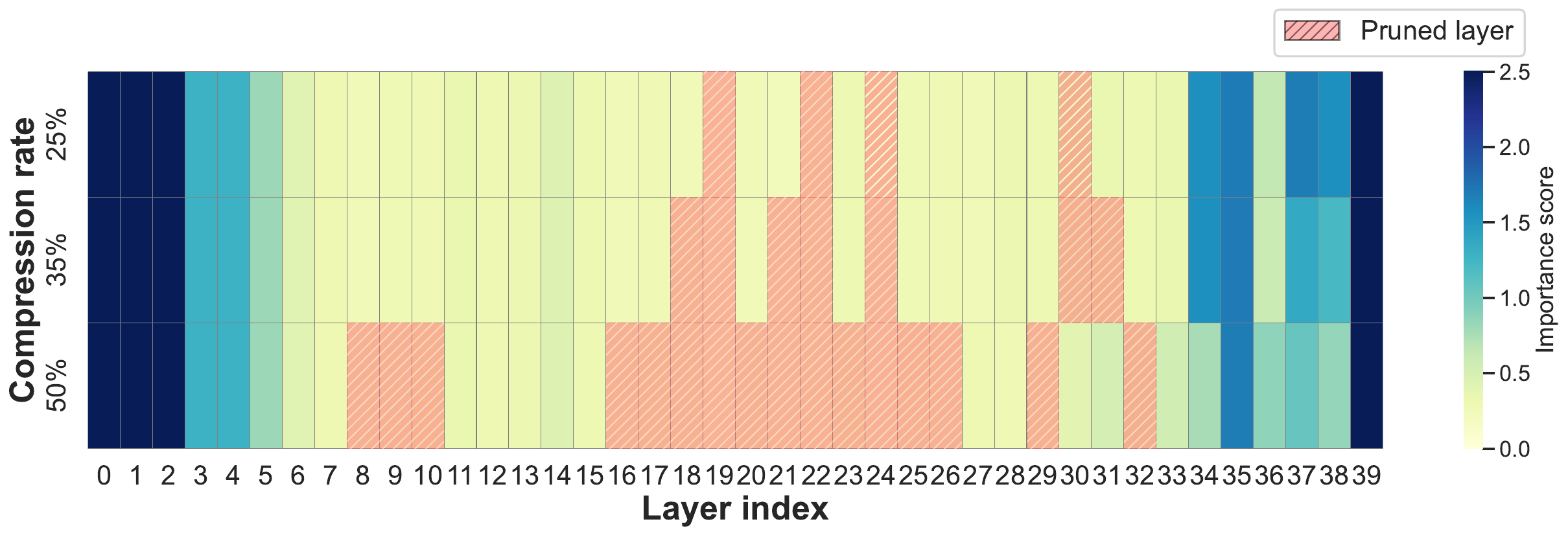}
    \caption{Layer-wise importance scores and pruning decisions for Phi-4-14B across multiple compression ratios. 
    The substantial overlap in pruned layers across compression ratios indicates stable, interpretable importance estimates that reflect intrinsic architectural redundancy rather than budget-specific noise.}
    \label{fig:consistency}
\end{figure*}
Figure~\ref{fig:consistency} highlights an important qualitative property of \pruner: beyond improved performance, it yields an interpretable measure of layer-wise importance. The learned scores exhibit a clear structure --- consistently high importance assigned to early and late transformer layers and systematically lower importance in intermediate layers --- with pruned layers predominantly drawn from the middle while boundary layers are preserved. Moreover, substantial overlap in evicted layers across compression budgets indicates that importance estimates capture stable architectural salience rather than budget-specific noise. This structured pattern aligns with known functional roles of transformer layers, suggesting that \pruner\ provides a principled, human-interpretable signal for understanding redundancy and capacity allocation in LLMs.

\begin{figure*}[!htb]
    \centering
    \begin{subfigure}[t]{0.45\linewidth}
        \centering
    \includegraphics[width=\linewidth]{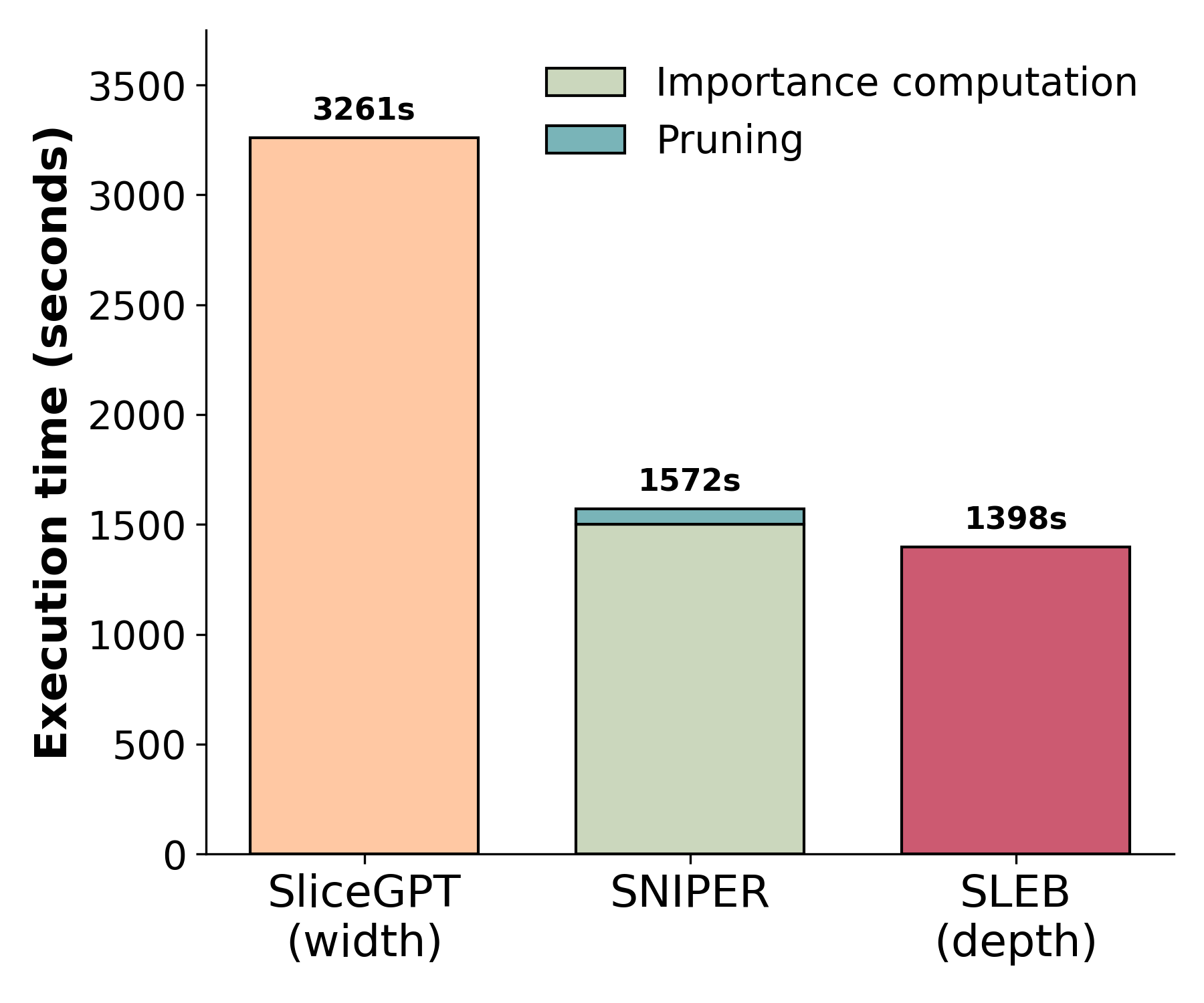}
        \caption{Comparison of pruning time.}
        \label{fig:runtime}
    \end{subfigure}
    \begin{subfigure}[t]{0.5\linewidth}
        \centering
        \includegraphics[width=\linewidth]{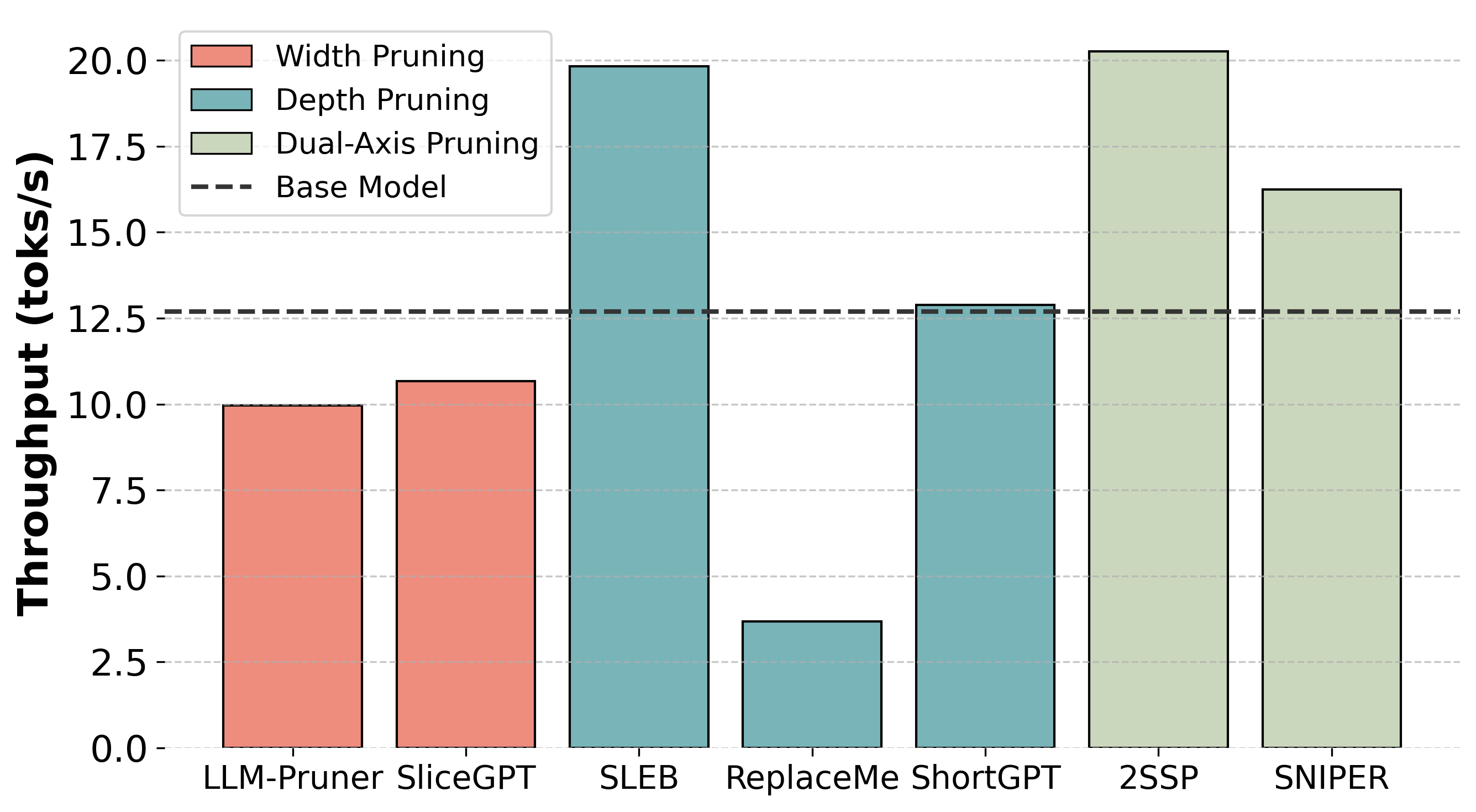}
        \caption{Post-pruning inference speed comparison.}
        \label{fig:latency-eval}
    \end{subfigure}
    
    \caption{Pruning runtime (a) and post-pruning inference speed (b) for different structured pruning method. All benchmarking conducted on Qwen3-8B model at 25\% compression.}
    \label{fig:latency-runtime}
\end{figure*}
\subsection{Runtime analysis, latency, and practical utility}
Figure~\ref{fig:runtime} compares the end-to-end runtime of representative depth-, width-, and mixed-granularity pruning methods on Qwen3-8B. Greedy depth pruners such as SLEB are the fastest ($\approx$ 23 minutes) as importance estimation is performed over a small number of coarse atomic units, whereas width-based methods such as SliceGPT incur substantially higher runtime ($\approx$ 55 minutes) due to computations over large sets of fine-grained elements. \pruner\ incurs a total runtime that is only about 4 minutes higher than depth-only methods, with approximately 96\% of runtime devoted to importance estimation while the actual pruning step requires only 56 seconds. This separation enables amortization of the dominant cost across multiple pruning runs by reusing importance scores (Appendix~\ref{app:transferability}), while requiring GPU acceleration only during importance estimation. We further analyze post-pruning inference speedup at 25\% compression in Figure~\ref{fig:latency-eval}. Width pruners fall below the base model due to hardware-unfriendly irregular tensor shapes whereas pure depth pruners achieve the largest raw speedups by eliminating entire compute blocks. Furthermore, both dual-axis approaches yield notable latency improvements by offsetting irregular tensor penalties through component elimination. We note that while \pruner\ lags behind 2SSP in realized efficiency gains (16.24 vs. 20.26 toks/s), it compensates with consistently strong performance and robustness across architectures and pruning regimes. We hypothesize that 2SSP's latency gains stem from its treatment of attention blocks as the sole coarse-grained component, whereas \pruner\ also considers MLP blocks at this granularity, trading some throughput for more balanced structural coverage. 

\begin{table*}[!htb]
\centering
\begin{tabular}{lcc}
\hline
\textbf{$\alpha$} & \textbf{CRAFT} & \textbf{Run-time of DP Solver (in seconds)} \\
\hline
8 & 0.98 & 22.18\\
32 & 0.98 & 5.84\\
128 & 0.98 & 1.46\\
512 & 0.98 & 0.28\\
8,192 & 0.98 & 0.02\\
32,768 & 0.98 & 0.01\\
100,000 & 1.79 (over-pruning) & <0.01\\
\hline
\end{tabular}%
\caption{Quantitative analysis of the effect of different values of the discretizing factor, $\alpha$, on CRAFT and algorithm runtime. These results were obtained on Qwen3-8B.}
\label{tab:discretization-analysis}
\end{table*}
\subsection{Analysis of Different Values of Discretizing Factor}
\label{app:discretizing-factor-sweep}
As demonstrated in Table~\ref{tab:discretization-analysis}, \pruner\ is robust to a large range of values for $\alpha$ and maintains a near-perfect CRAFT of 0.98 for all values from 8 to 32768. As expected, the runtime decreases as $\alpha$ increases. However, as $\alpha$ reaches exceptionally high values such as 100,000, it causes large components to be discretized into the same bucket as far smaller components, leading to a notable over-shooting of the compression budget. Therefore, to ensure $\alpha$ generalizes well across architectures while keeping runtime low, we select a conservative value of $\alpha=32$. Notably, for all $\alpha$ in the range 8–32768, the resulting pruned models are identical and therefore yield the same performance.

\section{Conclusion}
We presented \pruner, a dual-axis structured pruning framework that compresses along the model's depth via a knapsack-based objective, guaranteeing a conditionally optimal selection of coarse-grained components and follows that with a width-pruning stage. Through its two stage operation, \pruner\ achieves precise budget adherence, strong performance retention, and reduced task-level variance across diverse model architectures. Extensive experiments demonstrate consistent improvements over state-of-the-art structured pruners 
while performing competitively even under aggressive compression. Beyond performance gains, \pruner\ offers a practical deployment advantage by enabling amortized pruning through reusable importance estimates under strict budget constraints. Together, these results underscore the value of non-greedy optimization as a principled foundation for reliable and deployable LLM pruning.

\section{Limitations}
\label{sec:limitations}
\pruner\ currently employs homogeneous pruning signals across all model components. Incorporating non-homogeneous, expert-specific signals for Mixture-of-Experts (MoE) 
architectures represents a natural extension that could further enhance pruning efficacy 
in such settings. Additionally, while \pruner\ provides conditional optimality guarantees 
during component removal, extending theoretical analogous guarantees to other stages of 
the pipeline such as importance estimation, remains a compelling direction for future work.
\section{Ethical Considerations}
This work presents \pruner, a dual-axis structured pruning framework for compressing LLMs. We reflect on the ethical implications of our research below.
\paragraph{Utilization of Public Artifacts.}
In this work, we employ several publicly available models such as Qwen3-8B \citep{qwen3technicalreport}, LLaMA-3.1-8B-Instruct \citep{grattafiori2024llama3herdmodels}, Phi-4 \citep{abdin2024phi4technicalreport}, and GPT-OSS-20B \citep{openai2025gptoss120bgptoss20bmodel}. We also use the publicly available Slim Orca \citep{SlimOrca} dataset for pre-compression calibration and post-compression RFT. Furthermore, we utilize the Alpaca \citep{alpaca} and C4 \citep{c4-dataset} datasets to test the effect of calibration data distribution on the pruning decisions made by each method. We strictly adhere to the terms of use of each of the artifacts used by us. 
\paragraph{Intended Use.} 
\pruner\ is designed to
prune LLMs and make them more accessible to practitioners working in resource-constrained environments. However, we recognize that increasing accessibility also increases the likelihood of a pruned model being misused. Therefore, we implore users of our framework to use compressed models by adhering to the usage policies of the respective base models. 
\paragraph{Environmental Impact.}
As LLMs continue to grow in size, the resources required to run them have reached a concerning point, and will likely continue to grow. However, through model compression, we primarily aim to limit the amount of such required resources, including electricity for powering machines that run LLMs and water for cooling them. We believe that this work will have a positive effect on the environment by making it possible to run more economical variants of resource-intensive LLMs.
\section*{Acknowledgment}
 T. Chakraborty acknowledges the support of the Microsoft Research India Research Grant and the Rajiv Khemani Young Faculty Chair Professorship in Artificial Intelligence.


\bibliographystyle{IEEEtranN}
\small
\bibliography{custom}

@article{slidingwindowmerging,
   title={Sliding-Window Merging for Compacting Patch-Redundant Layers in LLMs},
   volume={40},
   ISSN={2159-5399},
   url={http://dx.doi.org/10.1609/aaai.v40i25.39222},
   DOI={10.1609/aaai.v40i25.39222},
   number={25},
   journal={Proceedings of the AAAI Conference on Artificial Intelligence},
   publisher={Association for the Advancement of Artificial Intelligence (AAAI)},
   author={Ding, Xuan and Sun, Rui and Zhang, Yunjian and Yan, Xiu and Zhou, Yueqi and Huang, Kaihao and Fu, Suzhong and Aviles-Rivero, Angelica I and Xie, Chuanlong and Zhu, Yao},
   year={2026},
   month=Mar, pages={20826–20834} }

@misc{kim2024shortenedllamadepthpruning,
      title={Shortened LLaMA: Depth Pruning for Large Language Models with Comparison of Retraining Methods}, 
      author={Bo-Kyeong Kim and Geonmin Kim and Tae-Ho Kim and Thibault Castells and Shinkook Choi and Junho Shin and Hyoung-Kyu Song},
      year={2024},
      eprint={2402.02834},
      archivePrefix={arXiv},
      primaryClass={cs.LG},
      url={https://arxiv.org/abs/2402.02834}, 
}

@article{sandri2025ssp,
title={2{SSP}: A Two-Stage Framework for Structured Pruning of {LLM}s},
author={Fabrizio Sandri and Elia Cunegatti and Giovanni Iacca},
journal={Transactions on Machine Learning Research},
issn={2835-8856},
year={2025},
url={https://openreview.net/forum?id=Qd7LzJBg21},
note={}
}

@misc{alpaca,
  author = {Rohan Taori and Ishaan Gulrajani and Tianyi Zhang and Yann Dubois and Xuechen Li and Carlos Guestrin and Percy Liang and Tatsunori B. Hashimoto },
  title = {Stanford Alpaca: An Instruction-following LLaMA model},
  year = {2023},
  publisher = {GitHub},
  journal = {GitHub repository},
  howpublished = {\url{https://github.com/tatsu-lab/stanford_alpaca}},
}

@article{c4-dataset,
author = {Raffel, Colin and Shazeer, Noam and Roberts, Adam and Lee, Katherine and Narang, Sharan and Matena, Michael and Zhou, Yanqi and Li, Wei and Liu, Peter J.},
title = {Exploring the limits of transfer learning with a unified text-to-text transformer},
year = {2020},
issue_date = {January 2020},
publisher = {JMLR.org},
volume = {21},
number = {1},
issn = {1532-4435},
journal = {J. Mach. Learn. Res.},
month = jan,
articleno = {140},
numpages = {67}
}

@inproceedings{
hu2022lora,
title={Lo{RA}: Low-Rank Adaptation of Large Language Models},
author={Edward J Hu and Yelong Shen and Phillip Wallis and Zeyuan Allen-Zhu and Yuanzhi Li and Shean Wang and Lu Wang and Weizhu Chen},
booktitle={International Conference on Learning Representations},
year={2022},
url={https://openreview.net/forum?id=nZeVKeeFYf9}
}

@inproceedings{wolf-etal-2020-transformers,
    title = "Transformers: State-of-the-Art Natural Language Processing",
    author = "Wolf, Thomas  and
      Debut, Lysandre  and
      Sanh, Victor  and
      Chaumond, Julien  and
      Delangue, Clement  and
      Moi, Anthony  and
      Cistac, Pierric  and
      Rault, Tim  and
      Louf, Remi  and
      Funtowicz, Morgan  and
      Davison, Joe  and
      Shleifer, Sam  and
      von Platen, Patrick  and
      Ma, Clara  and
      Jernite, Yacine  and
      Plu, Julien  and
      Xu, Canwen  and
      Le Scao, Teven  and
      Gugger, Sylvain  and
      Drame, Mariama  and
      Lhoest, Quentin  and
      Rush, Alexander",
    editor = "Liu, Qun  and
      Schlangen, David",
    booktitle = "Proceedings of the 2020 Conference on Empirical Methods in Natural Language Processing: System Demonstrations",
    month = oct,
    year = "2020",
    address = "Online",
    publisher = "Association for Computational Linguistics",
    url = "https://aclanthology.org/2020.emnlp-demos.6/",
    doi = "10.18653/v1/2020.emnlp-demos.6",
    pages = "38--45"
}

@inproceedings{paszke-pytorch,
 author = {Paszke, Adam and Gross, Sam and Massa, Francisco and Lerer, Adam and Bradbury, James and Chanan, Gregory and Killeen, Trevor and Lin, Zeming and Gimelshein, Natalia and Antiga, Luca and Desmaison, Alban and Kopf, Andreas and Yang, Edward and DeVito, Zachary and Raison, Martin and Tejani, Alykhan and Chilamkurthy, Sasank and Steiner, Benoit and Fang, Lu and Bai, Junjie and Chintala, Soumith},
 booktitle = {Advances in Neural Information Processing Systems},
 editor = {H. Wallach and H. Larochelle and A. Beygelzimer and F. d\textquotesingle Alch\'{e}-Buc and E. Fox and R. Garnett},
 pages = {},
 publisher = {Curran Associates, Inc.},
 title = {PyTorch: An Imperative Style, High-Performance Deep Learning Library},
 url = {https://proceedings.neurips.cc/paper_files/paper/2019/file/bdbca288fee7f92f2bfa9f7012727740-Paper.pdf},
 volume = {32},
 year = {2019}
}

@inproceedings{clark2019boolq,
    title = "{B}ool{Q}: Exploring the Surprising Difficulty of Natural Yes/No Questions",
    author = "Clark, Christopher  and
      Lee, Kenton  and
      Chang, Ming-Wei  and
      Kwiatkowski, Tom  and
      Collins, Michael  and
      Toutanova, Kristina",
    editor = "Burstein, Jill  and
      Doran, Christy  and
      Solorio, Thamar",
    booktitle = "Proceedings of the 2019 Conference of the No rth {A}merican Chapter of the Association for Computational Linguistics: Human Language Technologies, Volume 1 (Long and Short Papers)",
    month = jun,
    year = "2019",
    address = "Minneapolis, Minnesota",
    publisher = "Association for Computational Linguistics",
    url = "https://aclanthology.org/N19-1300/",
    doi = "10.18653/v1/N19-1300",
    pages = "2924--2936"
}

@article{warstadt2020blimp,
    author = {Warstadt, Alex and Parrish, Alicia and Liu, Haokun and Mohananey, Anhad and Peng, Wei and Wang, Sheng-Fu and Bowman, Samuel R.},
    title = {BLiMP: The Benchmark of Linguistic Minimal Pairs for English},
    journal = {Transactions of the Association for Computational Linguistics},
    volume = {8},
    number = {},
    pages = {377-392},
    year = {2020},
    doi = {10.1162/tacl\_a\_00321},
    URL = {https://doi.org/10.1162/tacl_a_00321},
    eprint = {https://doi.org/10.1162/tacl_a_00321}
}

@inproceedings{talmor-etal-2019-commonsenseqa,
    title = "{C}ommonsense{QA}: A Question Answering Challenge Targeting Commonsense Knowledge",
    author = "Talmor, Alon  and
      Herzig, Jonathan  and
      Lourie, Nicholas  and
      Berant, Jonathan",
    booktitle = "Proceedings of the 2019 Conference of the North {A}merican Chapter of the Association for Computational Linguistics: Human Language Technologies, Volume 1 (Long and Short Papers)",
    month = jun,
    year = "2019",
    address = "Minneapolis, Minnesota",
    publisher = "Association for Computational Linguistics",
    url = "https://aclanthology.org/N19-1421",
    doi = "10.18653/v1/N19-1421",
    pages = "4149--4158",
    archivePrefix = "arXiv",
    eprint        = "1811.00937",
    primaryClass  = "cs",
}

@inproceedings{aroca-ouellette-etal-2021-prost,
    title = "{PROST}: {P}hysical Reasoning about Objects through Space and Time",
    author = "Aroca-Ouellette, St{\'e}phane  and
      Paik, Cory  and
      Roncone, Alessandro  and
      Kann, Katharina",
    editor = "Zong, Chengqing  and
      Xia, Fei  and
      Li, Wenjie  and
      Navigli, Roberto",
    booktitle = "Findings of the Association for Computational Linguistics: ACL-IJCNLP 2021",
    month = aug,
    year = "2021",
    address = "Online",
    publisher = "Association for Computational Linguistics",
    url = "https://aclanthology.org/2021.findings-acl.404/",
    doi = "10.18653/v1/2021.findings-acl.404",
    pages = "4597--4608"
}

@article{jin2021disease,
AUTHOR = {Jin, Di and Pan, Eileen and Oufattole, Nassim and Weng, Wei-Hung and Fang, Hanyi and Szolovits, Peter},
TITLE = {What Disease Does This Patient Have? A Large-Scale Open Domain Question Answering Dataset from Medical Exams},
JOURNAL = {Applied Sciences},
VOLUME = {11},
YEAR = {2021},
NUMBER = {14},
ARTICLE-NUMBER = {6421},
URL = {https://www.mdpi.com/2076-3417/11/14/6421},
ISSN = {2076-3417},
DOI = {10.3390/app11146421}
}

@misc{clark2018thinksolvedquestionanswering,
      title={Think you have Solved Question Answering? Try ARC, the AI2 Reasoning Challenge}, 
      author={Peter Clark and Isaac Cowhey and Oren Etzioni and Tushar Khot and Ashish Sabharwal and Carissa Schoenick and Oyvind Tafjord},
      year={2018},
      eprint={1803.05457},
      archivePrefix={arXiv},
      primaryClass={cs.AI},
      url={https://arxiv.org/abs/1803.05457}, 
}

@inproceedings{amini-etal-2019-mathqa,
    title = "{M}ath{QA}: Towards Interpretable Math Word Problem Solving with Operation-Based Formalisms",
    author = "Amini, Aida  and
      Gabriel, Saadia  and
      Lin, Shanchuan  and
      Koncel-Kedziorski, Rik  and
      Choi, Yejin  and
      Hajishirzi, Hannaneh",
    booktitle = "Proceedings of the 2019 Conference of the North {A}merican Chapter of the Association for Computational Linguistics: Human Language Technologies, Volume 1 (Long and Short Papers)",
    month = jun,
    year = "2019",
    address = "Minneapolis, Minnesota",
    publisher = "Association for Computational Linguistics",
    url = "https://aclanthology.org/N19-1245",
    doi = "10.18653/v1/N19-1245",
    pages = "2357--2367",
}

@misc{openai2025gptoss120bgptoss20bmodel,
      title={gpt-oss-120b \& gpt-oss-20b Model Card}, 
      author={OpenAI and : and Sandhini Agarwal and Lama Ahmad and Jason Ai and Sam Altman and Andy Applebaum and Edwin Arbus and Rahul K. Arora and Yu Bai and Bowen Baker and Haiming Bao and Boaz Barak and Ally Bennett and Tyler Bertao and Nivedita Brett and Eugene Brevdo and Greg Brockman and Sebastien Bubeck and Che Chang and Kai Chen and Mark Chen and Enoch Cheung and Aidan Clark and Dan Cook and Marat Dukhan and Casey Dvorak and Kevin Fives and Vlad Fomenko and Timur Garipov and Kristian Georgiev and Mia Glaese and Tarun Gogineni and Adam Goucher and Lukas Gross and Katia Gil Guzman and John Hallman and Jackie Hehir and Johannes Heidecke and Alec Helyar and Haitang Hu and Romain Huet and Jacob Huh and Saachi Jain and Zach Johnson and Chris Koch and Irina Kofman and Dominik Kundel and Jason Kwon and Volodymyr Kyrylov and Elaine Ya Le and Guillaume Leclerc and James Park Lennon and Scott Lessans and Mario Lezcano-Casado and Yuanzhi Li and Zhuohan Li and Ji Lin and Jordan Liss and Lily and Liu and Jiancheng Liu and Kevin Lu and Chris Lu and Zoran Martinovic and Lindsay McCallum and Josh McGrath and Scott McKinney and Aidan McLaughlin and Song Mei and Steve Mostovoy and Tong Mu and Gideon Myles and Alexander Neitz and Alex Nichol and Jakub Pachocki and Alex Paino and Dana Palmie and Ashley Pantuliano and Giambattista Parascandolo and Jongsoo Park and Leher Pathak and Carolina Paz and Ludovic Peran and Dmitry Pimenov and Michelle Pokrass and Elizabeth Proehl and Huida Qiu and Gaby Raila and Filippo Raso and Hongyu Ren and Kimmy Richardson and David Robinson and Bob Rotsted and Hadi Salman and Suvansh Sanjeev and Max Schwarzer and D. Sculley and Harshit Sikchi and Kendal Simon and Karan Singhal and Yang Song and Dane Stuckey and Zhiqing Sun and Philippe Tillet and Sam Toizer and Foivos Tsimpourlas and Nikhil Vyas and Eric Wallace and Xin Wang and Miles Wang and Olivia Watkins and Kevin Weil and Amy Wendling and Kevin Whinnery and Cedric Whitney and Hannah Wong and Lin Yang and Yu Yang and Michihiro Yasunaga and Kristen Ying and Wojciech Zaremba and Wenting Zhan and Cyril Zhang and Brian Zhang and Eddie Zhang and Shengjia Zhao},
      year={2025},
      eprint={2508.10925},
      archivePrefix={arXiv},
      primaryClass={cs.CL},
      url={https://arxiv.org/abs/2508.10925}, 
}

@misc{abdin2024phi4technicalreport,
      title={Phi-4 Technical Report}, 
      author={Marah Abdin and Jyoti Aneja and Harkirat Behl and Sébastien Bubeck and Ronen Eldan and Suriya Gunasekar and Michael Harrison and Russell J. Hewett and Mojan Javaheripi and Piero Kauffmann and James R. Lee and Yin Tat Lee and Yuanzhi Li and Weishung Liu and Caio C. T. Mendes and Anh Nguyen and Eric Price and Gustavo de Rosa and Olli Saarikivi and Adil Salim and Shital Shah and Xin Wang and Rachel Ward and Yue Wu and Dingli Yu and Cyril Zhang and Yi Zhang},
      year={2024},
      eprint={2412.08905},
      archivePrefix={arXiv},
      primaryClass={cs.CL},
      url={https://arxiv.org/abs/2412.08905}, 
}

@misc{grattafiori2024llama3herdmodels,
      title={The Llama 3 Herd of Models}, 
      author={Aaron Grattafiori and Abhimanyu Dubey and Abhinav Jauhri and Abhinav Pandey and Abhishek Kadian and Ahmad Al-Dahle and Aiesha Letman and Akhil Mathur and Alan Schelten and Alex Vaughan and Amy Yang and Angela Fan and Anirudh Goyal and Anthony Hartshorn and Aobo Yang and Archi Mitra and Archie Sravankumar and Artem Korenev and Arthur Hinsvark and Arun Rao and Aston Zhang and Aurelien Rodriguez and Austen Gregerson and Ava Spataru and Baptiste Roziere and Bethany Biron and Binh Tang and Bobbie Chern and Charlotte Caucheteux and Chaya Nayak and Chloe Bi and Chris Marra and Chris McConnell and Christian Keller and Christophe Touret and Chunyang Wu and Corinne Wong and Cristian Canton Ferrer and Cyrus Nikolaidis and Damien Allonsius and Daniel Song and Danielle Pintz and Danny Livshits and Danny Wyatt and David Esiobu and Dhruv Choudhary and Dhruv Mahajan and Diego Garcia-Olano and Diego Perino and Dieuwke Hupkes and Egor Lakomkin and Ehab AlBadawy and Elina Lobanova and Emily Dinan and Eric Michael Smith and Filip Radenovic and Francisco Guzmán and Frank Zhang and Gabriel Synnaeve and Gabrielle Lee and Georgia Lewis Anderson and Govind Thattai and Graeme Nail and Gregoire Mialon and Guan Pang and Guillem Cucurell and Hailey Nguyen and Hannah Korevaar and Hu Xu and Hugo Touvron and Iliyan Zarov and Imanol Arrieta Ibarra and Isabel Kloumann and Ishan Misra and Ivan Evtimov and Jack Zhang and Jade Copet and Jaewon Lee and Jan Geffert and Jana Vranes and Jason Park and Jay Mahadeokar and Jeet Shah and Jelmer van der Linde and Jennifer Billock and Jenny Hong and Jenya Lee and Jeremy Fu and Jianfeng Chi and Jianyu Huang and Jiawen Liu and Jie Wang and Jiecao Yu and Joanna Bitton and Joe Spisak and Jongsoo Park and Joseph Rocca and Joshua Johnstun and Joshua Saxe and Junteng Jia and Kalyan Vasuden Alwala and Karthik Prasad and Kartikeya Upasani and Kate Plawiak and Ke Li and Kenneth Heafield and Kevin Stone and Khalid El-Arini and Krithika Iyer and Kshitiz Malik and Kuenley Chiu and Kunal Bhalla and Kushal Lakhotia and Lauren Rantala-Yeary and Laurens van der Maaten and Lawrence Chen and Liang Tan and Liz Jenkins and Louis Martin and Lovish Madaan and Lubo Malo and Lukas Blecher and Lukas Landzaat and Luke de Oliveira and Madeline Muzzi and Mahesh Pasupuleti and Mannat Singh and Manohar Paluri and Marcin Kardas and Maria Tsimpoukelli and Mathew Oldham and Mathieu Rita and Maya Pavlova and Melanie Kambadur and Mike Lewis and Min Si and Mitesh Kumar Singh and Mona Hassan and Naman Goyal and Narjes Torabi and Nikolay Bashlykov and Nikolay Bogoychev and Niladri Chatterji and Ning Zhang and Olivier Duchenne and Onur Çelebi and Patrick Alrassy and Pengchuan Zhang and Pengwei Li and Petar Vasic and Peter Weng and Prajjwal Bhargava and Pratik Dubal and Praveen Krishnan and Punit Singh Koura and Puxin Xu and Qing He and Qingxiao Dong and Ragavan Srinivasan and Raj Ganapathy and Ramon Calderer and Ricardo Silveira Cabral and Robert Stojnic and Roberta Raileanu and Rohan Maheswari and Rohit Girdhar and Rohit Patel and Romain Sauvestre and Ronnie Polidoro and Roshan Sumbaly and Ross Taylor and Ruan Silva and Rui Hou and Rui Wang and Saghar Hosseini and Sahana Chennabasappa and Sanjay Singh and Sean Bell and Seohyun Sonia Kim and Sergey Edunov and Shaoliang Nie and Sharan Narang and Sharath Raparthy and Sheng Shen and Shengye Wan and Shruti Bhosale and Shun Zhang and Simon Vandenhende and Soumya Batra and Spencer Whitman and Sten Sootla and Stephane Collot and Suchin Gururangan and Sydney Borodinsky and Tamar Herman and Tara Fowler and Tarek Sheasha and Thomas Georgiou and Thomas Scialom and Tobias Speckbacher and Todor Mihaylov and Tong Xiao and Ujjwal Karn and Vedanuj Goswami and Vibhor Gupta and Vignesh Ramanathan and Viktor Kerkez and Vincent Gonguet and Virginie Do and Vish Vogeti and Vítor Albiero and Vladan Petrovic and Weiwei Chu and Wenhan Xiong and Wenyin Fu and Whitney Meers and Xavier Martinet and Xiaodong Wang and Xiaofang Wang and Xiaoqing Ellen Tan and Xide Xia and Xinfeng Xie and Xuchao Jia and Xuewei Wang and Yaelle Goldschlag and Yashesh Gaur and Yasmine Babaei and Yi Wen and Yiwen Song and Yuchen Zhang and Yue Li and Yuning Mao and Zacharie Delpierre Coudert and Zheng Yan and Zhengxing Chen and Zoe Papakipos and Aaditya Singh and Aayushi Srivastava and Abha Jain and Adam Kelsey and Adam Shajnfeld and Adithya Gangidi and Adolfo Victoria and Ahuva Goldstand and Ajay Menon and Ajay Sharma and Alex Boesenberg and Alexei Baevski and Allie Feinstein and Amanda Kallet and Amit Sangani and Amos Teo and Anam Yunus and Andrei Lupu and Andres Alvarado and Andrew Caples and Andrew Gu and Andrew Ho and Andrew Poulton and Andrew Ryan and Ankit Ramchandani and Annie Dong and Annie Franco and Anuj Goyal and Aparajita Saraf and Arkabandhu Chowdhury and Ashley Gabriel and Ashwin Bharambe and Assaf Eisenman and Azadeh Yazdan and Beau James and Ben Maurer and Benjamin Leonhardi and Bernie Huang and Beth Loyd and Beto De Paola and Bhargavi Paranjape and Bing Liu and Bo Wu and Boyu Ni and Braden Hancock and Bram Wasti and Brandon Spence and Brani Stojkovic and Brian Gamido and Britt Montalvo and Carl Parker and Carly Burton and Catalina Mejia and Ce Liu and Changhan Wang and Changkyu Kim and Chao Zhou and Chester Hu and Ching-Hsiang Chu and Chris Cai and Chris Tindal and Christoph Feichtenhofer and Cynthia Gao and Damon Civin and Dana Beaty and Daniel Kreymer and Daniel Li and David Adkins and David Xu and Davide Testuggine and Delia David and Devi Parikh and Diana Liskovich and Didem Foss and Dingkang Wang and Duc Le and Dustin Holland and Edward Dowling and Eissa Jamil and Elaine Montgomery and Eleonora Presani and Emily Hahn and Emily Wood and Eric-Tuan Le and Erik Brinkman and Esteban Arcaute and Evan Dunbar and Evan Smothers and Fei Sun and Felix Kreuk and Feng Tian and Filippos Kokkinos and Firat Ozgenel and Francesco Caggioni and Frank Kanayet and Frank Seide and Gabriela Medina Florez and Gabriella Schwarz and Gada Badeer and Georgia Swee and Gil Halpern and Grant Herman and Grigory Sizov and Guangyi and Zhang and Guna Lakshminarayanan and Hakan Inan and Hamid Shojanazeri and Han Zou and Hannah Wang and Hanwen Zha and Haroun Habeeb and Harrison Rudolph and Helen Suk and Henry Aspegren and Hunter Goldman and Hongyuan Zhan and Ibrahim Damlaj and Igor Molybog and Igor Tufanov and Ilias Leontiadis and Irina-Elena Veliche and Itai Gat and Jake Weissman and James Geboski and James Kohli and Janice Lam and Japhet Asher and Jean-Baptiste Gaya and Jeff Marcus and Jeff Tang and Jennifer Chan and Jenny Zhen and Jeremy Reizenstein and Jeremy Teboul and Jessica Zhong and Jian Jin and Jingyi Yang and Joe Cummings and Jon Carvill and Jon Shepard and Jonathan McPhie and Jonathan Torres and Josh Ginsburg and Junjie Wang and Kai Wu and Kam Hou U and Karan Saxena and Kartikay Khandelwal and Katayoun Zand and Kathy Matosich and Kaushik Veeraraghavan and Kelly Michelena and Keqian Li and Kiran Jagadeesh and Kun Huang and Kunal Chawla and Kyle Huang and Lailin Chen and Lakshya Garg and Lavender A and Leandro Silva and Lee Bell and Lei Zhang and Liangpeng Guo and Licheng Yu and Liron Moshkovich and Luca Wehrstedt and Madian Khabsa and Manav Avalani and Manish Bhatt and Martynas Mankus and Matan Hasson and Matthew Lennie and Matthias Reso and Maxim Groshev and Maxim Naumov and Maya Lathi and Meghan Keneally and Miao Liu and Michael L. Seltzer and Michal Valko and Michelle Restrepo and Mihir Patel and Mik Vyatskov and Mikayel Samvelyan and Mike Clark and Mike Macey and Mike Wang and Miquel Jubert Hermoso and Mo Metanat and Mohammad Rastegari and Munish Bansal and Nandhini Santhanam and Natascha Parks and Natasha White and Navyata Bawa and Nayan Singhal and Nick Egebo and Nicolas Usunier and Nikhil Mehta and Nikolay Pavlovich Laptev and Ning Dong and Norman Cheng and Oleg Chernoguz and Olivia Hart and Omkar Salpekar and Ozlem Kalinli and Parkin Kent and Parth Parekh and Paul Saab and Pavan Balaji and Pedro Rittner and Philip Bontrager and Pierre Roux and Piotr Dollar and Polina Zvyagina and Prashant Ratanchandani and Pritish Yuvraj and Qian Liang and Rachad Alao and Rachel Rodriguez and Rafi Ayub and Raghotham Murthy and Raghu Nayani and Rahul Mitra and Rangaprabhu Parthasarathy and Raymond Li and Rebekkah Hogan and Robin Battey and Rocky Wang and Russ Howes and Ruty Rinott and Sachin Mehta and Sachin Siby and Sai Jayesh Bondu and Samyak Datta and Sara Chugh and Sara Hunt and Sargun Dhillon and Sasha Sidorov and Satadru Pan and Saurabh Mahajan and Saurabh Verma and Seiji Yamamoto and Sharadh Ramaswamy and Shaun Lindsay and Shaun Lindsay and Sheng Feng and Shenghao Lin and Shengxin Cindy Zha and Shishir Patil and Shiva Shankar and Shuqiang Zhang and Shuqiang Zhang and Sinong Wang and Sneha Agarwal and Soji Sajuyigbe and Soumith Chintala and Stephanie Max and Stephen Chen and Steve Kehoe and Steve Satterfield and Sudarshan Govindaprasad and Sumit Gupta and Summer Deng and Sungmin Cho and Sunny Virk and Suraj Subramanian and Sy Choudhury and Sydney Goldman and Tal Remez and Tamar Glaser and Tamara Best and Thilo Koehler and Thomas Robinson and Tianhe Li and Tianjun Zhang and Tim Matthews and Timothy Chou and Tzook Shaked and Varun Vontimitta and Victoria Ajayi and Victoria Montanez and Vijai Mohan and Vinay Satish Kumar and Vishal Mangla and Vlad Ionescu and Vlad Poenaru and Vlad Tiberiu Mihailescu and Vladimir Ivanov and Wei Li and Wenchen Wang and Wenwen Jiang and Wes Bouaziz and Will Constable and Xiaocheng Tang and Xiaojian Wu and Xiaolan Wang and Xilun Wu and Xinbo Gao and Yaniv Kleinman and Yanjun Chen and Ye Hu and Ye Jia and Ye Qi and Yenda Li and Yilin Zhang and Ying Zhang and Yossi Adi and Youngjin Nam and Yu and Wang and Yu Zhao and Yuchen Hao and Yundi Qian and Yunlu Li and Yuzi He and Zach Rait and Zachary DeVito and Zef Rosnbrick and Zhaoduo Wen and Zhenyu Yang and Zhiwei Zhao and Zhiyu Ma},
      year={2024},
      eprint={2407.21783},
      archivePrefix={arXiv},
      primaryClass={cs.AI},
      url={https://arxiv.org/abs/2407.21783}, 
}

@misc{qwen3technicalreport,
      title={Qwen3 Technical Report}, 
      author={An Yang and Anfeng Li and Baosong Yang and Beichen Zhang and Binyuan Hui and Bo Zheng and Bowen Yu and Chang Gao and Chengen Huang and Chenxu Lv and Chujie Zheng and Dayiheng Liu and Fan Zhou and Fei Huang and Feng Hu and Hao Ge and Haoran Wei and Huan Lin and Jialong Tang and Jian Yang and Jianhong Tu and Jianwei Zhang and Jianxin Yang and Jiaxi Yang and Jing Zhou and Jingren Zhou and Junyang Lin and Kai Dang and Keqin Bao and Kexin Yang and Le Yu and Lianghao Deng and Mei Li and Mingfeng Xue and Mingze Li and Pei Zhang and Peng Wang and Qin Zhu and Rui Men and Ruize Gao and Shixuan Liu and Shuang Luo and Tianhao Li and Tianyi Tang and Wenbiao Yin and Xingzhang Ren and Xinyu Wang and Xinyu Zhang and Xuancheng Ren and Yang Fan and Yang Su and Yichang Zhang and Yinger Zhang and Yu Wan and Yuqiong Liu and Zekun Wang and Zeyu Cui and Zhenru Zhang and Zhipeng Zhou and Zihan Qiu},
      year={2025},
      eprint={2505.09388},
      archivePrefix={arXiv},
      primaryClass={cs.CL},
      url={https://arxiv.org/abs/2505.09388}, 
}

@inproceedings{dettmers2023qloraefficientfinetuningquantized,
author = {Dettmers, Tim and Pagnoni, Artidoro and Holtzman, Ari and Zettlemoyer, Luke},
title = {QLORA: efficient finetuning of quantized LLMs},
year = {2023},
publisher = {Curran Associates Inc.},
address = {Red Hook, NY, USA},
booktitle = {Proceedings of the 37th International Conference on Neural Information Processing Systems},
articleno = {441},
numpages = {28},
location = {New Orleans, LA, USA},
series = {NIPS '23}
}

@inproceedings{sengupta2023good,
title={A good learner can teach better: Teacher-student collaborative knowledge distillation},
author={Sengupta, Ayan and Dixit, Shantanu and Akhtar, Md Shad and Chakraborty, Tanmoy},
booktitle={Proceedings of the The Twelfth International Conference on Learning Representations, Virtual Event},
pages={25--29},
year={2023}
}

@misc{wang2025metalearnedmodalityweightedknowledgedistillation,
      title={Meta-Learned Modality-Weighted Knowledge Distillation for Robust Multi-Modal Learning with Missing Data}, 
      author={Hu Wang and Salma Hassan and Yuyuan Liu and Congbo Ma and Yuanhong Chen and Qing Li and Jiahui Geng and Bingjie Wang and Yu Tian and Yutong Xie and Jodie Avery and Louise Hull and Ian Reid and Mohammad Yaqub and Gustavo Carneiro},
      year={2025},
      eprint={2405.07155},
      archivePrefix={arXiv},
      primaryClass={cs.CV},
      url={https://arxiv.org/abs/2405.07155}, 
}

@misc{hinton2015distillingknowledgeneuralnetwork,
      title={Distilling the Knowledge in a Neural Network}, 
      author={Geoffrey Hinton and Oriol Vinyals and Jeff Dean},
      year={2015},
      eprint={1503.02531},
      archivePrefix={arXiv},
      primaryClass={stat.ML},
      url={https://arxiv.org/abs/1503.02531}, 
}

@inproceedings{jiao-etal-2020-tinybert,
    title = "{T}iny{BERT}: Distilling {BERT} for Natural Language Understanding",
    author = "Jiao, Xiaoqi  and
      Yin, Yichun  and
      Shang, Lifeng  and
      Jiang, Xin  and
      Chen, Xiao  and
      Li, Linlin  and
      Wang, Fang  and
      Liu, Qun",
    editor = "Cohn, Trevor  and
      He, Yulan  and
      Liu, Yang",
    booktitle = "Findings of the Association for Computational Linguistics: EMNLP 2020",
    month = nov,
    year = "2020",
    address = "Online",
    publisher = "Association for Computational Linguistics",
    url = "https://aclanthology.org/2020.findings-emnlp.372/",
    doi = "10.18653/v1/2020.findings-emnlp.372",
    pages = "4163--4174"
}

@misc{zhang2025rsparserankawareactivationsparsity,
      title={R-Sparse: Rank-Aware Activation Sparsity for Efficient LLM Inference}, 
      author={Zhenyu Zhang and Zechun Liu and Yuandong Tian and Harshit Khaitan and Zhangyang Wang and Steven Li},
      year={2025},
      eprint={2504.19449},
      archivePrefix={arXiv},
      primaryClass={cs.LG},
      url={https://arxiv.org/abs/2504.19449}, 
}

@misc{liu2025trainingfreeactivationsparsitylarge,
      title={Training-Free Activation Sparsity in Large Language Models}, 
      author={James Liu and Pragaash Ponnusamy and Tianle Cai and Han Guo and Yoon Kim and Ben Athiwaratkun},
      year={2025},
      eprint={2408.14690},
      archivePrefix={arXiv},
      primaryClass={cs.CL},
      url={https://arxiv.org/abs/2408.14690}, 
}

@inproceedings{
guo2025slimllm,
title={Slim{LLM}: Accurate Structured Pruning for Large Language Models},
author={Jialong Guo and Xinghao Chen and Yehui Tang and Yunhe Wang},
booktitle={Forty-second International Conference on Machine Learning},
year={2025},
url={https://openreview.net/forum?id=2xjUkU7FDb}
}

@misc{luo2025sparsinglawlargelanguage,
      title={Sparsing Law: Towards Large Language Models with Greater Activation Sparsity}, 
      author={Yuqi Luo and Chenyang Song and Xu Han and Yingfa Chen and Chaojun Xiao and Xiaojun Meng and Liqun Deng and Jiansheng Wei and Zhiyuan Liu and Maosong Sun},
      year={2025},
      eprint={2411.02335},
      archivePrefix={arXiv},
      primaryClass={cs.LG},
      url={https://arxiv.org/abs/2411.02335}, 
}

@inproceedings{Dhar_2024,
   title={Activation Sparsity Opportunities for Compressing General Large Language Models},
   url={http://dx.doi.org/10.1109/IPCCC59868.2024.10850382},
   DOI={10.1109/ipccc59868.2024.10850382},
   booktitle={2024 IEEE International Performance, Computing, and Communications Conference (IPCCC)},
   publisher={IEEE},
   author={Dhar, Nobel and Deng, Bobin and Islam, Md Romyull and Ahmad Nasif, Kazi Fahim and Zhao, Liang and Suo, Kun},
   year={2024},
   month=nov, pages={1–9} }

@inproceedings{
ding2025cbqcrossblockquantizationlarge,
title={{CBQ}: Cross-Block Quantization for Large Language Models},
author={Xin Ding and Xiaoyu Liu and Zhijun Tu and Yun Zhang and Wei Li and Jie Hu and Hanting Chen and Yehui Tang and Zhiwei Xiong and Baoqun Yin and Yunhe Wang},
booktitle={The Thirteenth International Conference on Learning Representations},
year={2025},
url={https://openreview.net/forum?id=eW4yh6HKz4}
}

@inproceedings{Guan_2024, series={DAC ’24},
   title={APTQ: Attention-aware Post-Training Mixed-Precision Quantization for Large Language Models},
   url={http://dx.doi.org/10.1145/3649329.3658498},
   DOI={10.1145/3649329.3658498},
   booktitle={Proceedings of the 61st ACM/IEEE Design Automation Conference},
   publisher={ACM},
   author={Guan, Ziyi and Huang, Hantao and Su, Yupeng and Huang, Hong and Wong, Ngai and Yu, Hao},
   year={2024},
   month=jun, pages={1–6},
   collection={DAC ’24} }

@inproceedings{Liu_2023,
   title={LLM-FP4: 4-Bit Floating-Point Quantized Transformers},
   url={http://dx.doi.org/10.18653/v1/2023.emnlp-main.39},
   DOI={10.18653/v1/2023.emnlp-main.39},
   booktitle={Proceedings of the 2023 Conference on Empirical Methods in Natural Language Processing},
   publisher={Association for Computational Linguistics},
   author={Liu, Shih-yang and Liu, Zechun and Huang, Xijie and Dong, Pingcheng and Cheng, Kwang-Ting},
   year={2023},
   pages={592–605} }

@InProceedings{kim-integer-only-quantization,
  title = 	 {I-BERT: Integer-only BERT Quantization},
  author =       {Kim, Sehoon and Gholami, Amir and Yao, Zhewei and Mahoney, Michael W. and Keutzer, Kurt},
  booktitle = 	 {Proceedings of the 38th International Conference on Machine Learning},
  pages = 	 {5506--5518},
  year = 	 {2021},
  editor = 	 {Meila, Marina and Zhang, Tong},
  volume = 	 {139},
  series = 	 {Proceedings of Machine Learning Research},
  month = 	 {18--24 Jul},
  publisher =    {PMLR},
  url = 	 {https://proceedings.mlr.press/v139/kim21d.html}
}

@misc{bhandare2019efficient8bitquantizationtransformer,
      title={Efficient 8-Bit Quantization of Transformer Neural Machine Language Translation Model}, 
      author={Aishwarya Bhandare and Vamsi Sripathi and Deepthi Karkada and Vivek Menon and Sun Choi and Kushal Datta and Vikram Saletore},
      year={2019},
      eprint={1906.00532},
      archivePrefix={arXiv},
      primaryClass={cs.LG},
      url={https://arxiv.org/abs/1906.00532}, 
}

@misc{shopkhoev2025replaceme0,
      title={ReplaceMe: Network Simplification via Depth Pruning and Transformer Block Linearization}, 
      author={Dmitriy Shopkhoev and Ammar Ali and Magauiya Zhussip and Valentin Malykh and Stamatios Lefkimmiatis and Nikos Komodakis and Sergey Zagoruyko},
      year={2025},
      eprint={2505.02819},
      archivePrefix={arXiv},
      primaryClass={cs.CL},
      url={https://arxiv.org/abs/2505.02819}, 
}

@InProceedings{sleb,
  title = 	 {{SLEB}: Streamlining {LLM}s through Redundancy Verification and Elimination of Transformer Blocks},
  author =       {Song, Jiwon and Oh, Kyungseok and Kim, Taesu and Kim, Hyungjun and Kim, Yulhwa and Kim, Jae-Joon},
  booktitle = 	 {Proceedings of the 41st International Conference on Machine Learning},
  pages = 	 {46136--46155},
  year = 	 {2024},
  editor = 	 {Salakhutdinov, Ruslan and Kolter, Zico and Heller, Katherine and Weller, Adrian and Oliver, Nuria and Scarlett, Jonathan and Berkenkamp, Felix},
  volume = 	 {235},
  series = 	 {Proceedings of Machine Learning Research},
  month = 	 {21--27 Jul},
  publisher =    {PMLR},
  url = 	 {https://proceedings.mlr.press/v235/song24f.html}
}

@inproceedings{voita-etal-2019-analyzing-heavy-lifting,
    title = "Analyzing Multi-Head Self-Attention: Specialized Heads Do the Heavy Lifting, the Rest Can Be Pruned",
    author = "Voita, Elena  and
      Talbot, David  and
      Moiseev, Fedor  and
      Sennrich, Rico  and
      Titov, Ivan",
    editor = "Korhonen, Anna  and
      Traum, David  and
      M{\`a}rquez, Llu{\'i}s",
    booktitle = "Proceedings of the 57th Annual Meeting of the Association for Computational Linguistics",
    month = jul,
    year = "2019",
    address = "Florence, Italy",
    publisher = "Association for Computational Linguistics",
    url = "https://aclanthology.org/P19-1580/",
    doi = "10.18653/v1/P19-1580",
    pages = "5797--5808"
}

@inproceedings{aresixteenheadsbetterthanone,
 author = {Michel, Paul and Levy, Omer and Neubig, Graham},
 booktitle = {Advances in Neural Information Processing Systems},
 editor = {H. Wallach and H. Larochelle and A. Beygelzimer and F. d\textquotesingle Alch\'{e}-Buc and E. Fox and R. Garnett},
 pages = {},
 publisher = {Curran Associates, Inc.},
 title = {Are Sixteen Heads Really Better than One?},
 url = {https://proceedings.neurips.cc/paper_files/paper/2019/file/2c601ad9d2ff9bc8b282670cdd54f69f-Paper.pdf},
 volume = {32},
 year = {2019}
}

@inproceedings{optimalbrainsurgeon,
 author = {Hassibi, Babak and Stork, David},
 booktitle = {Advances in Neural Information Processing Systems},
 editor = {S. Hanson and J. Cowan and C. Giles},
 pages = {},
 publisher = {Morgan-Kaufmann},
 title = {Second order derivatives for network pruning: Optimal Brain Surgeon},
 url = {https://proceedings.neurips.cc/paper_files/paper/1992/file/303ed4c69846ab36c2904d3ba8573050-Paper.pdf},
 volume = {5},
 year = {1992}
}

@misc{men2024shortgptlayerslargelanguage,
      title={ShortGPT: Layers in Large Language Models are More Redundant Than You Expect}, 
      author={Xin Men and Mingyu Xu and Qingyu Zhang and Bingning Wang and Hongyu Lin and Yaojie Lu and Xianpei Han and Weipeng Chen},
      year={2024},
      eprint={2403.03853},
      archivePrefix={arXiv},
      primaryClass={cs.CL},
      url={https://arxiv.org/abs/2403.03853}, 
}

@inproceedings{frantar2023sparsegptmassivelanguagemodels,
author = {Frantar, Elias and Alistarh, Dan},
title = {SparseGPT: massive language models can be accurately pruned in one-shot},
year = {2023},
publisher = {JMLR.org},
booktitle = {Proceedings of the 40th International Conference on Machine Learning},
articleno = {414},
numpages = {15},
location = {Honolulu, Hawaii, USA},
series = {ICML'23}
}

@misc{han2015learningweightsconnectionsefficient,
      title={Learning both Weights and Connections for Efficient Neural Networks}, 
      author={Song Han and Jeff Pool and John Tran and William J. Dally},
      year={2015},
      eprint={1506.02626},
      archivePrefix={arXiv},
      primaryClass={cs.NE},
      url={https://arxiv.org/abs/1506.02626}, 
}

@misc{ashkboos2024slicegptcompresslargelanguage,
      title={SliceGPT: Compress Large Language Models by Deleting Rows and Columns}, 
      author={Saleh Ashkboos and Maximilian L. Croci and Marcelo Gennari do Nascimento and Torsten Hoefler and James Hensman},
      year={2024},
      eprint={2401.15024},
      archivePrefix={arXiv},
      primaryClass={cs.LG},
      url={https://arxiv.org/abs/2401.15024}, 
}

@inproceedings{ma2023llmpruner,
author = {Ma, Xinyin and Fang, Gongfan and Wang, Xinchao},
title = {LLM-pruner: on the structural pruning of large language models},
year = {2023},
publisher = {Curran Associates Inc.},
address = {Red Hook, NY, USA},
booktitle = {Proceedings of the 37th International Conference on Neural Information Processing Systems},
articleno = {950},
numpages = {19},
location = {New Orleans, LA, USA},
series = {NIPS '23}
}

@misc{wang2024svd,
      title={SVD-LLM: Truncation-aware Singular Value Decomposition for Large Language Model Compression}, 
      author={Xin Wang and Yu Zheng and Zhongwei Wan and Mi Zhang},
      year={2025},
      eprint={2403.07378},
      archivePrefix={arXiv},
      primaryClass={cs.CL},
      url={https://arxiv.org/abs/2403.07378}, 
}

@misc{merity2016pointer,
      title={Pointer Sentinel Mixture Models}, 
      author={Stephen Merity and Caiming Xiong and James Bradbury and Richard Socher},
      year={2016},
      eprint={1609.07843},
      archivePrefix={arXiv},
      primaryClass={cs.CL},
      url={https://arxiv.org/abs/1609.07843}, 
}

@misc{eval-harness,
  author       = {Gao, Leo and Tow, Jonathan and Abbasi, Baber and Biderman, Stella and Black, Sid and DiPofi, Anthony and Foster, Charles and Golding, Laurence and Hsu, Jeffrey and Le Noac'h, Alain and Li, Haonan and McDonell, Kyle and Muennighoff, Niklas and Ociepa, Chris and Phang, Jason and Reynolds, Laria and Schoelkopf, Hailey and Skowron, Aviya and Sutawika, Lintang and Tang, Eric and Thite, Anish and Wang, Ben and Wang, Kevin and Zou, Andy},
  title        = {A framework for few-shot language model evaluation},
  month        = 07,
  year         = 2024,
  publisher    = {Zenodo},
  version      = {v0.4.3},
  doi          = {10.5281/zenodo.12608602},
  url          = {https://zenodo.org/records/12608602}
}

@article{reddy2019coqa,
    title = "{C}o{QA}: A Conversational Question Answering Challenge",
    author = "Reddy, Siva  and
      Chen, Danqi  and
      Manning, Christopher D.",
    editor = "Lee, Lillian  and
      Johnson, Mark  and
      Roark, Brian  and
      Nenkova, Ani",
    journal = "Transactions of the Association for Computational Linguistics",
    volume = "7",
    year = "2019",
    address = "Cambridge, MA",
    publisher = "MIT Press",
    url = "https://aclanthology.org/Q19-1016/",
    doi = "10.1162/tacl_a_00266",
    pages = "249--266"
}

@inproceedings{lin-etal-2022-truthfulqa,
    title = "{T}ruthful{QA}: Measuring How Models Mimic Human Falsehoods",
    author = "Lin, Stephanie  and
      Hilton, Jacob  and
      Evans, Owain",
    editor = "Muresan, Smaranda  and
      Nakov, Preslav  and
      Villavicencio, Aline",
    booktitle = "Proceedings of the 60th Annual Meeting of the Association for Computational Linguistics (Volume 1: Long Papers)",
    month = may,
    year = "2022",
    address = "Dublin, Ireland",
    publisher = "Association for Computational Linguistics",
    url = "https://aclanthology.org/2022.acl-long.229/",
    doi = "10.18653/v1/2022.acl-long.229",
    pages = "3214--3252"
}

@inproceedings{rudinger-etal-2018-gender,
    title = "Gender Bias in Coreference Resolution",
    author = "Rudinger, Rachel  and
      Naradowsky, Jason  and
      Leonard, Brian  and
      Van Durme, Benjamin",
    editor = "Walker, Marilyn  and
      Ji, Heng  and
      Stent, Amanda",
    booktitle = "Proceedings of the 2018 Conference of the North {A}merican Chapter of the Association for Computational Linguistics: Human Language Technologies, Volume 2 (Short Papers)",
    month = jun,
    year = "2018",
    address = "New Orleans, Louisiana",
    publisher = "Association for Computational Linguistics",
    url = "https://aclanthology.org/N18-2002/",
    doi = "10.18653/v1/N18-2002",
    pages = "8--14"
}

@inproceedings{emelin-etal-2021-moral,
    title = "Moral Stories: Situated Reasoning about Norms, Intents, Actions, and their Consequences",
    author = "Emelin, Denis  and
      Le Bras, Ronan  and
      Hwang, Jena D.  and
      Forbes, Maxwell  and
      Choi, Yejin",
    editor = "Moens, Marie-Francine  and
      Huang, Xuanjing  and
      Specia, Lucia  and
      Yih, Scott Wen-tau",
    booktitle = "Proceedings of the 2021 Conference on Empirical Methods in Natural Language Processing",
    month = nov,
    year = "2021",
    address = "Online and Punta Cana, Dominican Republic",
    publisher = "Association for Computational Linguistics",
    url = "https://aclanthology.org/2021.emnlp-main.54/",
    doi = "10.18653/v1/2021.emnlp-main.54",
    pages = "698--718"
}

@misc{SlimOrca,
  title = {SlimOrca: An Open Dataset of GPT-4 Augmented FLAN Reasoning Traces, with Verification},
  author = {Wing Lian and Guan Wang and Bleys Goodson and Eugene Pentland and Austin Cook and Chanvichet Vong and "Teknium"},
  year = {2023},
  publisher = {HuggingFace},
  url = {https://huggingface.co/datasets/Open-Orca/SlimOrca}
}

@misc{mukherjee2023orca,
      title={Orca: Progressive Learning from Complex Explanation Traces of GPT-4}, 
      author={Subhabrata Mukherjee and Arindam Mitra and Ganesh Jawahar and Sahaj Agarwal and Hamid Palangi and Ahmed Awadallah},
      year={2023},
      eprint={2306.02707},
      archivePrefix={arXiv},
      primaryClass={cs.CL},
      url={https://arxiv.org/abs/2306.02707}, 
}

@misc{qwen_qwen2._2025,
      title={Qwen2.5 Technical Report}, 
      author={Qwen and : and An Yang and Baosong Yang and Beichen Zhang and Binyuan Hui and Bo Zheng and Bowen Yu and Chengyuan Li and Dayiheng Liu and Fei Huang and Haoran Wei and Huan Lin and Jian Yang and Jianhong Tu and Jianwei Zhang and Jianxin Yang and Jiaxi Yang and Jingren Zhou and Junyang Lin and Kai Dang and Keming Lu and Keqin Bao and Kexin Yang and Le Yu and Mei Li and Mingfeng Xue and Pei Zhang and Qin Zhu and Rui Men and Runji Lin and Tianhao Li and Tianyi Tang and Tingyu Xia and Xingzhang Ren and Xuancheng Ren and Yang Fan and Yang Su and Yichang Zhang and Yu Wan and Yuqiong Liu and Zeyu Cui and Zhenru Zhang and Zihan Qiu},
      year={2025},
      eprint={2412.15115},
      archivePrefix={arXiv},
      primaryClass={cs.CL},
      url={https://arxiv.org/abs/2412.15115}, 
}

@misc{nguyen_survey_2024,
      title={A Survey of Small Language Models}, 
      author={Chien Van Nguyen and Xuan Shen and Ryan Aponte and Yu Xia and Samyadeep Basu and Zhengmian Hu and Jian Chen and Mihir Parmar and Sasidhar Kunapuli and Joe Barrow and Junda Wu and Ashish Singh and Yu Wang and Jiuxiang Gu and Franck Dernoncourt and Nesreen K. Ahmed and Nedim Lipka and Ruiyi Zhang and Xiang Chen and Tong Yu and Sungchul Kim and Hanieh Deilamsalehy and Namyong Park and Mike Rimer and Zhehao Zhang and Huanrui Yang and Ryan A. Rossi and Thien Huu Nguyen},
      year={2024},
      eprint={2410.20011},
      archivePrefix={arXiv},
      primaryClass={cs.CL},
      url={https://arxiv.org/abs/2410.20011}, 
}

@misc{yuan2023asvd,
      title={ASVD: Activation-aware Singular Value Decomposition for Compressing Large Language Models}, 
      author={Zhihang Yuan and Yuzhang Shang and Yue Song and Dawei Yang and Qiang Wu and Yan Yan and Guangyu Sun},
      year={2025},
      eprint={2312.05821},
      archivePrefix={arXiv},
      primaryClass={cs.CL},
      url={https://arxiv.org/abs/2312.05821}, 
}

@incollection{kellerer2013knapsack,
  title={Knapsack problems},
  author={Pisinger, David and Toth, Paolo},
  booktitle={Handbook of Combinatorial Optimization: Volume1--3},
  pages={299--428},
  year={1998},
  publisher={Springer}
}

@misc{zhu2024surveymodelcompressionlarge,
      title={A Survey on Model Compression for Large Language Models}, 
      author={Xunyu Zhu and Jian Li and Yong Liu and Can Ma and Weiping Wang},
      year={2024},
      eprint={2308.07633},
      archivePrefix={arXiv},
      primaryClass={cs.CL},
      url={https://arxiv.org/abs/2308.07633}, 
}

@misc{hsu2022language,
      title={Language model compression with weighted low-rank factorization}, 
      author={Yen-Chang Hsu and Ting Hua and Sungen Chang and Qian Lou and Yilin Shen and Hongxia Jin},
      year={2022},
      eprint={2207.00112},
      archivePrefix={arXiv},
      primaryClass={cs.LG},
      url={https://arxiv.org/abs/2207.00112}, 
}

@article{sakaguchi2021winogrande,
  title={Winogrande: An adversarial winograd schema challenge at scale},
  author={Sakaguchi, Keisuke and Bras, Ronan Le and Bhagavatula, Chandra and Choi, Yejin},
  journal={Communications of the ACM},
  volume={64},
  number={9},
  pages={99--106},
  year={2021},
  publisher={ACM New York, NY, USA}
}

@inproceedings{bisk2020piqa,
  author = {Yonatan Bisk and Rowan Zellers and
            Ronan Le Bras and Jianfeng Gao
            and Yejin Choi},
  title = {PIQA: Reasoning about Physical Commonsense in
           Natural Language},
  booktitle = {Thirty-Fourth AAAI Conference on
               Artificial Intelligence},
  year = {2020},
}

@misc{yang2025moregreencodelarge,
      title={Less is More: Towards Green Code Large Language Models via Unified Structural Pruning}, 
      author={Guang Yang and Yu Zhou and Xiangyu Zhang and Wei Cheng and Ke Liu and Xiang Chen and Terry Yue Zhuo and Taolue Chen},
      year={2025},
      eprint={2412.15921},
      archivePrefix={arXiv},
      primaryClass={cs.SE},
      url={https://arxiv.org/abs/2412.15921}, 
}

@misc{mishra2021acceleratingsparsedeepneural,
      title={Accelerating Sparse Deep Neural Networks}, 
      author={Asit Mishra and Jorge Albericio Latorre and Jeff Pool and Darko Stosic and Dusan Stosic and Ganesh Venkatesh and Chong Yu and Paulius Micikevicius},
      year={2021},
      eprint={2104.08378},
      archivePrefix={arXiv},
      primaryClass={cs.LG},
      url={https://arxiv.org/abs/2104.08378}, 
}

@inproceedings{paperno2016lambada,
    title = "The {LAMBADA} dataset: Word prediction requiring a broad discourse context",
    author = "Paperno, Denis  and
      Kruszewski, Germ{\'a}n  and
      Lazaridou, Angeliki  and
      Pham, Ngoc Quan  and
      Bernardi, Raffaella  and
      Pezzelle, Sandro  and
      Baroni, Marco  and
      Boleda, Gemma  and
      Fern{\'a}ndez, Raquel",
    editor = "Erk, Katrin  and
      Smith, Noah A.",
    booktitle = "Proceedings of the 54th Annual Meeting of the Association for Computational Linguistics (Volume 1: Long Papers)",
    month = aug,
    year = "2016",
    address = "Berlin, Germany",
    publisher = "Association for Computational Linguistics",
    url = "https://aclanthology.org/P16-1144/",
    doi = "10.18653/v1/P16-1144",
    pages = "1525--1534"
}

@inproceedings{
sengupta2025pruneoncedesigningcalibrationfree,
title={You Only Prune Once: Designing Calibration-Free Model Compression With Policy Learning},
author={Ayan Sengupta and Siddhant Chaudhary and Tanmoy Chakraborty},
booktitle={The Thirteenth International Conference on Learning Representations},
year={2025},
url={https://openreview.net/forum?id=5RZoYIT3u6}
}

@inproceedings{
chen2025dlp,
title={{DLP}: Dynamic Layerwise Pruning in Large Language Models},
author={Yuli Chen and Bo Cheng and Jiale Han and Yingying Zhang and Yingting Li and Shuhao Zhang},
booktitle={Forty-second International Conference on Machine Learning},
year={2025},
url={https://openreview.net/forum?id=11id5ppGZ8}
}

@inproceedings{
chen2025streamlining,
title={Streamlining Redundant Layers to Compress Large Language Models},
author={Xiaodong Chen and Yuxuan Hu and Jing Zhang and Yanling Wang and Cuiping Li and Hong Chen},
booktitle={The Thirteenth International Conference on Learning Representations},
year={2025},
url={https://openreview.net/forum?id=IC5RJvRoMp}
}

@inproceedings{
gromov2025the,
title={The Unreasonable Ineffectiveness of the Deeper Layers},
author={Andrey Gromov and Kushal Tirumala and Hassan Shapourian and Paolo Glorioso and Dan Roberts},
booktitle={The Thirteenth International Conference on Learning Representations},
year={2025},
url={https://openreview.net/forum?id=ngmEcEer8a}
}

@inproceedings{
wee2025promptbased,
title={Prompt-based Depth Pruning of Large Language Models},
author={Juyun Wee and Minjae Park and Jaeho Lee},
booktitle={Forty-second International Conference on Machine Learning},
year={2025},
url={https://openreview.net/forum?id=hRxHF1xPYB}
}

@inproceedings{
ouderaa2024the,
title={The {LLM} Surgeon},
author={Tycho F. A. van der Ouderaa and Markus Nagel and Mart Van Baalen and Tijmen Blankevoort},
booktitle={The Twelfth International Conference on Learning Representations},
year={2024},
url={https://openreview.net/forum?id=DYIIRgwg2i}
}

@inproceedings{
jibeaware2025,
title={Beware of Calibration Data for Pruning Large Language Models},
author={Yixin Ji and Yang Xiang and Juntao Li and Qingrong Xia and Ping Li and Xinyu Duan and Zhefeng Wang and Min Zhang},
booktitle={The Thirteenth International Conference on Learning Representations},
year={2025},
url={https://openreview.net/forum?id=x83w6yGIWb}
}

@misc{deepseekai2024deepseekv3technicalreport,
      title={DeepSeek-V3 Technical Report}, 
      author={DeepSeek-AI and Aixin Liu and Bei Feng and Bing Xue and Bingxuan Wang and Bochao Wu and Chengda Lu and Chenggang Zhao and Chengqi Deng and Chenyu Zhang and Chong Ruan and Damai Dai and Daya Guo and Dejian Yang and Deli Chen and Dongjie Ji and Erhang Li and Fangyun Lin and Fucong Dai and Fuli Luo and Guangbo Hao and Guanting Chen and Guowei Li and H. Zhang and Han Bao and Hanwei Xu and Haocheng Wang and Haowei Zhang and Honghui Ding and Huajian Xin and Huazuo Gao and Hui Li and Hui Qu and J. L. Cai and Jian Liang and Jianzhong Guo and Jiaqi Ni and Jiashi Li and Jiawei Wang and Jin Chen and Jingchang Chen and Jingyang Yuan and Junjie Qiu and Junlong Li and Junxiao Song and Kai Dong and Kai Hu and Kaige Gao and Kang Guan and Kexin Huang and Kuai Yu and Lean Wang and Lecong Zhang and Lei Xu and Leyi Xia and Liang Zhao and Litong Wang and Liyue Zhang and Meng Li and Miaojun Wang and Mingchuan Zhang and Minghua Zhang and Minghui Tang and Mingming Li and Ning Tian and Panpan Huang and Peiyi Wang and Peng Zhang and Qiancheng Wang and Qihao Zhu and Qinyu Chen and Qiushi Du and R. J. Chen and R. L. Jin and Ruiqi Ge and Ruisong Zhang and Ruizhe Pan and Runji Wang and Runxin Xu and Ruoyu Zhang and Ruyi Chen and S. S. Li and Shanghao Lu and Shangyan Zhou and Shanhuang Chen and Shaoqing Wu and Shengfeng Ye and Shengfeng Ye and Shirong Ma and Shiyu Wang and Shuang Zhou and Shuiping Yu and Shunfeng Zhou and Shuting Pan and T. Wang and Tao Yun and Tian Pei and Tianyu Sun and W. L. Xiao and Wangding Zeng and Wanjia Zhao and Wei An and Wen Liu and Wenfeng Liang and Wenjun Gao and Wenqin Yu and Wentao Zhang and X. Q. Li and Xiangyue Jin and Xianzu Wang and Xiao Bi and Xiaodong Liu and Xiaohan Wang and Xiaojin Shen and Xiaokang Chen and Xiaokang Zhang and Xiaosha Chen and Xiaotao Nie and Xiaowen Sun and Xiaoxiang Wang and Xin Cheng and Xin Liu and Xin Xie and Xingchao Liu and Xingkai Yu and Xinnan Song and Xinxia Shan and Xinyi Zhou and Xinyu Yang and Xinyuan Li and Xuecheng Su and Xuheng Lin and Y. K. Li and Y. Q. Wang and Y. X. Wei and Y. X. Zhu and Yang Zhang and Yanhong Xu and Yanhong Xu and Yanping Huang and Yao Li and Yao Zhao and Yaofeng Sun and Yaohui Li and Yaohui Wang and Yi Yu and Yi Zheng and Yichao Zhang and Yifan Shi and Yiliang Xiong and Ying He and Ying Tang and Yishi Piao and Yisong Wang and Yixuan Tan and Yiyang Ma and Yiyuan Liu and Yongqiang Guo and Yu Wu and Yuan Ou and Yuchen Zhu and Yuduan Wang and Yue Gong and Yuheng Zou and Yujia He and Yukun Zha and Yunfan Xiong and Yunxian Ma and Yuting Yan and Yuxiang Luo and Yuxiang You and Yuxuan Liu and Yuyang Zhou and Z. F. Wu and Z. Z. Ren and Zehui Ren and Zhangli Sha and Zhe Fu and Zhean Xu and Zhen Huang and Zhen Zhang and Zhenda Xie and Zhengyan Zhang and Zhewen Hao and Zhibin Gou and Zhicheng Ma and Zhigang Yan and Zhihong Shao and Zhipeng Xu and Zhiyu Wu and Zhongyu Zhang and Zhuoshu Li and Zihui Gu and Zijia Zhu and Zijun Liu and Zilin Li and Ziwei Xie and Ziyang Song and Ziyi Gao and Zizheng Pan},
      year={2024},
      eprint={2412.19437},
      archivePrefix={arXiv},
      primaryClass={cs.CL},
      url={https://arxiv.org/abs/2412.19437}, 
}

@misc{sun2024simpleeffectivepruningapproach,
      title={A Simple and Effective Pruning Approach for Large Language Models}, 
      author={Mingjie Sun and Zhuang Liu and Anna Bair and J. Zico Kolter},
      year={2024},
      eprint={2306.11695},
      archivePrefix={arXiv},
      primaryClass={cs.CL},
      url={https://arxiv.org/abs/2306.11695}, 
}

@misc{gu2023knowledge,
      title={MiniLLM: Knowledge Distillation of Large Language Models}, 
      author={Yuxian Gu and Li Dong and Furu Wei and Minlie Huang},
      year={2025},
      eprint={2306.08543},
      archivePrefix={arXiv},
      primaryClass={cs.CL},
      url={https://arxiv.org/abs/2306.08543}, 
}

@inproceedings{OpenBookQA2018,
    title = "Can a Suit of Armor Conduct Electricity? A New Dataset for Open Book Question Answering",
    author = "Mihaylov, Todor  and
      Clark, Peter  and
      Khot, Tushar  and
      Sabharwal, Ashish",
    editor = "Riloff, Ellen  and
      Chiang, David  and
      Hockenmaier, Julia  and
      Tsujii, Jun{'}ichi",
    booktitle = "Proceedings of the 2018 Conference on Empirical Methods in Natural Language Processing",
    month = oct # "-" # nov,
    year = "2018",
    address = "Brussels, Belgium",
    publisher = "Association for Computational Linguistics",
    url = "https://aclanthology.org/D18-1260/",
    doi = "10.18653/v1/D18-1260",
    pages = "2381--2391"
}












\clearpage
\beginappendix
\setcounter{tocdepth}{3}

\section{Implementation Details}
\label{appendix:implementation-details}
We implement \pruner\ using PyTorch~\citep{paszke-pytorch} and the HuggingFace Transformers library~\citep{wolf-etal-2020-transformers}. All experiments are conducted on a single NVIDIA A100 GPU with 80GB of VRAM and evaluated using the Language Model Evaluation Harness~\citep{eval-harness}. For importance estimation (Section~\ref{sec:importance_estimation}), we use a calibration set of 50 samples randomly drawn from the SlimOrca dataset~\citep{SlimOrca,mukherjee2023orca}, with a context length of 512 tokens.

For post-pruning recovery fine-tuning, we train the pruned models on 2,000 SlimOrca samples with a context length of 1,024 tokens. To maintain computational efficiency, we employ Low-Rank Adaptation~\citep{hu2022lora} rather than full fine-tuning. We use a LoRA rank of 64, a scaling factor of 16, and a dropout probability of 0.05. All models are fine-tuned for a single epoch with a batch size of 2 and a learning rate of $2 \times 10^{-4}$.

\section{Description of Evaluation Tasks}
\label{appendix:dataset-descriptions}
We devise a suite of 18 diverse tasks to evaluate different pruning methodologies comprehensively. We provide a description of each benchmark task in this section.

\paragraph{Generative Performance.} We compute log-perplexity on the WikiText-2 \citep{merity2016pointer} and LAMBADA \citep{paperno2016lambada} datasets to assess generative performance. These are language modeling datasets consisting of unlabeled English texts requiring models to understand local and global context to predict subsequent words.

\paragraph{World Understanding.} In order to assess models' common sense capabilities and general understanding of the physical world, we utilize the PIQA \citep{bisk2020piqa}, PROST \citep{aroca-ouellette-etal-2021-prost}, and CommonsenseQA \citep{talmor-etal-2019-commonsenseqa} datasets. PIQA contains multiple-choice questions (MCQs) testing a model's general understanding of the physical world with questions such as whether the process of boiling eggs requires water or not. PROST specifically focuses on evaluating model's understanding of each of 10 physical reasoning concepts such as direction, mass and height. CommonsenseQA also contains MCQs but extends its scope of evaluation to include different types of commonsense knowledge instead of specifically physical understanding.

\paragraph{Domain-Specific Knowledge.} We employ the ARC-Easy and ARC-Challenge \citep{clark2018thinksolvedquestionanswering} tasks which contain grade-school level multiple-choice science questions and are designed to test a model's knowledge of basic science. As their names suggest, the former is an easier version of the task than the latter. We evaluate mathematical capabilities using the MathQA \citep{amini-etal-2019-mathqa} task which contains multiple-choice mathematical word problems. Additionally, we employ the MedQA \citep{jin2021disease} task which is a challenging dataset containing multiple-choice medical questions from professional medical exams. Lastly, we evaluate a model's ability to combine its commonsense, domain knowledge and text comprehension capabilities by testing it on the OpenbookQA benchmark \citep{OpenBookQA2018} which contains complex reasoning questions that require models to refer to a given corpus of salient facts to answer correctly.

\paragraph{Natural Language Understanding and Inference.} We gauge a model's semantic understanding on natural understanding with the help of the BLIMP \citep{warstadt2020blimp}, BoolQ \citep{clark2019boolq}, LAMBADA \citep{paperno2016lambada}, Winogrande \citep{sakaguchi2021winogrande}, and CoQA \citep{reddy2019coqa} benchmarks. BLIMP gauges grammatical understanding by making models choose between pairs of sentences that differ in morphology, semantics, or syntax. On the other hand, BoolQ contains binary True/False-style questions based on a reference text, thereby testing a model's comprehension abilities. While we employ LAMBADA to assess generative performance by computing log-perplexity over it, we employ its next word prediction format that offers provides models with an incomplete passage and assesses the accuracy with which they are able to predict the next word in the passage by understanding long context. Winogrande tests semantic understanding by providing fill-in-the-blanks questions with binary options, requiring models to choose the most appropriate option given the context. CoQA contains questions based on a multi-turn conversation between two actors, thereby evaluating a model's ability to understand the nuances of long conversations and how they differ from regular text passages.

\paragraph{Safety, Bias \& and Ethics.} We assess a pruned model's alignment to safety and ethical principles by evaluating it on the challenging Winogender \citep{rudinger-etal-2018-gender}, TruthfulQA \citep{lin-etal-2022-truthfulqa}, and Moral Stories \citep{emelin-etal-2021-moral} tasks. Winogender assess systemic bias by offering pairs of highly similar texts that differ only by the gender of a pronoun (e.g., ``he'' changed to ``she''). TruthfulQA poses questions to models which humans may answer incorrectly due to personal biases or misconceptions, assessing the ability of a model to identify and ignore these biases to answer questions truthfully. Lastly, Moral Stories is a dataset containing narratives describing moral and immoral decisions taken by various actors with the goal of assessing whether LLMs are able to discriminate between moral and immoral actions, thereby gauging their alignment with socially acceptable norms.

\section{Detailed Results}
\label{appendix:detailed-results}
\subsection{Comparison of CRAFT Across Methods}
\label{appendix:craft-detailed-results}
\begin{table*}[!htb]
\centering
\begin{tabular}{ccccc}
\hline
\textbf{Method} & \textbf{Target CR (\%) $= r_t$} & \textbf{Observed CR (\%) $= r_o$} & $\Delta = r_t - r_o$ & \textbf{Avg. CRAFT} $= \frac{r_o}{r_t}$ \\
 &&& ($\downarrow$) & ($\uparrow$) \\
\hline
& 25 & 21.19 & 3.81 & \multirow{5}{*}{0.84}\\
ReplaceMe & 35 & 28.23 & 6.77 &\\
SLEB& 50 & 42.38 & 7.62 &\\
ShortGPT & 60 & 51.79 & 8.21 &\\
& 75 & 63.57 & 11.43 &\\
\hline
\multirow{5}{*}{SliceGPT} & 25 & 14.60 & 10.40 & \multirow{5}{*}{0.79}\\
& 35 & 25.99 & 9.01 &\\
& 50 & 42.13 & 7.87 &\\
& 60 & 52.82 & 7.18 &\\
& 75 & 67.86 & 7.14 &\\
\hline
\multirow{5}{*}{LLM-Pruner} & 25 & 17.63 & 7.37 & \multirow{5}{*}{0.58}\\
& 35 & 19.33 & 15.67 &\\
& 50 & 27.66 & 22.34 &\\
& 60 & 33.16 & 26.84 &\\
& 75 & 41.49 & 33.51 &\\
\hline
\multirow{5}{*}{2SSP} & 25 & 21.19 & 3.81 & \multirow{5}{*}{0.85}\\
& 35 & 29.64 & 5.36 &\\
& 50 & 42.38 & 7.62 &\\
& 60 & 50.87 & 9.13 &\\
& 75 & 63.57 & 11.43 &\\
\hline
\multirow{5}{*}{\pruner} & 25 & 24.62 & 0.38 & \multirow{5}{*}{0.97}\\
& 35 & 33.97 & 1.03 &\\
& 50 & 48.64 & 1.36 &\\
& 60 & 58.71 & 1.29 &\\
& 75 & 71.24 & 3.76 &\\
\hline
\end{tabular}%
\caption{Quantitative analysis of the \textit{Compression Adherence Factor} (CRAFT) for each method over a diverse range of target compression ratios (25\%, 35\%, 50\%, 60\%, 75\%).}
\label{tab:craft-analysis}
\end{table*}
As depicted in Figure~\ref{fig:motivation}, we observe significant disparities between the target and achieved compression ratios for the analyzed baselines, serving as the primary motivator for the dual-stage mixed-granularity design of \pruner. We provide the quantitative results for this key observation in Table~\ref{tab:craft-analysis}. As is evident, baselines exhibit poor adherence to the target compression ratio with CRAFTs going as low as 0.58, thereby reducing reliability and trust. On the contrary, we demonstrate that \pruner\ adheres tightly to the target CRs, making it much more reliable in practical settings. Note that since ReplaceMe, SLEB, and ShortGPT are depth pruning baselines, they prune a fixed number of layers for a given compression ratio, thereby leading to the same $\Delta$ and CRAFT.
\begin{table*}[!htb]
\centering
\resizebox{\textwidth}{!}{%
\begin{tabular}{|cc|cc|ccc|ccccc|}
\hline
\multirow{3}{*}{\textbf{CR}} & \multirow{3}{*}{\textbf{Method}} & \multicolumn{2}{c}{Generative} & \multicolumn{3}{|c|}{World Understanding} & \multicolumn{5}{c|}{Domain-Specific}\\
\cline{3-12}
& & \textbf{Wikitext} & \textbf{Lambada} & \textbf{PIQA} & \textbf{PROST} & \textbf{CommonsenseQA} & \textbf{OpenbookQA} & \textbf{MathQA} & \textbf{ARC-Easy} & \textbf{ARC-Challenge} & \textbf{MedQA} \\
 & & (Log-PPL $\downarrow$) & (Log-PPL $\downarrow$) & (Acc $\uparrow$) & (Acc $\uparrow$) & (Acc $\uparrow$) & (Acc $\uparrow$) & (Acc $\uparrow$) & (Acc $\uparrow$) & (Acc $\uparrow$) & (Acc $\uparrow$) \\
\hline
\multirow{1}{*}{-} & Base & 2.16 & 1.22 & 80.96 & 41.28 & 77.40 & 43.60 & 39.60 & 79.67 & 55.03 & 63.47 \\
\cdashline{1-12}
\multirow{7}{*}{25\%} 
 & ReplaceMe & 4.47 & 5.36 & 68.23 & 35.96 & 73.71 & 33.20 & 27.07 & 53.45 & 37.63 & 52.55\\
 & SliceGPT & 4.51 & 5.72 & 55.39 & 29.85 & 37.10 & 25.80 & 23.52 & 35.61 & 24.83 & 24.59\\
 & LLM-Pruner & 3.50 & 4.43 & 68.44 & 31.46 & 21.79 & 29.60 & 24.99 & 52.57 & 29.18 & 24.35\\
 & SLEB & 3.44 & 3.17 & 70.89 & 29.33 & 26.37 & 34.80 & 26.40 & 60.35 & 33.45 & 30.32\\
 & ShortGPT & 10.75 & 15.65 & 64.36 & 33.89 & 47.83 & 29.20 & 25.06 & 42.97 & 33.36 & 27.81\\
 & 2SSP & 3.27 & 2.27 & 68.66 & 30.28 & 49.06 & 33.00 & 27.71 & 57.74 & 35.92 & 28.99\\
 &  \pruner & 3.42 & 2.89 & 71.22 & 31.80 & 52.42 & 36.40 & 26.40 & 59.30 & 37.37 & 35.51\\
\hline
\multirow{7}{*}{35\%} 
 & ReplaceMe & 6.04 & 5.68 & 60.99 & 34.27 & 19.98 & 29.80 & 23.75 & 41.84 & 23.21 & 26.47 \\
 & SliceGPT & 5.19 & 7.70 & 54.08 & 30.75 & 30.47 & 27.00 & 21.84 & 31.82 & 22.44 & 25.30\\
 & LLM-Pruner & 4.27 & 5.61 & 62.02 & 28.48 & 21.05 & 28.00 & 25.13 & 43.22 & 24.83 & 26.87\\
 & SLEB & 4.13 & 4.81 & 67.36 & 29.77 & 20.48 & 29.20 & 24.05 & 47.39 & 27.82 & 26.47\\
 & ShortGPT & 11.85 & 14.43 & 59.58 & 31.78 & 30.96 & 27.00 & 22.25 & 34.26 & 30.12 & 31.34  \\
 & 2SSP & 3.97 & 3.57 & 60.45 & 28.64 & 22.77 & 26.60 & 25.33 & 40.99 & 25.26 & 28.75\\
 &  \pruner & 4.41 & 5.00 & 65.83 & 29.55 & 28.09 & 28.60 & 24.32 & 46.84 & 33.11 & 28.04\\
 \hline
\end{tabular}%
}

\vspace{0.3em}
\textit{(continued)}
\vspace{0.3em}

\resizebox{\textwidth}{!}{%
\begin{tabular}{|cc|ccccc|ccc|cc|}
\hline
\multirow{3}{*}{\textbf{CR}} & \multirow{3}{*}{\textbf{Method}} & \multicolumn{5}{c}{NLU \& NLI} & \multicolumn{3}{|c|}{Safety, Bias \& Ethics} & \multirow{3}{*}{\textbf{Avg RP (\%)}} & \multirow{3}{*}{\textbf{Std RP (\%)}}\\
\cline{3-10}
& & \textbf{BLIMP} & \textbf{BoolQ} & \textbf{Lambada} & \textbf{Winogrande} & \textbf{CoQA} & \textbf{Winogender} & \textbf{TruthfulQA} & \textbf{Moral Stories} & & \\
& & (Acc $\uparrow$) & (Acc $\uparrow$) & (Acc $\uparrow$) & (Acc $\uparrow$) & (Acc $\uparrow$) & (Acc $\uparrow$) & (Acc $\uparrow$) & (Acc $\uparrow$) & ($\uparrow$) & ($\downarrow$) \\
\hline
\multirow{1}{*}{-} & Base & 83.42 & 83.91 & 73.08 & 73.40 & 78.80 & 62.50 & 54.00 & 50.17 & - & -\\
\cdashline{1-12}
\multirow{7}{*}{25\%} 
 & ReplaceMe & 74.97 & 79.76 & 14.17 & 66.14 & 39.80 & 57.64 & 55.20 & 54.01 & 74.86 & 20.40 \\
 & SliceGPT & 75.45 & 59.45 & 18.44 & 50.59 & 64.80 & 49.44 & 45.91 & 55.33 & 62.06 & 20.20  \\
 & LLM-Pruner & 80.15 & 63.95 & 21.54 & 53.20 & 30.40 & 53.47 & 45.05 & 57.09 & 64.56 & 22.40  \\
 & SLEB & 80.26 & 57.00 & 37.57 & 58.48 & 20.60 & 53.47 & 44.60 & 54.98 & 67.91 & 20.30  \\
 & ShortGPT & 69.01 & 37.83 & 2.83 & 58.25 & 4.60 & 49.44 & 51.88 & 51.71 & 57.50 & 26.10  \\
 & 2SSP & 82.92 & 68.26 & 55.29 & 62.12 & 72.10 & 55.42 & 42.08 & 55.29 & 76.65 & 14.00\\
 &  \pruner & 80.30 & 65.08 & 47.27 & 65.51 & 66.60 & 57.78 & 47.19 & 54.10 & \textbf{77.03} & \textbf{12.70}\\
\hline
\multirow{7}{*}{35\%} 
 & ReplaceMe & 76.94 & 37.89 & 13.16 & 51.86 & 0.60 & 50.42 & 48.76 & 54.53 & 56.25 & 27.10 \\
 & SliceGPT & 72.83 & 45.38 & 8.37 & 50.12 & 36.90 & 50.83 & 46.25 & 55.50 & 56.74 & 22.50  \\
 & LLM-Pruner & 78.69 & 58.75 & 13.76 & 49.88 & 21.90 & 50.56 & 46.61 & 56.83 & 59.66 & 23.70  \\
 & SLEB & 75.99 & 61.56 & 21.46 & 52.72 & 8.90 & 52.78 & 44.54 & 55.29 & 60.72 & 24.10 \\
 & ShortGPT & 61.67 & 59.30 & 1.32 & 57.70 & 2.60 & 55.83 & 52.16 & 57.10 & 56.14 & 28.00  \\
 & 2SSP & 82.54 & 56.15 & 36.50 & 53.99 & 48.60 & 52.92 & 42.79 & 55.78 & 64.21 & 19.10\\
 &  \pruner & 76.98 & 62.63 & 21.25 & 60.46 & 46.30 & 52.50 & 46.69 & 54.69 & \textbf{64.96} & \textbf{19.50}\\
 \hline
\end{tabular}%
}
\caption{Task-specific pruning results for different pruning methods on Llama-3.1-8B-Instruct across various compression ratios (CR) without RFT.}
\label{tab:main_results_llama_full_norft}
\end{table*}

\begin{table*}[!htb]
\centering
\resizebox{\textwidth}{!}{%
\begin{tabular}{|cc|cc|ccc|ccccc|}
\hline
\multirow{3}{*}{\textbf{CR}} & \multirow{3}{*}{\textbf{Method}} & \multicolumn{2}{c}{Generative} & \multicolumn{3}{|c|}{World Understanding} & \multicolumn{5}{c|}{Domain-Specific}\\
\cline{3-12}
& & \textbf{Wikitext} & \textbf{Lambada} & \textbf{PIQA} & \textbf{PROST} & \textbf{CommonsenseQA} & \textbf{OpenbookQA} & \textbf{MathQA} & \textbf{ARC-Easy} & \textbf{ARC-Challenge} & \textbf{MedQA} \\
 & & (Log-PPL $\downarrow$) & (Log-PPL $\downarrow$) & (Acc $\uparrow$) & (Acc $\uparrow$) & (Acc $\uparrow$) & (Acc $\uparrow$) & (Acc $\uparrow$) & (Acc $\uparrow$) & (Acc $\uparrow$) & (Acc $\uparrow$) \\
\hline
\multirow{1}{*}{-} & Base & 2.16 & 1.22 & 80.96 & 41.28 & 77.40 & 43.60 & 39.60 & 79.67 & 55.03 & 63.47 \\
\cdashline{1-12}
\multirow{7}{*}{25\%} 
 & ReplaceMe & 2.92 & 1.80 & 70.73 & 33.61 & 70.60 & 34.40 & 27.94 & 59.34 & 39.59 & 55.46\\
 & SliceGPT & 3.29 & 3.91 & 61.37 & 32.53 & 53.40 & 26.40 & 23.89 & 44.66 & 30.97 & 24.51\\
 & LLM-Pruner & 2.79 & 2.29 & 72.63 & 29.52 & 30.96 & 34.20 & 27.14 & 58.92 & 36.60 & 24.75\\
 & SLEB & 2.72 & 1.79 & 73.56 & 24.79 & 44.23 & 37.60 & 29.62 & 60.56 & 36.86 & 29.30\\
 & ShortGPT & 2.95 & 1.98 & 70.62 & 32.93 & 74.12 & 37.40 & 26.94 & 56.65 & 38.65 & 52.47\\
 & 2SSP & 2.64 & 1.63 & 73.56 & 29.18 & 59.95 & 34.60 & 31.36 & 62.54 & 40.19 & 30.09\\
 &  \pruner & 2.85 & 1.72 & 70.40 & 34.73 & 68.55 & 35.60 & 28.51 & 58.17 & 40.19 & 46.11\\
\hline
\multirow{7}{*}{35\%} 
 & ReplaceMe & 3.40 & 2.98 & 68.72 & 30.39 & 20.31 & 28.60 & 24.29 & 45.83 & 26.28 & 23.10\\
 & SliceGPT & 3.65 & 5.37 & 58.98 & 29.48 & 45.86 & 27.20 & 22.78 & 39.94 & 27.30 & 24.82\\
 & LLM-Pruner & 3.07 & 2.81 & 67.41 & 25.13 & 22.93 & 32.10 & 26.43 & 51.05 & 31.06 & 25.22\\
 & SLEB & 3.05 & 2.36 & 69.37 & 27.21 & 25.14 & 31.80 & 25.93 & 52.82 & 31.49 & 22.70\\
 & ShortGPT & 3.34 & 2.54 & 66.38 & 29.75 & 34.81 & 29.20 & 24.26 & 47.77 & 32.94 & 28.36  \\
 & 2SSP & 2.90 & 2.00 & 69.70 & 26.48 & 46.68 & 30.40 & 28.88 & 55.43 & 34.90 & 28.67\\
 &  \pruner & 3.09 & 2.16 & 68.77 & 26.92 & 49.88 & 29.40 & 26.30 & 51.31 & 32.08 & 34.56\\
 \hline
\end{tabular}%
}

\vspace{0.3em}
\textit{(continued)}
\vspace{0.3em}

\resizebox{\textwidth}{!}{%
\begin{tabular}{|cc|ccccc|ccc|cc|}
\hline
\multirow{3}{*}{\textbf{CR}} & \multirow{3}{*}{\textbf{Method}} & \multicolumn{5}{c}{NLU \& NLI} & \multicolumn{3}{|c|}{Safety, Bias \& Ethics} & \multirow{3}{*}{\textbf{Avg RP (\%)}} & \multirow{3}{*}{\textbf{Std RP (\%)}}\\
\cline{3-10}
& & \textbf{BLIMP} & \textbf{BoolQ} & \textbf{Lambada} & \textbf{Winogrande} & \textbf{CoQA} & \textbf{Winogender} & \textbf{TruthfulQA} & \textbf{Moral Stories} & & \\
& & (Acc $\uparrow$) & (Acc $\uparrow$) & (Acc $\uparrow$) & (Acc $\uparrow$) & (Acc $\uparrow$) & (Acc $\uparrow$) & (Acc $\uparrow$) & (Acc $\uparrow$) & ($\uparrow$) & ($\downarrow$) \\
\hline
\multirow{1}{*}{-} & Base & 83.42 & 83.91 & 73.08 & 73.40 & 78.80 & 62.50 & 54.00 & 50.17 & - & -\\
\cdashline{1-12}
\multirow{7}{*}{25\%} 
 & ReplaceMe & 82.67 & 82.66 & 64.93 & 70.64 & 78.10 & 57.64 & 53.74 & 52.99 & \textbf{88.90} & 10.50 \\
 & SliceGPT & 79.84 & 71.96 & 37.96 & 55.88 & 66.80 & 49.58 & 44.43 & 54.48 & 72.49 & 17.30  \\
 & LLM-Pruner & 83.66 & 55.14 & 51.80 & 59.67 & 61.90 & 54.17 & 44.14 & 54.68 & 75.11 & 17.70  \\
 & SLEB & 84.02 & 48.87 & 61.32 & 63.69 & 72.00 & 54.86 & 41.88 & 55.48 & 78.45 & 16.80  \\
 & ShortGPT & 81.75 & 83.21 & 62.49 & 70.09 & 76.30 & 59.86 & 52.90 & 53.26 & 88.47 & 11.20  \\
 & 2SSP & 85.42 & 63.27 & 64.68 & 65.67 & 78.90 & 57.50 & 43.45 & 53.31 & 83.19 & 13.90\\
 &  \pruner & 81.89 & 80.85 & 66.16 & 67.88 & 75.30 & 58.89 & 50.60 & 52.42 & 87.34 & \textbf{10.00}\\
\hline
\multirow{7}{*}{35\%} 
 & ReplaceMe & 82.47 & 57.13 & 36.66 & 51.70 & 24.80 & 51.53 & 41.93 & 55.28 & 65.16 & 22.60 \\
 & SliceGPT & 78.30 & 64.37 & 29.05 & 54.78 & 49.60 & 51.81 & 48.21 & 53.68 & 68.08 & 18.60 \\
 & LLM-Pruner & 83.53 & 55.14 & 42.67 & 58.17 & 53.00 & 52.78 & 43.38 & 54.69 & 69.93 & 19.40  \\
 & SLEB & 82.75 & 41.96 & 51.60 & 57.46 & 55.60 & 55.00 & 43.35 & 55.52 & 70.58 & 20.10 \\
 & ShortGPT & 79.03 & 77.46 & 53.66 & 67.72 & 71.70 &60.14 & 49.75 & 54.12 & 76.98 & 18.70\\
 & 2SSP & 84.86 & 61.50 & 57.73 & 62.43 & 73.60 & 56.11 & 39.94 & 54.51 & 77.26 & 15.80\\
 &  \pruner & 82.43 & 69.33 & 57.46 & 62.51 & 69.80 & 55.97 & 41.87 & 55.01 & \textbf{77.26} & \textbf{14.80}\\
 \hline
\end{tabular}%
}
\caption{Task-specific pruning results for different pruning methods on Llama-3.1-8B-Instruct across various compression ratios (CR) with RFT.}
\label{tab:main_results_llama_full_rft}
\end{table*}

\begin{table*}[!htb]
\centering
\resizebox{\textwidth}{!}{%
\begin{tabular}{|cc|cc|ccc|ccccc|}
\hline
\multirow{3}{*}{\textbf{CR}} & \multirow{3}{*}{\textbf{Method}} & \multicolumn{2}{c}{Generative} & \multicolumn{3}{|c|}{World Understanding} & \multicolumn{5}{c|}{Domain-Specific}\\
\cline{3-12}
& & \textbf{Wikitext} & \textbf{Lambada} & \textbf{PIQA} & \textbf{PROST} & \textbf{CommonsenseQA} & \textbf{OpenbookQA} & \textbf{MathQA} & \textbf{ARC-Easy} & \textbf{ARC-Challenge} & \textbf{MedQA} \\
 & & (Log-PPL $\downarrow$) & (Log-PPL $\downarrow$) & (Acc $\uparrow$) & (Acc $\uparrow$) & (Acc $\uparrow$) & (Acc $\uparrow$) & (Acc $\uparrow$) & (Acc $\uparrow$) & (Acc $\uparrow$) & (Acc $\uparrow$) \\
\hline
\multirow{1}{*}{-} & Base & 2.50 & 1.52 & 77.53 & 42.90 & 78.54 & 41.60 & 49.75 & 81.06 & 56.23 & 64.34 \\
\cdashline{1-12}
\multirow{7}{*}{25\%} 
 & ReplaceMe & 4.11 & 4.64 & 62.30 & 28.13 & 21.79 & 31.60 & 23.15 & 42.59 & 25.51 & 28.20\\
 & SliceGPT & 13.03 & 18.41 & 51.47 & 29.60 & 20.97 & 25.60 & 19.43 & 26.31 & 26.20 & 25.69\\
 & LLM-Pruner & 3.32 & 3.09 & 64.26 & 28.46 & 53.48 & 28.80 & 28.41 & 46.59 & 25.68 & 29.77\\
 & SLEB & 3.84 & 3.66 & 66.76 & 28.72 & 33.84 & 31.80 & 25.80 & 59.72 & 37.17 & 30.17\\
 & ShortGPT & 6.73 & 11.23 & 62.08 & 29.55 & 39.64 & 29.60 & 21.94 & 41.92 & 31.14 & 40.85\\
 & 2SSP & 5.12 & 7.62 & 55.50 & 30.68 & 18.18 & 27.80 & 21.11 & 32.41 & 23.29 & 26.94\\
 &  \pruner & 3.55 & 3.75 &63.66 & 30.04 & 58.72 & 30.60 & 26.57 & 52.44 & 34.81 & 36.61\\
\hline
\multirow{7}{*}{35\%} 
 & ReplaceMe & 6.12 & 5.87 & 63.06 & 30.67 & 19.17 & 32.00 & 23.05 & 48.49 & 29.35 & 28.59\\
& SliceGPT & 13.42 & 18.85 & 51.52 & 29.55 & 19.41 & 26.00 & 20.24 & 26.52 & 24.92 & 26.24\\
 & LLM-Pruner & 4.55 & 5.95 & 56.45 & 28.42 & 32.84 & 28.00 & 22.21 & 32.41 & 23.12 & 26.63\\
 & SLEB & 4.90 & 4.99 & 60.94 & 32.75 & 21.70 & 29.60 & 23.75 & 49.50 & 28.24 & 28.67\\
 & ShortGPT &  8.98 & 12.36 & 58.98 & 30.19 & 47.50 & 32.00 & 22.71 & 33.33 & 29.44 & 37.31\\
 & 2SSP & 5.63 & 9.21 & 52.61 & 32.42 & 19.08 & 26.00 & 21.94 & 30.22 & 23.72 & 27.73\\
&  \pruner & 4.45 & 5.77 & 59.68 & 32.15 & 40.30 & 27.20 & 22.88 & 42.72 & 27.05 & 32.29\\
 \hline
\end{tabular}%
}

\vspace{0.3em}
\textit{(continued)}
\vspace{0.3em}

\resizebox{\textwidth}{!}{%
\begin{tabular}{|cc|ccccc|ccc|cc|}
\hline
\multirow{3}{*}{\textbf{CR}} & \multirow{3}{*}{\textbf{Method}} & \multicolumn{5}{c}{NLU \& NLI} & \multicolumn{3}{|c|}{Safety, Bias \& Ethics} & \multirow{3}{*}{\textbf{Avg RP (\%)}} & \multirow{3}{*}{\textbf{Std RP (\%)}}\\
\cline{3-10}
& & \textbf{BLIMP} & \textbf{BoolQ} & \textbf{Lambada} & \textbf{Winogrande} & \textbf{CoQA} & \textbf{Winogender} & \textbf{TruthfulQA} & \textbf{Moral Stories} & & \\
& & (Acc $\uparrow$) & (Acc $\uparrow$) & (Acc $\uparrow$) & (Acc $\uparrow$) & (Acc $\uparrow$) & (Acc $\uparrow$) & (Acc $\uparrow$) & (Acc $\uparrow$) & ($\uparrow$) & ($\downarrow$) \\
\hline
\multirow{1}{*}{-} & Base & 80.72 & 86.54 & 64.10 & 67.72 & 81.90 & 62.22 & 54.39 & 51.08 & - & -\\
\cdashline{1-12}
\multirow{7}{*}{25\%} 
 & ReplaceMe & 78.15 & 40.31 & 24.65 & 51.38 & 12.70 & 49.44 & 45.44 & 56.53 & 59.92 & 23.80 \\
 & SliceGPT & 57.66 & 40.43 & 7.80 & 48.15 & 1.00 & 49.17 & 48.50 & 55.35 & 48.68 & 27.50\\
 & LLM-Pruner & 79.78 & 72.39 & 41.39 & 55.17 & 70.70 & 54.72 & 44.77 & 53.48 & 72.64 & 15.90\\
 & SLEB & 74.04 & 66.48 & 30.04 & 54.93 & 18.70 & 53.06 & 45.73 & 55.67 & 67.53 & 20.20\\
 & ShortGPT & 70.86 & 68.10 & 10.77 & 57.06 & 28.10 & 58.75 & 49.17 & 57.13 & 63.03 & 22.90  \\
 & 2SSP & 61.04 & 41.44 & 11.97 & 49.96 & 35.50 & 50.56 & 50.61 & 55.41 & 56.09 & 23.10\\
 &  \pruner & 74.03 & 69.57 & 33.34 & 57.54 & 70.70 & 59.58 & 50.22 & 54.90 & \textbf{74.41} & \textbf{15.00}\\
\hline
\multirow{7}{*}{35\%} 
 & ReplaceMe & 80.35 & 48.23 & 17.45 & 51.46 & 6.40 & 49.86 & 47.80 & 56.76 & 59.40 & 25.90\\
 & SliceGPT & 55.95 & 40.67 & 5.80 & 49.88 & 1.10 & 47.50 & 49.78 & 55.68 & 48.67 & 27.70\\
 & LLM-Pruner & 76.43 & 61.90 & 13.12 & 52.49 & 46.30 & 51.67 & 44.15 & 56.44 & 60.63 & 21.90\\
 & SLEB & 70.73 & 50.25 & 19.91 & 50.28 & 3.80 & 50.69 & 46.63 & 57.33 & 59.67 & 24.70\\
 & ShortGPT & 67.94 & 62.17 & 13.60 & 53.51 & 6.70 & 55.56 & 49.73 & 56.78 & 58.69 & 27.10\\
& 2SSP & 59.32 & 42.69 & 6.60 & 50.36 & 24.20 & 49.03 & 49.35 & 56.35 & 54.15 & 24.40\\
 &  \pruner & 70.96 & 58.59 & 14.38 & 53.59 & 45.90 & 53.19 & 48.43 & 56.65 & \textbf{63.71} & \textbf{20.10}\\
 \hline
\end{tabular}%
}
\caption{Task-specific pruning results for different pruning methods on Qwen3-8B across various compression ratios (CR) without RFT.}
\label{tab:main_results_qwen_full_norft}
\end{table*}

\begin{table*}[!htb]
\centering
\resizebox{\textwidth}{!}{%
\begin{tabular}{|cc|cc|ccc|ccccc|}
\hline
\multirow{3}{*}{\textbf{CR}} & \multirow{3}{*}{\textbf{Method}} & \multicolumn{2}{c}{Generative} & \multicolumn{3}{|c|}{World Understanding} & \multicolumn{5}{c|}{Domain-Specific}\\
\cline{3-12}
& & \textbf{Wikitext} & \textbf{Lambada} & \textbf{PIQA} & \textbf{PROST} & \textbf{CommonsenseQA} & \textbf{OpenbookQA} & \textbf{MathQA} & \textbf{ARC-Easy} & \textbf{ARC-Challenge} & \textbf{MedQA} \\
 & & (Log-PPL $\downarrow$) & (Log-PPL $\downarrow$) & (Acc $\uparrow$) & (Acc $\uparrow$) & (Acc $\uparrow$) & (Acc $\uparrow$) & (Acc $\uparrow$) & (Acc $\uparrow$) & (Acc $\uparrow$) & (Acc $\uparrow$) \\
\hline
\multirow{1}{*}{-} & Base & 2.50 & 1.52 & 77.53 & 42.90 & 78.54 & 41.60 & 49.75 & 81.06 & 56.23 & 64.34 \\
\cdashline{1-12}
\multirow{7}{*}{25\%} 
 & ReplaceMe & 2.93 & 2.60 & 70.46 & 27.00 & 34.48 & 36.20 & 27.77 & 57.03 & 34.73 & 28.44 \\
 & SliceGPT & 8.59 & 12.92 & 51.20 & 32.42 & 19.82 & 25.80 & 20.07 & 27.65 & 25.09 & 26.63 \\
 & LLM-Pruner & 2.83 & 2.09 & 70.02 & 28.14 & 62.16 & 34.00 & 31.66 & 55.81 & 33.96 & 30.95 \\
 & SLEB & 2.98 & 2.24 & 69.80 & 30.39 & 51.02 & 34.40 & 29.08 & 55.85 & 37.37 & 35.19\\
 & ShortGPT & 3.44 & 2.85 & 67.03 & 32.18 & 68.39 & 30.20 & 25.90 & 52.36 & 33.96 & 54.28\\
 & 2SSP & 2.82 & 2.04 & 69.75 & 32.21 & 68.96 & 32.20 & 26.73 & 61.36 & 38.14 & 35.98\\
 &  \pruner & 3.04 & 2.41 & 68.01 & 32.29 & 67.49 & 31.40 & 31.09 & 62.88 & 38.99 & 42.42\\
\hline
\multirow{7}{*}{35\%} 
 & ReplaceMe & 4.15 & 4.25 & 61.86 & 29.81 & 22.69 & 28.40 & 22.14 & 39.94 & 27.73 & 26.08\\
 & SliceGPT & 8.52 & 13.13 & 51.36 & 32.16 & 19.66 & 24.60 & 19.46 & 26.98 & 24.06 & 27.81\\
 & LLM-Pruner & 3.30 & 3.63 & 63.87 & 27.82 & 44.06 & 29.00 & 23.45 & 51.85 & 28.75 & 26.24\\
 & SLEB & 3.38 & 2.83 & 64.91 & 30.82 & 39.72 & 32.00 & 25.93 & 52.82 & 32.68 & 28.59\\
 & ShortGPT & 3.80 & 4.17 & 64.04 & 31.85 & 54.87 & 31.00 & 23.75 & 45.96 & 29.95 & 43.44\\
 & 2SSP & 3.52 & 3.59 & 64.04 & 28.23 & 46.85 & 29.20 & 23.45 & 47.14 & 30.29 & 29.62\\
 &  \pruner & 3.36 & 3.06 & 65.29 & 31.45 & 55.69 & 29.20 & 26.20 & 53.66 & 31.66 & 36.53\\
 \hline
\end{tabular}%
}

\vspace{0.3em}
\textit{(continued)}
\vspace{0.3em}

\resizebox{\textwidth}{!}{%
\begin{tabular}{|cc|ccccc|ccc|cc|}
\hline
\multirow{3}{*}{\textbf{CR}} & \multirow{3}{*}{\textbf{Method}} & \multicolumn{5}{c}{NLU \& NLI} & \multicolumn{3}{|c|}{Safety, Bias \& Ethics} & \multirow{3}{*}{\textbf{Avg RP (\%)}} & \multirow{3}{*}{\textbf{Std RP (\%)}}\\
\cline{3-10}
& & \textbf{BLIMP} & \textbf{BoolQ} & \textbf{Lambada} & \textbf{Winogrande} & \textbf{CoQA} & \textbf{Winogender} & \textbf{TruthfulQA} & \textbf{Moral Stories} & & \\
& & (Acc $\uparrow$) & (Acc $\uparrow$) & (Acc $\uparrow$) & (Acc $\uparrow$) & (Acc $\uparrow$) & (Acc $\uparrow$) & (Acc $\uparrow$) & (Acc $\uparrow$) & ($\uparrow$) & ($\downarrow$) \\
\hline
\multirow{1}{*}{-} & Base & 80.72 & 86.54 & 64.10 & 67.72 & 81.90 & 62.22 & 54.39 & 51.08 & - & -\\
\cdashline{1-12}
\multirow{7}{*}{25\%} 
 & ReplaceMe & 82.84 & 41.10 & 42.81 & 56.43 & 54.40 & 52.36 & 44.35 & 55.28 & 72.30 & 18.60\\
 & SliceGPT & 57.22 & 38.47 & 0.00 & 50.91 & 1.10 & 47.22 & 49.76 & 57.07 & 50.07 & 30.00\\
 & LLM-Pruner & 82.69 & 59.02 & 55.02 & 62.12 & 76.40 & 55.28 & 47.07 & 54.25 & 80.11 & 15.00\\
 & SLEB & 79.76 & 60.40 & 54.05 & 58.71 & 69.60 & 56.25 & 48.93 & 54.18 & 78.89 & 13.80\\
 & ShortGPT & 79.28 & 80.06 & 47.70 & 63.46 & 69.40 & 59.31 & 48.38 & 54.73 & 80.21 & 15.20\\
 & 2SSP & 82.76 & 63.39 & 57.99 & 64.80 & 77.80 & 57.50 & 49.79 & 52.11 & 82.77 & 13.90\\
 &  \pruner & 80.85 & 77.19 & 51.12 & 63.22 & 74.20 & 58.61 & 47.99 & 54.51 & \textbf{82.64} & \textbf{12.20}\\
\hline
\multirow{7}{*}{35\%} 
 & ReplaceMe & 76.07 & 53.27 & 32.82 & 54.78 & 44.40 & 56.67 & 44.30 & 56.82 & 64.00 & 21.60\\
 & SliceGPT & 58.83 & 38.14 & 0.00 & 51.38 & 1.00 & 50.69 & 49.18 & 57.44 & 50.17 & 30.40 \\
 & LLM-Pruner & 81.68 & 48.84 & 31.36 & 53.51 & 59.70 & 53.33 & 44.85 & 55.98 & 68.34 & 19.10\\
 & SLEB & 79.03 & 43.98 & 44.17 & 54.54 & 46.10 & 53.61 & 48.66 & 55.02 & 70.47 & 17.50\\
 & ShortGPT & 77.52 & 68.01 & 32.02 & 58.96 & 61.20 & 60.00 & 43.61 & 55.73 & 72.27 & 18.30\\
& 2SSP & 73.65 & 53.82 & 35.15 & 56.20 & 64.60 & 53.33 & 46.23 & 55.47 & 69.23 & 17.50\\
 &  \pruner & 81.71 & 72.20 & 41.51 & 57.93 & 63.40 & 55.00 & 47.49 & 55.95 & \textbf{75.13} & \textbf{15.90}\\
 \hline
\end{tabular}%
}
\caption{Task-specific pruning results for different pruning methods on Qwen3-8B across various compression ratios (CR) with RFT.}
\label{tab:main_results_qwen_full_rft}
\end{table*}

\begin{table*}[!htb]
\centering
\resizebox{\textwidth}{!}{%
\begin{tabular}{|cc|cc|ccc|ccccc|}
\hline
\multirow{3}{*}{\textbf{CR}} & \multirow{3}{*}{\textbf{Method}} & \multicolumn{2}{c}{Generative} & \multicolumn{3}{|c|}{World Understanding} & \multicolumn{5}{c|}{Domain-Specific}\\
\cline{3-12}
& & \textbf{Wikitext} & \textbf{Lambada} & \textbf{PIQA} & \textbf{PROST} & \textbf{CommonsenseQA} & \textbf{OpenbookQA} & \textbf{MathQA} & \textbf{ARC-Easy} & \textbf{ARC-Challenge} & \textbf{MedQA} \\
 & & (Log-PPL $\downarrow$) & (Log-PPL $\downarrow$) & (Acc $\uparrow$) & (Acc $\uparrow$) & (Acc $\uparrow$) & (Acc $\uparrow$) & (Acc $\uparrow$) & (Acc $\uparrow$) & (Acc $\uparrow$) & (Acc $\uparrow$)\\
\hline
\multirow{1}{*}{-} & Base & 2.05 & 1.25 & 81.34 & 54.46 & 74.53 & 45.20 & 47.07 & 72.81 & 56.06 & 62.53 \\
\cdashline{1-12}
\multirow{3}{*}{35\%} 
 & ShortGPT & 3.38 & 3.32 & 68.83 & 54.65 & 39.07 & 36.40 & 30.55 & 60.14 & 41.55 & 31.58\\
 & 2SSP & 2.77 & 3.38 & 65.51 & 34.81 & 42.51 & 28.00 & 24.99 & 50.46 & 32.34 & 30.95\\
 &  \pruner & 2.77 & 2.33 & 73.94 & 34.00 & 66.26 & 37.60 & 32.63 & 66.08 & 42.58 & 41.79\\
\hline
\end{tabular}%
}

\vspace{0.3em}
\textit{(continued)}
\vspace{0.3em}

\resizebox{\textwidth}{!}{%
\begin{tabular}{|cc|ccccc|ccc|cc|}
\hline
\multirow{3}{*}{\textbf{CR}} & \multirow{3}{*}{\textbf{Method}} & \multicolumn{5}{c}{NLU \& NLI} & \multicolumn{3}{|c|}{Safety, Bias \& Ethics} & \multirow{3}{*}{\textbf{Avg RP (\%)}} & \multirow{3}{*}{\textbf{Std RP (\%)}}\\
\cline{3-10}
& & \textbf{BLIMP} & \textbf{BoolQ} & \textbf{Lambada} & \textbf{Winogrande} & \textbf{CoQA} & \textbf{Winogender} & \textbf{TruthfulQA} & \textbf{Moral Stories} & & \\
& & (Acc $\uparrow$) & (Acc $\uparrow$) & (Acc $\uparrow$) & (Acc $\uparrow$) & (Acc $\uparrow$) & (Acc $\uparrow$) & (Acc $\uparrow$) & (Acc $\uparrow$) & ($\uparrow$) & ($\downarrow$) \\
\hline
\multirow{1}{*}{-} & Base & 81.69 & 86.18 & 72.31 & 76.56 & 82.70 & 71.67 &59.34 & 47.38 & - & -\\
\cdashline{1-12}
\multirow{3}{*}{35\%} 
 & ShortGPT & 78.28 & 65.05 & 41.06 & 71.67 & 70.90 & 68.61 & 49.89 & 53.83 & 77.18 & 16.60\\
 & 2SSP & 80.41 & 60.55 & 42.97 & 54.85 & 74.50 & 52.22 & 48.24 & 54.39 & 70.16 & 16.20\\
 &  \pruner & 82.45 & 75.44 & 50.51 & 63.38 & 78.70 & 62.50 & 44.65 & 53.95 & \textbf{81.58} & \textbf{12.60}\\
\hline
\end{tabular}%
}
\caption{Task-specific pruning results for different pruning methods on Phi-4 for a CR of 35\% with RFT.}
\label{tab:main_results_phi_full_rft}
\end{table*}
\begin{table*}[!htb]
\centering
\resizebox{\textwidth}{!}{%
\begin{tabular}{|cc|cc|ccc|ccccc|}
\hline
\multirow{3}{*}{\textbf{CR}} & \multirow{3}{*}{\textbf{Method}} & \multicolumn{2}{c}{Generative} & \multicolumn{3}{|c|}{World Understanding} & \multicolumn{5}{c|}{Domain-Specific}\\
\cline{3-12}
& & \textbf{Wikitext} & \textbf{Lambada} & \textbf{PIQA} & \textbf{PROST} & \textbf{CommonsenseQA} & \textbf{OpenbookQA} & \textbf{MathQA} & \textbf{ARC-Easy} & \textbf{ARC-Challenge} & \textbf{MedQA} \\
 & & (Log-PPL $\downarrow$) & (Log-PPL $\downarrow$) & (Acc $\uparrow$) & (Acc $\uparrow$) & (Acc $\uparrow$) & (Acc $\uparrow$) & (Acc $\uparrow$) & (Acc $\uparrow$) & (Acc $\uparrow$) & (Acc $\uparrow$) \\
\hline
\multirow{1}{*}{-} & Base & 2.59 & 1.86 & 78.62 & 45.28 & 73.06 & 40.60 & 47.84 & 77.53 & 51.88 & 62.53 \\
\cdashline{1-12}
\multirow{3}{*}{35\%} 
 & ShortGPT & 3.17 & 2.95 & 69.64 & 30.56 & 38.33 & 32.00 & 33.87 & 60.40 & 33.79 & 30.09\\
 & 2SSP & 3.26 & 2.73 & 73.12 & 33.43 & 61.92 & 35.20 & 31.99 & 65.28 & 40.27 & 39.91\\
 &  \pruner & 3.04 & 2.38 & 72.96 & 32.39 & 61.51 & 37.00 & 35.75 & 66.16 & 42.41 & 45.09\\
\hline
\end{tabular}%
}

\vspace{0.3em}
\textit{(continued)}
\vspace{0.3em}

\resizebox{\textwidth}{!}{%
\begin{tabular}{|cc|ccccc|ccc|cc|}
\hline
\multirow{3}{*}{\textbf{CR}} & \multirow{3}{*}{\textbf{Method}} & \multicolumn{5}{c}{NLU \& NLI} & \multicolumn{3}{|c|}{Safety, Bias \& Ethics} & \multirow{3}{*}{\textbf{Avg RP (\%)}} & \multirow{3}{*}{\textbf{Std RP (\%)}}\\
\cline{3-10}
& & \textbf{BLIMP} & \textbf{BoolQ} & \textbf{Lambada} & \textbf{Winogrande} & \textbf{CoQA} & \textbf{Winogender} & \textbf{TruthfulQA} & \textbf{Moral Stories} & & \\
& & (Acc $\uparrow$) & (Acc $\uparrow$) & (Acc $\uparrow$) & (Acc $\uparrow$) & (Acc $\uparrow$) & (Acc $\uparrow$) & (Acc $\uparrow$) & (Acc $\uparrow$) & ($\uparrow$) & ($\downarrow$) \\
\hline
\multirow{1}{*}{-} & Base & 81.00 & 86.42 & 60.76 & 67.88 & 80.80 & 62.22 & 59.13 & 53.13 & - & -\\
\cdashline{1-12}
\multirow{3}{*}{35\%} 
 & ShortGPT & 77.70 & 54.43 & 40.07 & 53.51 & 0.10 & 53.61 & 46.73 & 56.38 & 70.51 & 22.10\\
 & 2SSP & 80.60 & 75.11 & 46.87 & 63.30 & 79.10 & 61.53 & 53.83 & 53.79 & 84.68 & 11.40\\
 &  \pruner & 80.44 & 79.82 & 52.05 & 61.72 & 75.70 & 58.89 & 51.79 & 55.72 & \textbf{86.99} & \textbf{8.80}\\
\hline
\end{tabular}%
}
\caption{Task-specific pruning results for different pruning methods on GPT-OSS-20B for a CR of 35\% with RFT.}
\label{tab:main_results_gpt_full_rft}
\end{table*}
\subsection{Task-Wise Performance}
\label{appendix:task-wise-performance}
Tables~\ref{tab:main_results_llama_full_norft}, \ref{tab:main_results_llama_full_rft}, \ref{tab:main_results_qwen_full_norft}, and \ref{tab:main_results_qwen_full_rft} provide the detailed task-wise results for each method on Llama-3.1-8B-Instruct and Qwen3-8B with and without RFT at compression ratios of 25\% and 35\%. \pruner\ demonstrates noteworthy superiority over its baselines across the different testing configurations, repeatedly outperforming them on average and exhibiting the highest per-task consistency among them. Even in the instance where \pruner\ lags slightly behind a baseline in average performance (Table~\ref{tab:main_results_llama_full_rft}, 25\% CR), it still retains the lowest per-task standard deviation of the group, providing a strong testimony to its excellent robustness to distributional bias.   

We also provide the task-wise results for Phi-4 and GPT-OSS-20B in Tables~\ref{tab:main_results_phi_full_rft} and \ref{tab:main_results_gpt_full_rft}, respectively, where \pruner\ surpasses ShortGPT and 2SSP on Avg RP by notable margins. In addition to delivering excellent performance, \pruner\ also maintains a considerably lower Std RP compared to them, indicating its impressive ability to adapt to unique architectures.     

\subsection{Sensitivity to Calibration Data}
\label{appendix:calibration-results}
\begin{table*}[!htb]
\centering
\resizebox{\textwidth}{!}{%
\begin{tabular}{|cc|cc|ccc|ccccc|}
\hline
\multirow{3}{*}{\textbf{\# Calibration Samples}} & \multirow{3}{*}{\textbf{Method}} & \multicolumn{2}{c}{Generative} & \multicolumn{3}{|c|}{World Understanding} & \multicolumn{5}{c|}{Domain-Specific}\\
\cline{3-12}
& & \textbf{Wikitext} & \textbf{Lambada} & \textbf{PIQA} & \textbf{PROST} & \textbf{CommonsenseQA} & \textbf{OpenbookQA} & \textbf{MathQA} & \textbf{ARC-Easy} & \textbf{ARC-Challenge} & \textbf{MedQA} \\
 & & (Log-PPL $\downarrow$) & (Log-PPL $\downarrow$) & (Acc $\uparrow$) & (Acc $\uparrow$) & (Acc $\uparrow$) & (Acc $\uparrow$) & (Acc $\uparrow$) & (Acc $\uparrow$) & (Acc $\uparrow$) & (Acc $\uparrow$) \\
\hline
\multirow{7}{*}{50} 
 & ReplaceMe & 2.93 & 2.60 & 70.46 & 27.00 & 34.48 & 36.20 & 27.77 & 57.03 & 34.73 & 28.44 \\
 & SliceGPT & 8.59 & 12.92 & 51.20 & 32.42 & 19.82 & 25.80 & 20.07 & 27.65 & 25.09 & 26.63 \\
 & LLM-Pruner & 2.83 & 2.09 & 70.02 & 28.14 & 62.16 & 34.00 & 31.66 & 55.81 & 33.96 & 30.95 \\
 & SLEB & 2.98 & 2.24 & 69.80 & 30.39 & 51.02 & 34.40 & 29.08 & 55.85 & 37.37 & 35.19\\
 & ShortGPT & 3.44 & 2.85 & 67.03 & 32.18 & 68.39 & 30.20 & 25.90 & 52.36 & 33.96 & 54.28\\
 & 2SSP & 2.82 & 2.04 & 69.75 & 32.21 & 68.96 & 32.20 & 26.73 & 61.36 & 38.14 & 35.98\\
 & \pruner & 3.04 & 2.41 & 68.01 & 32.29 & 67.49 & 31.40 & 31.09 & 62.88 & 38.99 & 42.42\\
 \hline
\multirow{7}{*}{250} 
 & ReplaceMe & 2.91 & 2.54 & 71.00 & 27.04 & 33.99 & 36.00 & 27.97 & 57.74 & 36.52 & 28.12\\
 & SliceGPT & 8.39 & 13.10 & 50.11 & 34.69 & 19.66 & 28.00 & 20.30 & 27.69 & 24.40 & 27.34 \\
 & LLM-Pruner & 2.86 & 2.29 & 70.24 & 30.17 & 60.61 & 32.80 & 31.66 & 63.34 & 38.06 & 30.09\\
 & SLEB & 3.00 & 2.55 & 72.04 & 28.57 & 31.12 & 39.00 & 29.01 & 59.51 & 37.12 & 29.22\\
 & ShortGPT & 3.41 & 2.79 & 66.81 & 34.26 & 72.97 & 29.80 & 26.16 & 52.27 & 35.50 & 56.01\\
 & 2SSP & 3.08 & 2.95 & 65.83 & 32.11 & 57.08 & 29.80 & 24.29 & 50.84 & 31.91 & 31.27\\
 &  \pruner & 3.01 & 2.50 & 69.59 & 32.43 & 61.02 & 32.80 & 28.74 & 62.46 & 38.48 & 38.57\\
\hline
\multirow{7}{*}{1000} 
 & ReplaceMe & 2.91 & 2.54 & 71.00 & 27.04 & 33.99 & 36.00 & 27.97 & 57.74 & 36.52 & 28.12\\
 & SliceGPT & 7.92 & 11.98 & 51.09 & 28.98 & 18.76 & 27.20 & 20.34 & 27.65 & 23.38 & 26.24\\
 & LLM-Pruner & 2.94 & 2.35 & 68.55 & 27.78 & 59.54 & 33.60 & 29.05 & 60.99 & 37.88 & 30.56\\
 & SLEB & 2.98 & 2.58 & 70.89 & 29.77 & 37.35 & 35.20 & 29.95 & 60.19 & 34.73 & 31.19\\
 & ShortGPT & 3.41 & 2.78 & 67.14 & 33.22 & 75.43 & 30.20 & 25.60 & 51.05 & 34.90 & 58.05\\
 & 2SSP & 2.97 & 3.02 & 67.57 & 30.68 & 60.69 & 29.00 & 24.26 & 53.45 & 32.85 & 29.93\\
 &  \pruner & 3.01 & 2.49 & 69.70 & 31.47 & 60.20 & 32.80 & 28.74 & 62.16 & 37.71 & 38.96\\
\hline
\end{tabular}%
}

\vspace{0.3em}
\textit{(continued)}
\vspace{0.3em}

\resizebox{\textwidth}{!}{%
\begin{tabular}{|cc|ccccc|ccc|cc|}
\hline
\multirow{3}{*}{\textbf{\# Calibration Samples}} & \multirow{3}{*}{\textbf{Method}} & \multicolumn{5}{c}{NLU \& NLI} & \multicolumn{3}{|c|}{Safety, Bias \& Ethics} & \multirow{3}{*}{\textbf{Avg RP (\%)}} & \multirow{3}{*}{\textbf{Std RP (\%)}}\\
\cline{3-10}
& & \textbf{BLIMP} & \textbf{BoolQ} & \textbf{Lambada} & \textbf{Winogrande} & \textbf{CoQA} & \textbf{Winogender} & \textbf{TruthfulQA} & \textbf{Moral Stories} & & \\
& & (Acc $\uparrow$) & (Acc $\uparrow$) & (Acc $\uparrow$) & (Acc $\uparrow$) & (Acc $\uparrow$) & (Acc $\uparrow$) & (Acc $\uparrow$) & (Acc $\uparrow$) & ($\uparrow$) & ($\downarrow$) \\
\hline
\multirow{7}{*}{50} 
 & ReplaceMe & 82.84 & 41.10 & 42.81 & 56.43 & 54.40 & 52.36 & 44.35 & 55.28 & 72.30 & 18.60\\
 & SliceGPT & 57.22 & 38.47 & 0.00 & 50.91 & 1.10 & 47.22 & 49.76 & 57.07 & 50.07 & 30.00\\
 & LLM-Pruner & 82.69 & 59.02 & 55.02 & 62.12 & 76.40 & 55.28 & 47.07 & 54.25 & 80.11 & 15.00\\
 & SLEB & 79.76 & 60.40 & 54.05 & 58.71 & 69.60 & 56.25 & 48.93 & 54.18 & 78.89 & 13.80\\
 & ShortGPT & 79.28 & 80.06 & 47.70 & 63.46 & 69.40 & 59.31 & 48.38 & 54.73 & 80.21 & 15.20\\
 & 2SSP & 82.76 & 63.39 & 57.99 & 64.80 & 77.80 & 57.50 & 49.79 & 52.11 & 82.77 & 13.90\\
 &  \pruner & 80.85 & 77.19 & 51.12 & 63.22 & 74.20 & 58.61 & 47.99 & 54.51 & \textbf{82.64} & \textbf{12.20}\\
\hline
\multirow{7}{*}{250} 
 & ReplaceMe & 83.41 & 44.04 & 44.87 & 56.59 & 54.20 & 52.50 & 44.71 & 55.03 & 73.08 & 18.00 \\
 & SliceGPT & 58.83 & 38.38 & 0.00 & 50.43 & 0.60 & 49.31 & 49.24 & 57.73 & 50.91 & 28.90  \\
 & LLM-Pruner & 82.49 & 42.72 & 52.20 & 62.35 & 77.10 & 56.11 & 45.88 & 53.89 & 79.26 & 15.60\\
 & SLEB & 81.46 & 58.87 & 45.49 & 57.93 & 43.10 & 53.61 & 42.78 & 55.74 & 73.99 & 18.20 \\
 & ShortGPT & 78.62 & 80.83 & 48.61 & 62.75 & 67.50 & 58.06 & 48.92 & 54.72 & 81.00 & 13.40 \\
 & 2SSP & 74.26 & 72.17 & 45.12 & 57.70 & 68.70 & 58.47 & 45.98 & 55.44 & 75.31 & 15.10\\
 &  \pruner & 80.29 & 76.70 & 47.70 & 62.98 & 73.60 & 59.17 & 47.58 & 53.98 & \textbf{81.28} & \textbf{12.60}\\
\hline
\multirow{7}{*}{1000} 
 & ReplaceMe & 83.41 & 44.04 & 44.87 & 56.59 & 54.20 & 52.50 & 44.71 & 55.03 &73.08 & 18.00\\
 & SliceGPT & 60.30 & 41.28 & 5.80 & 50.28 & 0.80 & 48.75 & 49.49 & 56.71 & 50.19 & 28.20\\
 & LLM-Pruner & 82.16 & 46.02 & 51.56 & 59.91 & 75.60 & 55.00 & 45.54 & 54.98 & 78.02 & 15.50\\
 & SLEB & 81.75 & 42.60 & 45.08 & 55.09 & 45.30 & 54.44 & 43.26 & 56.09 & 73.08 & 17.50\\
 & ShortGPT & 78.95 & 80.64 & 49.00 & 63.22 & 67.50 & 59.03 & 48.45 & 54.78 & 81.23 & 14.10\\
 & 2SSP & 74.77 & 55.72 & 44.15 & 58.49 & 67.10 & 51.94 & 47.81 & 56.85 & 74.26 & 15.40\\
 &  \pruner & 80.38 & 77.98 & 47.99 & 63.46 & 76.70 & 59.44 & 48.33 & 54.01 & \textbf{81.52} & \textbf{13.00}\\
\hline
\end{tabular}%
}
\caption{Detailed results for the impact of varying number of calibration samples selected from the Slim Orca dataset to identify model components to prune from Qwen3-8B. All results have been obtained after RFT.}
\label{tab:calibration-size-all-results}
\end{table*}
\begin{table*}[!htb]
\centering
\resizebox{\textwidth}{!}{%
\begin{tabular}{|cc|cc|ccc|ccccc|}
\hline
\multirow{3}{*}{\textbf{Calibration Dataset}} & \multirow{3}{*}{\textbf{Method}} & \multicolumn{2}{c}{Generative} & \multicolumn{3}{|c|}{World Understanding} & \multicolumn{5}{c|}{Domain-Specific}\\
\cline{3-12}
& & \textbf{Wikitext} & \textbf{Lambada} & \textbf{PIQA} & \textbf{PROST} & \textbf{CommonsenseQA} & \textbf{OpenbookQA} & \textbf{MathQA} & \textbf{ARC-Easy} & \textbf{ARC-Challenge} & \textbf{MedQA} \\
 & & (Log-PPL $\downarrow$) & (Log-PPL $\downarrow$) & (Acc $\uparrow$) & (Acc $\uparrow$) & (Acc $\uparrow$) & (Acc $\uparrow$) & (Acc $\uparrow$) & (Acc $\uparrow$) & (Acc $\uparrow$) & (Acc $\uparrow$) \\
\hline
\multirow{6}{*}{Slim Orca} 
 & ReplaceMe & 2.93 & 2.60 & 70.46 & 27.00 & 34.48 & 36.20 & 27.77 & 57.03 & 34.73 & 28.44 \\
 \multirow{6}{*}{\citep{SlimOrca}} & SliceGPT & 8.59 & 12.92 & 51.20 & 32.42 & 19.82 & 25.80 & 20.07 & 27.65 & 25.09 & 26.63 \\
 & LLM-Pruner & 2.83 & 2.09 & 70.02 & 28.14 & 62.16 & 34.00 & 31.66 & 55.81 & 33.96 & 30.95 \\
 & SLEB & 2.98 & 2.24 & 69.80 & 30.39 & 51.02 & 34.40 & 29.08 & 55.85 & 37.37 & 35.19\\
 & ShortGPT & 3.44 & 2.85 & 67.03 & 32.18 & 68.39 & 30.20 & 25.90 & 52.36 & 33.96 & 54.28\\
 & 2SSP & 2.82 & 2.04 & 69.75 & 32.21 & 68.96 & 32.20 & 26.73 & 61.36 & 38.14 & 35.98\\
 & \pruner & 3.04 & 2.41 & 68.01 & 32.29 & 67.49 & 31.40 & 31.09 & 62.88 & 38.99 & 42.42\\
 \hline
\multirow{6}{*}{Alpaca} 
 & ReplaceMe & 3.78 & 4.11 & 65.78 & 42.47 & 40.05 & 32.60 & 25.96 & 52.32 & 35.41 & 30.32\\
\multirow{6}{*}{\citep{alpaca}} & SliceGPT & 9.55 & 13.84 & 50.82 & 34.40 & 19.57 & 26.40 & 20.37 & 27.95 & 25.60 & 27.73\\
& LLM-Pruner & 3.00 & 2.73 & 68.50 & 29.68 & 51.84 & 32.60 & 30.52 & 61.62 & 36.78 & 32.05\\
 & SLEB & 3.51 & 3.27 & 65.45 & 32.86 & 41.77 & 29.40 & 29.08 & 52.02 & 32.59 & 33.46\\
 & ShortGPT & 3.29 & 2.56 & 67.57 & 35.32 & 72.56 & 33.80 & 26.97 & 55.43 & 37.20 & 52.55\\
 & 2SSP & 2.94 & 2.49 & 68.23 & 31.08 & 61.02 & 31.00 & 28.91 & 61.07 & 38.74 & 36.61\\
 & \pruner & 3.01 & 2.49 & 70.19 & 32.86 & 61.34 & 32.60 & 30.32 & 61.95 & 39.33 & 38.26\\
\hline
\multirow{6}{*}{C4} 
 & ReplaceMe & 3.78 & 4.11 & 65.78 & 42.47 & 40.05 & 32.60 & 25.96 & 52.32 & 35.41 & 30.32\\
\multirow{6}{*}{\citep{c4-dataset}} & SliceGPT & 9.24 & 13.91 & 48.37 & 34.01 & 19.57 & 26.20 & 20.54 & 27.69 & 25.00 & 27.73\\
 & LLM-Pruner & 3.00 & 2.72 & 67.41 & 28.78 & 53.15 & 33.80 & 30.25 & 61.62 & 36.35 & 30.48\\
 & SLEB & 3.21 & 2.57 & 69.48 & 36.92 & 55.61 & 34.60 & 31.39 & 62.00 & 39.76 & 33.78\\
 & ShortGPT & 3.63 & 3.68 & 66.05 & 43.07 & 74.45 & 31.20 & 26.57 & 52.65 & 34.04 & 51.93\\
 & 2SSP & 3.25 & 3.44 & 62.57 &26.57 & 29.81 & 28.40 & 23.99 & 46.17 & 29.35 & 27.34\\
& \pruner & 3.01 & 2.49 & 69.04 & 33.52 & 61.75 & 32.60 & 29.98 & 62.46 & 39.08 & 39.20\\
\hline
\end{tabular}%
}

\vspace{0.3em}
\textit{(continued)}
\vspace{0.3em}

\resizebox{\textwidth}{!}{%
\begin{tabular}{|cc|ccccc|ccc|cc|}
\hline
\multirow{3}{*}{\textbf{CR}} & \multirow{3}{*}{\textbf{Method}} & \multicolumn{5}{c}{NLU \& NLI} & \multicolumn{3}{|c|}{Safety, Bias \& Ethics} & \multirow{3}{*}{\textbf{Avg RP (\%)}} & \multirow{3}{*}{\textbf{Std RP (\%)}}\\
\cline{3-10}
& & \textbf{BLIMP} & \textbf{BoolQ} & \textbf{Lambada} & \textbf{Winogrande} & \textbf{CoQA} & \textbf{Winogender} & \textbf{TruthfulQA} & \textbf{Moral Stories} & & \\
& & (Acc $\uparrow$) & (Acc $\uparrow$) & (Acc $\uparrow$) & (Acc $\uparrow$) & (Acc $\uparrow$) & (Acc $\uparrow$) & (Acc $\uparrow$) & (Acc $\uparrow$) & ($\uparrow$) & ($\downarrow$) \\
\hline
\multirow{6}{*}{Slim Orca} & ReplaceMe & 82.84 & 41.10 & 42.81 & 56.43 & 54.40 & 52.36 & 44.35 & 55.28 & 72.30 & 18.60\\
\multirow{6}{*}{\citep{SlimOrca}} & SliceGPT & 57.22 & 38.47 & 0.00 & 50.91 & 1.10 & 47.22 & 49.76 & 57.07 & 50.07 & 30.00\\
 & LLM-Pruner & 82.69 & 59.02 & 55.02 & 62.12 & 76.40 & 55.28 & 47.07 & 54.25 & 80.11 & 15.00\\
 & SLEB & 79.76 & 60.40 & 54.05 & 58.71 & 69.60 & 56.25 & 48.93 & 54.18 & 78.89 & 13.80\\
 & ShortGPT & 79.28 & 80.06 & 47.70 & 63.46 & 69.40 & 59.31 & 48.38 & 54.73 & 80.21 & 15.20\\
 & 2SSP & 82.76 & 63.39 & 57.99 & 64.80 & 77.80 & 57.50 & 49.79 & 52.11 & \textbf{82.77} & 13.90\\
 & \pruner & 80.85 & 77.19 & 51.12 & 63.22 & 74.20 & 58.61 & 47.99 & 54.51 & 82.64 & \textbf{12.20}\\
\hline
 \multirow{6}{*}{Alpaca} & ReplaceMe & 78.37 & 83.39 & 34.16 & 63.30 & 68.80 & 59.31 & 50.26 & 54.57 & 75.66 & 18.70\\
 \multirow{6}{*}{\citep{alpaca}} & SliceGPT & 56.02 & 37.92 & 0.00 & 51.30 & 0.40 & 49.86 & 49.29 & 57.03 & 50.45 & 28.60\\
 & LLM-Pruner & 81.05 & 55.72 & 42.93 & 59.20 & 69.40 & 54.72 & 45.48 & 56.65 & 76.67 & 14.40\\
 & SLEB & 75.80 & 71.32 & 43.97 & 56.99 & 59.90 & 57.64 & 48.01 & 55.94 & 73.78 & 15.00\\
 & ShortGPT & 79.83 & 81.65 & 50.69 & 62.90 & 68.10 & 58.75 & 47.52 & 54.93 & \textbf{82.59} & 13.50\\
 & 2SSP & 82.84 & 52.75 & 49.23 & 62.12 & 71.80 & 55.42 & 48.19 & 54.91 & 79.06 & 14.30\\
 & \pruner & 80.61 & 77.34 & 48.34 & 63.38 & 73.70 & 59.72 & 47.54 & 54.10 & 81.79 & \textbf{13.10}\\
\hline

\multirow{6}{*}{C4} & ReplaceMe & 78.37 & 83.39 & 34.16 & 63.30 & 68.80 & 59.31 & 50.26 & 54.57 & 75.66 & 18.70\\
\multirow{6}{*}{\citep{c4-dataset}} & SliceGPT & 53.69 & 37.83 & 0.00 & 49.49 & 0.80 & 50.83 & 54.53 & 57.28 & 50.50 & 29.10\\
 & LLM-Pruner & 80.33 & 57.40 & 43.06 & 60.62 & 70.70 & 53.19 & 45.26 & 57.37 & 76.73 & 14.80\\
 & SLEB & 78.83 & 74.68 & 49.31 & 62.75 & 62.20 & 55.00 & 52.28 & 54.48 & 80.57 & \textbf{12.70}\\
 & ShortGPT & 78.70 & 81.25 & 38.31 & 65.04 & 70.20 & 59.58 & 49.26 & 54.58 & 80.63 & 15.30\\
 & 2SSP & 71.82 & 43.52 & 37.20 & 57.06 & 65.90 & 51.81 & 47.60 & 57.18 & 67.48 & 18.80\\
 & \pruner & 80.50 & 77.46 & 48.25 & 63.14 & 73.70 & 60.69 & 47.61 & 54.04 & \textbf{81.93} & 13.10\\
\hline
\end{tabular}%
}
\caption{Detailed results for impact of calibration data distribution on post-pruning performance of Qwen3-8B at a compression ratio of 25\%. Sample count is kept constant at 50. All results have been obtained after RFT.}
\label{tab:calibration-distribution-all-results}
\end{table*}
We analyze the effect of varying calibration-related factors, namely, dataset size and data distribution, on each method's performance. We provide these results in Tables~\ref{tab:calibration-size-all-results} and \ref{tab:calibration-distribution-all-results}, respectively. In line with the analysis presented in Section~\ref{sec:discussion}, \pruner\ demonstrates excellent robustness at varying calibration configurations where existing methods tend to deteriorate either in terms of average performance or per-task consistency.    

\subsection{Transferability of Importance Scores}
\label{appendix:transferability}
\begin{table*}[!htb]
\centering
\resizebox{\textwidth}{!}{%
\begin{tabular}{|cc|cc|ccc|ccccc|}
\hline
\multirow{3}{*}{\textbf{CR}} & \multirow{3}{*}{\textbf{Transfer Regime}} & \multicolumn{2}{c}{Generative} & \multicolumn{3}{|c|}{World Understanding} & \multicolumn{5}{c|}{Domain-Specific}\\
\cline{3-12}
& & \textbf{Wikitext} & \textbf{Lambada} & \textbf{PIQA} & \textbf{PROST} & \textbf{CommonsenseQA} & \textbf{OpenbookQA} & \textbf{MathQA} & \textbf{ARC-Easy} & \textbf{ARC-Challenge} & \textbf{MedQA} \\
 & & (Log-PPL $\downarrow$) & (Log-PPL $\downarrow$) & (Acc $\uparrow$) & (Acc $\uparrow$) & (Acc $\uparrow$) & (Acc $\uparrow$) & (Acc $\uparrow$) & (Acc $\uparrow$) & (Acc $\uparrow$) & (Acc $\uparrow$) \\
\hline
\multirow{3}{*}{25\%} 
 & $25\% \rightarrow 25\%$ (No transfer) & 3.55 & 3.75 &63.66 & 30.04 & 58.72 & 30.60 & 26.57 & 52.44 & 34.81 & 36.61\\
 & $35\% \rightarrow 25\%$ (High-to-low) & 3.55 & 3.76 & 63.44 & 29.93 & 58.40 & 30.60 & 26.80 & 52.57 & 35.07 & 35.59\\
 & $50\% \rightarrow 25\%$ (High-to-low) & 3.55 & 3.71 & 64.31 & 29.52 & 54.63 & 30.80 & 26.27 & 55.72 & 33.28 & 32.60\\
\hline
\multirow{3}{*}{35\%} 
 & $35\% \rightarrow 35\%$ (No transfer) & 4.41 & 5.00 & 65.83 & 29.55 & 28.09 & 28.60 & 24.32 & 46.84 & 33.11 & 28.04\\
 & $25\% \rightarrow 35\%$ (Low-to-high) & 4.38 & 5.76 & 60.34 & 31.49 & 36.36 & 25.40 & 23.18 & 40.95 & 26.20 & 30.72\\
 & $50\% \rightarrow 35\%$ (High-to-low) & 4.52 & 5.64 & 60.12 & 32.50 & 40.62 & 28.80 & 22.58 & 41.58 & 26.45 & 32.76\\
 \hline
 \multirow{3}{*}{50\%} 
 & $50\% \rightarrow 50\%$ (No transfer) & 7.74 & 15.39 & 55.44 & 32.61 & 21.05 & 28.80 & 20.17 & 31.52 & 25.94 & 27.57\\
 & $25\% \rightarrow 50\%$ (Low-to-high) & 7.46 & 11.65 & 55.82 & 31.47 & 19.25 & 27.00 & 19.30 & 31.86 & 23.98 & 27.18\\
 & $35\% \rightarrow 50\%$ (Low-to-high) & 8.43 & 14.50 & 55.99 & 33.95 & 20.48 & 25.40 & 20.54 & 32.96 & 23.98 & 28.28\\
 \hline
\end{tabular}%
}

\vspace{0.3em}
\textit{(continued)}
\vspace{0.3em}

\resizebox{\textwidth}{!}{%
\begin{tabular}{|cc|ccccc|ccc|cc|}
\hline
\multirow{3}{*}{\textbf{CR}} & \multirow{3}{*}{\textbf{Transfer Regime}} & \multicolumn{5}{c}{NLU \& NLI} & \multicolumn{3}{|c|}{Safety, Bias \& Ethics} & \multirow{3}{*}{\textbf{Avg RP (\%)}} & \multirow{3}{*}{\textbf{Std RP (\%)}}\\
\cline{3-10}
& & \textbf{BLIMP} & \textbf{BoolQ} & \textbf{Lambada} & \textbf{Winogrande} & \textbf{CoQA} & \textbf{Winogender} & \textbf{TruthfulQA} & \textbf{Moral Stories} & & \\
& & (Acc $\uparrow$) & (Acc $\uparrow$) & (Acc $\uparrow$) & (Acc $\uparrow$) & (Acc $\uparrow$) & (Acc $\uparrow$) & (Acc $\uparrow$) & (Acc $\uparrow$) & ($\uparrow$) & ($\downarrow$) \\
\hline
\multirow{3}{*}{25\%} 
 & $25\% \rightarrow 25\%$ (No transfer) & 74.03 & 69.57 & 33.34 & 57.54 & 70.70 & 59.58 & 50.22 & 54.90 & \textbf{74.41} & \textbf{15.00}\\
 & $35\% \rightarrow 25\%$ (High-to-low) & 73.90 & 69.76 & 33.28 & 58.09 & 68.90 & 59.44 & 50.19 & 55.03 & 74.24 & 17.00\\
 & $50\% \rightarrow 25\%$ (High-to-low) & 74.19 & 74.34 & 33.34 & 56.51 & 69.50 & 58.75 & 50.56 & 54.62 & 73.92 & 17.40\\
\hline
\multirow{3}{*}{35\%} 
 & $35\% \rightarrow 35\%$ (No transfer) & 70.96 & 58.59 & 14.38 & 53.59 & 45.90 & 53.19 & 48.43 & 56.65 & \textbf{63.71} & \textbf{20.10}\\
 & $25\% \rightarrow 35\%$ (Low-to-high) & 71.82 & 64.95 & 13.91 & 53.12 & 44.80 & 55.83 & 48.74 & 56.68 & 63.48 & 22.90\\
 & $50\% \rightarrow 35\%$ (High-to-low) & 70.95 & 62.97 & 15.82 & 53.67 & 33.80 & 58.33 & 47.72 & 56.38 & 63.83 & 22.90\\
 \hline
 \multirow{3}{*}{50\%} 
 & $50\% \rightarrow 50\%$ (No transfer) & 60.02 & 54.99 & 34.90 & 51.30 & 1.50 & 55.00 & 48.34 & 57.19 & \textbf{53.30} & \textbf{28.60}\\
 & $25\% \rightarrow 50\%$ (Low-to-high) & 61.88 & 38.10 & 2.72 & 50.43 & 0.30 & 49.58 & 48.42 & 57.92 & 51.46 & 30.00\\
 & $35\% \rightarrow 50\%$ (Low-to-high) & 62.77 & 46.33 & 1.52 & 49.01 & 0.10 & 53.61 & 49.78 & 57.73 & 52.45 & 30.90\\
 \hline
\end{tabular}%
}
\caption{Detailed results for \pruner\ under various transfer regimes wherein importance scores computed for a particular compression ratio are utilized to prune a model to a different capacity.}
\label{tab:transferability-detailed-results}
\end{table*}

\pruner\ demonstrates impressive flexibility by maintaining excellent performance even when utilizing heuristics that were not computed for the target compression ratio, as evident by the detailed results in Table~\ref{tab:transferability-detailed-results}. This characteristic enhances \pruner's practical utility by offering an avenue for users to compute these heuristics once and re-use them for varying target compression ratios.  

\end{document}